\documentclass[lettersize,journal]{IEEEtran}

\pdfoutput=1

\usepackage{amsmath,amsfonts}
\usepackage{amssymb}
\usepackage{array}
\usepackage{textcomp}
\usepackage{stfloats}
\usepackage{url}
\usepackage{verbatim}
\usepackage{graphicx}
\usepackage{cite}

\usepackage[utf8]{inputenc} % allow utf-8 input
\usepackage[T1]{fontenc}    % use 8-bit T1 fonts
\usepackage{booktabs}       % professional-quality tables
\usepackage{nicefrac}       % compact symbols for 1/2, etc.
\usepackage{microtype}      % microtypography
\usepackage{multicol}
\usepackage{color}
\usepackage[table]{xcolor}
\usepackage{bm}
\usepackage{cuted}
\usepackage{mathrsfs}
\usepackage{mathtools}
\usepackage{multirow}
\usepackage{makecell}
\usepackage{subfigure}
\usepackage{enumerate}
\usepackage[ruled,linesnumbered]{algorithm2e}
\usepackage{float}				  
\usepackage{enumitem}

\begin{document}

\title{HyCubE: Efficient Knowledge Hypergraph 3D Circular Convolutional Embedding}

\author{Zhao Li,
        Xin Wang,~\IEEEmembership{Member,~IEEE,}
        Jun Zhao,      
        Wenbin Guo,     
        Jianxin Li,~\IEEEmembership{Senior Member,~IEEE}
        % <-this % stops a space

\thanks{Manuscript received June 3, 2024.}
\thanks{(Corresponding author: Xin Wang.)}
\thanks{This work is supported by the National Science and Technology Major Project of China (2020AAA0108504), the Key Research and Development Program of Ningxia Hui Autonomous Region (2023BEG02067), the National Natural Science Foundation of China (62472311), the Ant Group Research Fund (2023061517131),  and the Australian Research Council Linkage Project (LP180100750).}

\IEEEcompsocitemizethanks{\IEEEcompsocthanksitem Zhao Li, Xin Wang, and Wenbin Guo are with the College of Intelligence and Computing, Tianjin University, Tianjin 300354, China, and also with the Tianjin Key Laboratory of Cognitive Computing and Application, Tianjin 300354, China. E-mail: \{lizh, wangx, Wenff\}@tju.edu.cn
\IEEEcompsocthanksitem Jun Zhao is with the School of Economics and Management, Ningxia University, Yinchuan 750021, China. E-mail: tzhaoj@nxu.edu.cn
\IEEEcompsocthanksitem Jianxin Li is with the Discipline of Business Systems and Operations, School of Business and Law, Edith Cowan University, Joondalup, WA, Australia. E-mail: jianxin.li@ecu.edu.au}
% <-this % stops an unwanted space
}

% The paper headers
\markboth{Journal of \LaTeX\ Class Files,~Vol.~14, No.~8, August~2024}%
{Shell \MakeLowercase{\textit{et al.}}: A Sample Article Using IEEEtran.cls for IEEE Journals}

% \IEEEpubid{0000--0000/00\$00.00~\copyright~2024 IEEE}
% Remember, if you use this you must call \IEEEpubidadjcol in the second
% column for its text to clear the IEEEpubid mark.

\maketitle

\begin{abstract}
    Knowledge hypergraph embedding models are usually computationally expensive due to the inherent complex semantic information. However, existing works mainly focus on improving the effectiveness of knowledge hypergraph embedding, making the model architecture more complex and redundant. It is desirable and challenging for knowledge hypergraph embedding to reach a trade-off between model effectiveness and efficiency. In this paper, we propose an end-to-end efficient knowledge hypergraph embedding model, HyCubE, which designs a novel \textit{3D circular convolutional neural network} and the \textit{alternate mask stack} strategy to enhance the interaction and extraction of feature information comprehensively. Furthermore, our proposed model achieves a better trade-off between effectiveness and efficiency by adaptively adjusting the 3D circular convolutional layer structure to handle $n$-ary knowledge tuples of different arities with fewer parameters. In addition, we use a knowledge hypergraph 1-N multilinear scoring way to accelerate the model training efficiency further. Finally, extensive experimental results on all datasets demonstrate that our proposed model consistently outperforms state-of-the-art baselines, with an average improvement of 8.22\% and a maximum improvement of 33.82\% across all metrics. Meanwhile, HyCubE is 6.12x faster, GPU memory usage is 52.67\% lower, and the number of parameters is reduced by 85.21\% compared with the average metric of the latest state-of-the-art baselines.
\end{abstract}


\section{Introduction}\label{sec:introduction}
\begin{figure}[ht]
    \centering
    \includegraphics[width=0.95\linewidth]{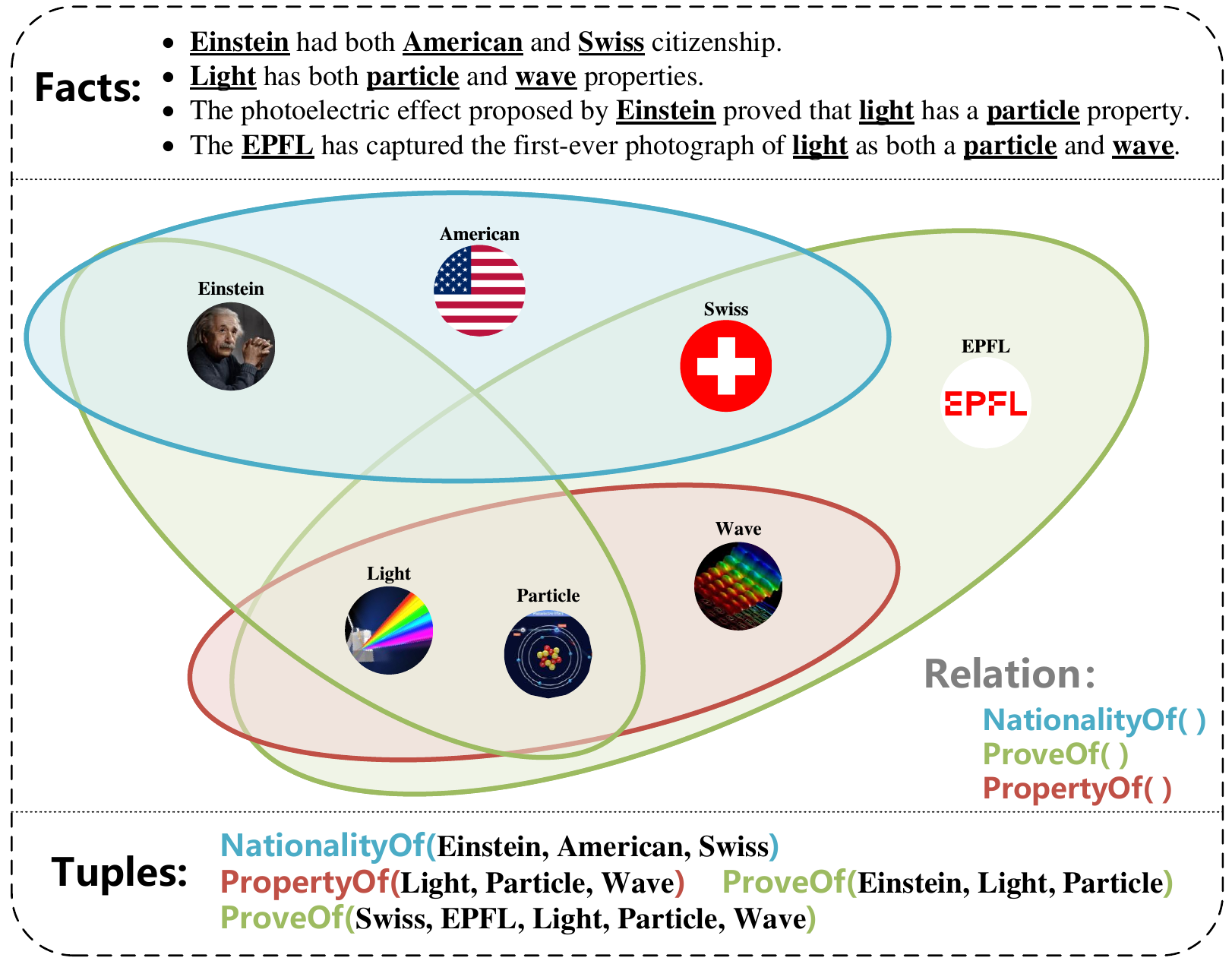}
    \caption{Examples of real-world knowledge hypergraphs.}
    \label{fig:khg}
\end{figure}

\IEEEPARstart{T}{raditional} binary relational knowledge graphs describe real-world facts in a triple form $r(e_h, e_t)$, where $r$ is the binary relation, $e_h$ and $e_t$ are the head and tail entities, respectively. However, recent studies find that considerable knowledge is beyond triple representation, e.g., in the Freebase knowledge base, more than one-third of entities are involved in non-binary relations~\cite{m-TransH}, and about 61\% of relations are beyond binary~\cite{HypE-HSimplE}. These results demonstrate that $n$-ary relational knowledge hypergraphs are ubiquitous in the real world. Moreover, $n$-ary relational knowledge hypergraphs can describe the relationships between $n \ge 2$ entities and contain more complex semantic knowledge, which is impossible with simple binary relational knowledge graphs. For example, the physicist \underline{\textit{Albert Einstein}} had both \underline{\textit{American}} and \underline{\textit{Swiss}} citizenship. As shown in Fig.~\ref{fig:khg}, the binary relational triple cannot express this fact, but the knowledge hypergraph can be defined intuitively and clearly as $\texttt{NationalityOf} (\textit{Einstein}, \textit{American}, \textit{Swiss})$.

% \textcolor{red}{For example, the mathematical community recognizes \underline{\textit{Newton}} and \underline{\textit{Leibniz}} as co-inventors of \underline{\textit{Calculus}}. The binary relation triple cannot express this fact, but the knowledge hypergraph can be defined intuitively and clearly as $\texttt{InventorOf} (\textit{Calculus}, \textit{Newton}, \textit{Leibniz})$.}

\textbf{New challenge.} Knowledge hypergraphs suffer from the same incompleteness issue as binary relational knowledge bases, which can be alleviated through link prediction (also known as knowledge completion)~\cite{ReAlE}. In recent years, knowledge embedding is an effective link prediction paradigm that has received extensive attention from academia and industry~\cite{DASFAA}. Compared with binary relational knowledge graph embedding methods, knowledge hypergraph embedding methods have significantly higher computational costs and memory usage due to more complex inherent semantic information, resulting in lower model training efficiency. However, previous knowledge hypergraph embedding models have generally focused on improving effectiveness so that model architectures have been designed to be increasingly complex and redundant. Importantly, in knowledge hypergraph embeddings with complex $n$-ary semantic and knowledge information, model efficiency is a condition that must be constrained, and it is a challenging dilemma to trade off model effectiveness and efficiency.

\textbf{Comparison of solutions.} Knowledge hypergraphs have more complex $n$-ary semantic structures than traditional binary relational knowledge graphs~\cite{RAM}~\cite{PosKHG}. Because of the powerful nonlinear complex semantic modeling capability of neural networks, convolutional neural networks (CNNs) are gradually becoming the most popular model architecture for knowledge hypergraph embedding~\cite{HyConvE}~\cite{HJE}. Furthermore, previous works have shown the inadequate ability of the Transformer architecture to capture the semantic structure information of the knowledge graphs~\cite{KGTransformer}~\cite{kgformer}. The recent knowledge hypergraph embedding model, RD-MPNN~\cite{RD-MPNN}, conducted an ablation study on the architecture of message-passing neural networks using CNNs and Transformer as decoders, respectively, and the results showed better effectiveness of the CNNs decoder (about 8\% average improvement). Consequently, the advanced natural language processing Transformer architecture is not competitive in knowledge hypergraph embeddings. In addition, existing works on binary relational knowledge graph embedding have shown that CNNs architecture can achieve comparable model effectiveness to Transformer architecture with fewer parameters~\cite{KGT5}~\cite{HittER}~\cite{SAttLE}. These researches show that CNNs architecture also has the advantage of model efficiency for knowledge embedding. Based on the analysis of previous works, the knowledge hypergraph embedding architecture using CNNs is undoubtedly superior to Transformer regarding the trade-off between model effectiveness and efficiency.

\begin{figure}[ht]
    \centering
    \includegraphics[width=1.0\linewidth]{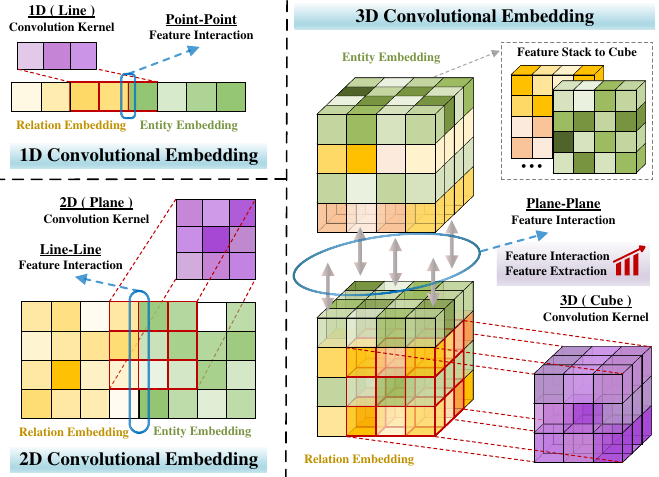}
    \caption{3D convolutional embedding vs. traditional 1D/2D convolutional embedding in feature interaction and extraction.}
    \label{fig:motivation}
\end{figure}

\textbf{Our idea.} Enhancing the feature interaction and extraction capabilities of convolutional neural networks is the key to improving the effectiveness of convolutional embedding models~\cite{ConvE}~\cite{InteractE}~\cite{Nature}~\cite{JointE}. As shown in Fig.~\ref{fig:motivation}, 3D convolutional embedding has significant feature interaction and extraction advantages over traditional 1D/2D convolutional embedding. In relation and entity feature interaction, traditional convolutional embedding can only complete the interaction of 1D (point-point) or 2D (line-line) local adjacent features. The difference is that 3D convolutional embedding builds a feature cube through the plane-plane pattern for stronger global feature interaction. Regarding feature extraction capability, the 3D convolutional kernel has a larger receptive field than the traditional convolutional 1D/2D kernel. It can complete more dimensional relation and entity feature combinations, further improving the performance of convolutional embedding. In addition, traditional knowledge hypergraph embedding models require complex and redundant operations such as decomposition and summation of different $n$-ary knowledge tuples. Accordingly, based on the property of 3D convolution with the higher dimensional receptive field, we design an end-to-end model architecture that can simultaneously process $n$-ary knowledge tuples of different arities, significantly improving the embedding efficiency of knowledge hypergraph.

\textbf{Contributions.} An excellent knowledge hypergraph embedding model should reach a trade-off between effectiveness and efficiency for real-world applications. In this paper, a novel \textit{3D circular convolutional neural network} and the \textit{alternate mask stack} strategy for knowledge hypergraph embedding are proposed to enhance complex semantic interaction and extraction with plain 3D convolution. Our proposed model, which always pays attention to the trade-off between the effectiveness and efficiency of knowledge hypergraph embedding, is named HyCubE. Specifically, HyCubE achieves end-to-end efficient knowledge hypergraph embedding by adaptively adjusting the structural parameters of 3D circular convolutional layers without the $n$-ary knowledge tuples decomposition and summation operations of other methods. This ensures that the relations in the $n$-ary knowledge tuples fully interact with each entity, reducing the amount of model parameter computation. In addition, we use a knowledge hypergraph 1-N multilinear scoring way to efficiently compute the plausibility of $n$-ary knowledge tuples to improve the model training efficiency further. Our contributions are summarized as follows:
\begin{itemize}
    \item \textbf{Enhanced Global Feature Interaction.} We propose a novel knowledge hypergraph embedding model, HyCubE, which uses the performance-enhanced \textit{3D circular convolutional neural network} and \textit{alternate mask stack} strategy to achieve global feature interactions between relation and each entity in $n$-ary knowledge tuples.
    \item \textbf{Complex Semantic Knowledge Efficient Embedding.} HyCubE attains end-to-end efficient knowledge hypergraph embedding with fewer parameters by adaptively adjusting the structural parameters of 3D circular convolutional layers. Additionally, we utilize a knowledge hypergraph 1-N multilinear scoring way to accelerate the model training efficiency further.
    \item \textbf{Consistently Better Performance.} Extensive experimental results demonstrate that our proposed model consistently outperforms state-of-the-art baselines. The metrics across all datasets improved by an average of 8.22\%, with a maximum improvement of 33.82\%.
    \item \textbf{Faster Speed and Less Memory Usage.} Compared with the latest state-of-the-art baselines, HyCubE speeds up by an average of 6.12x, reduces GPU memory usage by an average of 52.67\%, and the number of parameters is 85.21\% fewer on average.
\end{itemize}

\section{Related Work}
Existing knowledge hypergraph embedding methods can be classified into three categories: translation-based, semantic matching, and neural network methods.

\subsection{Tranlation-Based Methods}
The earliest translation-based knowledge hypergraph embedding model is m-TransH~\cite{m-TransH}, which extends the binary relational knowledge graph embedding model TransH to represent $n$-ary relations canonically. RAE~\cite{RAE} builds on the m-TransH model by using a fully connected network to add the correlation of related entities to the loss function to improve performance. NaLP~\cite{NaLP} is inspired by the RAE model, which explicitly models the relevance of all role-value pairs in $n$-ary relational facts. The translation-based methods are the classical models for early knowledge hypergraph embedding. Most of these methods struggle to capture the deep semantic information between entities and relations directly, and the models have limited expressive ability, making it difficult to achieve superior performance for complex tasks.

\subsection{Semantic Matching Methods}
GETD~\cite{GETD} extends the binary relational knowledge graph tensor decomposition model TuckER to $n$-ary relations. The storage complexity of GETD grows exponentially with the increase of relations and can only handle \textit{fixed arity} knowledge hypergraph data, which limits the practical application capability. HSimplE~\cite{HypE-HSimplE} takes inspiration from the SimplE method for binary relational knowledge graph embedding, which computes each entity in $n$-ary knowledge tuple separately. RAM~\cite{RAM} captures semantic information about entity roles using a linear combination of vectors that constrain semantically related roles to have similar representations. Similar to the RAM idea, PosKHG~\cite{PosKHG} uses a relational matrix to capture the position and role information of entities. ReAlE~\cite{ReAlE} studies knowledge hypergraph completion from the perspective of relational algebra and its core operations, which capture semantic information according to relational algebra. Most semantic matching methods are based on linear model architectures with tensor decomposition, which usually fail to capture the implicit and deep complex $n$-ary semantic information. The architecture of the models has gradually been designed to be more and more complicated, and the memory and time consumption during training have greatly limited the real-world applications of semantic matching methods.

\subsection{Neural Network Methods}
HypE~\cite{HypE-HSimplE} uses a 1D convolutional position filter to capture entity position information in each fact. HyperMLN~\cite{HyperMLN} is a statistical relational model using Markov Logic networks, which performs knowledge hypergraph embedding using the RAM in its model framework. tNaLP+~\cite{tNaLP} introduces type constraints on roles and role-value pairs using the CNN to explicitly model the relevance of role and role-value pairs in $n$-ary relational facts. RD-MPNN~\cite{RD-MPNN} uses relational message-passing neural networks as the model encoder and CNNs as the model decoder for knowledge hypergraph link prediction. HyConvE~\cite{HyConvE} jointly uses different convolutional paths to obtain superior knowledge embedding performance, but its complex model architecture results in an excessive number of redundant parameters. The comprehensive performance of such methods is generally good due to the powerful nonlinear modeling and latent semantic mining capabilities of neural networks. In contrast to state-of-the-art semantic matching models, neural network models can often support both \textit{mixed arity} and \textit{fixed arity} knowledge hypergraph modeling. As a result, neural network methods are gradually gaining attention from researchers.

\begin{figure*}[ht]
  \centering
  \includegraphics[width=1.0\linewidth]{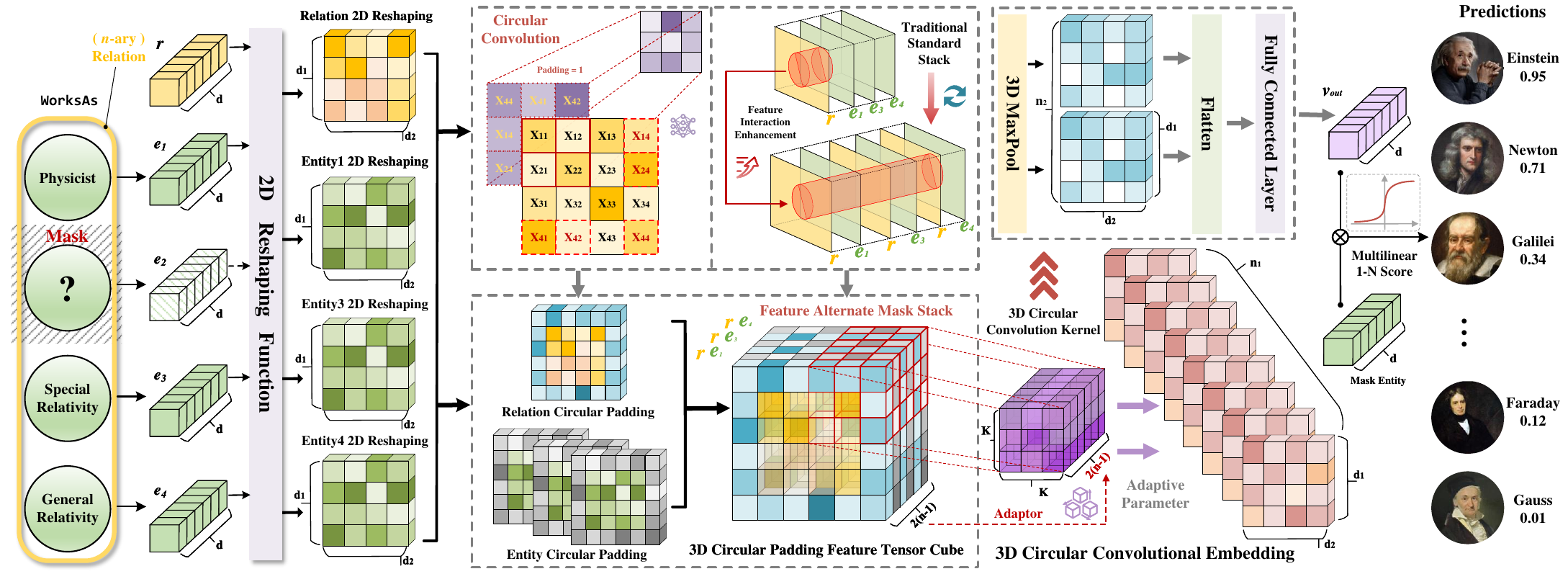}
  \caption{The overall framework of the HyCubE model.}
  \label{fig:model}
\end{figure*}

% \subsection{Problem Statement}
\section{Problem Statement}
% \subsection{Main Notations}
% The main notations used in this paper are summarized in TABLE~\ref{notations}.

% \begin{table}[h]
%     \caption{Main Notations}
%     \label{notations}
%     \centering
%     \begin{tabular}{c|c}
%     \toprule  
%     \textbf{Notations} & \textbf{Descriptions} \\
%     \hline
%     \hline
%     % $\mathcal{H}=(\mathcal{E},\mathcal{R},\mathcal{T})$ & $n$-ary knowledge hypergraph \\
%     $\mathcal{R}, \mathcal{E}$ & set of relations/entities \\
%     $\mathcal{T}$ & a knowledge tuple (fact) \\
%     $\boldsymbol{r}, \boldsymbol{e}$ & relation/entity embedding \\
%     $\overline{\boldsymbol{r}}, \overline{\boldsymbol{e}}$ & relation/entity 2D reshaping embedding \\
%     $||$ & concatenation operation \\
%     $\boldsymbol{\Lambda}$ & cumulative concatenation operation \\
%     $\mathscr{C}_{n}$ & 3D feature tensor cube \\
%     $\boldsymbol{\mathcal{C}_n}$ & 3D circular padding feature tensor cube \\    
%     $\boldsymbol{w_c}$ & 3D circular convolution kernel \\
%     $k$ & convolution kernel size \\
%     $\circledast$ & 3D circular convolution operation \\
%     $n_1$ & 3D circular convolution output channels \\
%     $\boldsymbol{\mathcal{F}_i}$ & feature maps \\
%     \bottomrule
%     \end{tabular}
% \end{table}

\subsection{Knowledge Hypergraph}
Given a finite set of entities $\mathcal{E}$, relations $\mathcal{R}$, and knowledge tuples $\mathcal{T}$, a knowledge hypergraph can be represented as $\mathcal{H} = (\mathcal{E}, \mathcal{R}, \mathcal{T})$. Each observed fact is in the form of a knowledge tuple $t=r(e_1, e_2, ..., e_i, ...,e_{n})$, where $r \in \mathcal{R}$, $e_i \in \mathcal{E}$, $t \in \mathcal{T}$, $i$ is the position of entity in knowledge tuple, and $n$ is the non-negative arity of relation $r$ representing the number of entities involved within each relation. A knowledge hypergraph consists of a large number of different knowledge tuples. When the knowledge hypergraph is \textit{mixed arity}, the arity of each knowledge tuple is different, i.e., the $n$-ary relation of each knowledge tuple contains a different number of entities ($n$ is variable). When the knowledge hypergraph is \textit{fixed arity}, the arity of each knowledge tuple is identical, i.e., the $n$-ary relation of each knowledge tuple contains the same number of entities ($n$ is constant). Generally, the \textit{fixed arity} knowledge hypergraph is a special case of the \textit{mixed arity} knowledge hypergraph. Obviously, a good knowledge hypergraph embedding model should work on both \textit{mixed arity} and \textit{fixed arity} datasets.
A traditional knowledge graph is a special case of a knowledge hypergraph where the arity of all relations is two.

\subsection{Knowledge Hypergraph Embedding}
Knowledge hypergraph embedding is an efficient link prediction (also known as completion) method that essentially learns an $n$-ary knowledge tuple $r(e_1, e_2, ..., e_{n})$ mapping function $f: \{ r \mapsto \boldsymbol{r} \in \mathbb{R}^{d}; e_{i} \mapsto \boldsymbol{e_{i}} \in \mathbb{R}^{d} \}$, where $\boldsymbol{r}$ is the embedding of the $n$-ary relation $r$, $\boldsymbol{e_{i}}$ is the embedding of the entities $e_{i}$, and $d$ is the dimension of the relation and entity embedding. Traditionally, knowledge embedding generally projects entities and relations into a low-dimensional continuous latent space to perform downstream tasks. The knowledge hypergraph link prediction task aims at predicting missing component in $n$-ary facts, where the missing component can be either an entity in the $i$-th position of the tuple $r(e_1,e_2,...,?,...,e_n)$ or an $n$-ary relation $?(e_1, e_2, ..., e_n)$.

\section{Methodology}
The overall framework of HyCubE is shown in Fig.~\ref{fig:model}. All the core components of the proposed model are designed to improve feature interaction and mine the potential deep complex semantic information. In particular, HyCubE focuses on a better trade-off between the effectiveness and efficiency of knowledge hypergraph embedding. In the following, we introduce the details of the HyCubE model and its variants. Without a special notation introduction, this paper uses lowercase letters for scalars, bold lowercase letters for vectors, and bold uppercase letters for matrices (or 3D tensors).

\subsection{Feature Alternate Mask Stack}
Consistent with previous convolutional neural network embedding methods~\cite{HyConvE}~\cite{ConvE}~\cite{JointE}~\cite{AcrE}, we must reshape the initial embedding vectors of $n$-ary relations and entities into 2D embeddings (matrices) to increase feature information interactions and better perform convolutional operations. Specifically, given an $n$-ary knowledge tuple $r(e_1, e_2, ..., e_i, ...,e_{n})$, we first randomly initialize the relation $r$ and the entities $e_{i}$ into $d$-dimensional embedding vectors $\boldsymbol{r} \in \mathbb{R}^{d}, \boldsymbol{e_i} \in \mathbb{R}^{d}$. Then, the 2D reshaping embedding of relation and entities for each knowledge tuple in the knowledge hypergraph is defined as
\begin{equation}
     \left ( \overline{\boldsymbol{r}}, \overline{\boldsymbol{e}}_{i} \right ) = \Psi \left ( \boldsymbol{r}, \boldsymbol{e_i} \right )
\end{equation}
where $\Psi$ is a 2D reshaping function: $\{ \mathbb{R}^{d} \mapsto \mathbb{R}^{d_1 \times d_2} \}$ transforms embedding vectors $\boldsymbol{r}$ and $\boldsymbol{e_i}$ into embedding matrices $\overline{\boldsymbol{r}} \in \mathbb{R}^{d_1 \times d_2}, \overline{\boldsymbol{e}}_{i} \in \mathbb{R}^{d_1 \times d_2}$, where $d_1 \times d_2 = d$.

To enhance feature interactions and better perform 3D convolutional embeddings, HyCubE stacks the 2D reshaping embeddings of relations and entities into a 3D feature tensor cube. Inspired by the mask mechanism of the large language model, HyCubE masks the predicted entity  (missing entity) $\boldsymbol{e_m}$. Furthermore, we design an \textit{alternate mask stack} strategy with feature interaction enhancement to construct an $n$-ary knowledge tuple 3D feature tensor cube, defined as
\begin{equation}
\label{cube}
     \mathscr{C}_{n} = \left [ ( \overline{\boldsymbol{r}}, \overline{\boldsymbol{e}}_{1} )||\cdots||( \overline{\boldsymbol{r}}, \overline{\boldsymbol{e}}_{m-1} )||( \overline{\boldsymbol{r}}, \overline{\boldsymbol{e}}_{m+1} )||\cdots||( \overline{\boldsymbol{r}}, \overline{\boldsymbol{e}}_{n} ) \right ]
\end{equation}
where $||$ is the concatenation operation, and the relation 2D reshaping embedding and each entity 2D reshaping embeddings in $\mathscr{C}_{n} \in \mathbb{R}^{d_1 \times d_2 \times 2(n-1)}$ generate feature interactions.

\subsection{3D Circular Convolutional Embedding}
Traditional 2D standard convolutional neural networks have been innovated into diverse improved versions for successful applications in various fields~\cite{ConvE}~\cite{JointE}~\cite{AcrE}~\cite{CircularConvolution}. As mentioned earlier, the 3D convolution in the HyCubE model architecture has powerful nonlinear deep feature modeling capabilities compared with other convolutional embedding models. Naturally, we design a novel \textit{3D circular convolutional neural network} for HyCubE to improve the feature interaction and extraction capabilities further. The \textit{3D circular convolutional neural network}, defined as
\begin{equation}
\label{convolution}
\begin{split}
  & \boldsymbol{\mathcal{F}}\left ( x, y, z \right ) = \left ( \boldsymbol{\mathcal{C}_n} \circledast \boldsymbol{w_c} \right )\left ( x, y, z \right ) = \\
  & \sum_{k_h}\sum_{k_w}\sum_{k_d} \boldsymbol{\mathcal{C}_n}(x-k_h,y-k_w,z-k_d) \boldsymbol{w_c}(k_h,k_w,k_d)
\end{split}
\end{equation}
where $\circledast$ is the 3D circular convolution operation, $\boldsymbol{w_c}$ is the 3D circular convolution kernel, and $(k_h, k_w, k_d)$ is the size of the 3D circular convolution kernel. $\boldsymbol{\mathcal{C}_n}$ is the feature tensor cube after a 3D circular padding of $\mathscr{C}_{n}$, and is defined as
\begin{equation}
\label{circular}
     \boldsymbol{\mathcal{C}_n} = \overset{\left \lfloor k_h/2 \right \rfloor}{\underset{x=-\left \lfloor k_h/2 \right \rfloor}{\boldsymbol{\Lambda}}} \overset{\left \lfloor k_w/2 \right \rfloor}{\underset{y=-\left \lfloor k_w/2 \right \rfloor}{\boldsymbol{\Lambda}}} \overset{\left \lfloor k_d/2 \right \rfloor}{\underset{z=-\left \lfloor k_d/2 \right \rfloor}{\boldsymbol{\Lambda}}} \mathscr{C}_{n}\left [ :||x, :||y, :||z \right ]
\end{equation}
where $\left \lfloor \cdot \right \rfloor$ denotes the floor function and $x, y, z \ne 0$. Also, $\boldsymbol{\Lambda}$ is the notation of the cumulative concatenation operation defined in this paper.

\begin{figure}[h]
  \centering
  \includegraphics[width=0.95\linewidth]{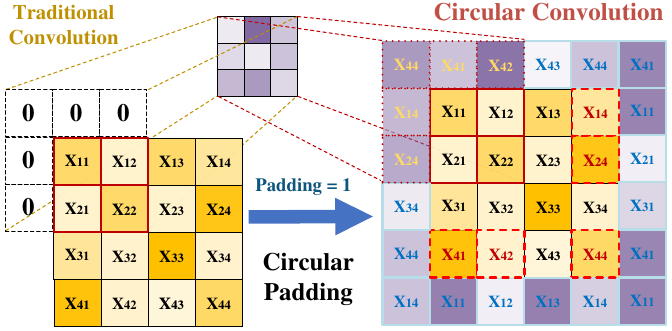}
  \caption{A front view of the 3D circular padding.}
  \label{fig:3D Circular Padding}
\end{figure}

A front view of 3D circular padding, as shown in Fig.~\ref{fig:3D Circular Padding}. Take the example of height dimension concatenation $\mathscr{C}_{n}\left [ :||i, :, : \right ]$. When $i$ is positive, it indicates that the $i$-th height plane $\mathscr{C}_{n}\left [ i, :, : \right ]$ is concatenated at the $i$-th position below the outside of the input feature $\mathscr{C}_{n}$; When $i$ is negative, it means that the reciprocal $i$-th height plane $\mathscr{C}_{n}\left [ -i, :, : \right ]$ is concatenated at the $|i|$-th position above the outside of the input feature $\mathscr{C}_{n}$. Compared with the traditional knowledge convolutional embedding methods, the \textit{3D circular convolutional neural network} performs circular padding of the 3D feature tensor cube to improve the interaction area between relations and entities, thus enhancing the effectiveness of the knowledge hypergraph embedding.

\begin{figure}[h]
  \centering
  \includegraphics[width=1.0\linewidth]{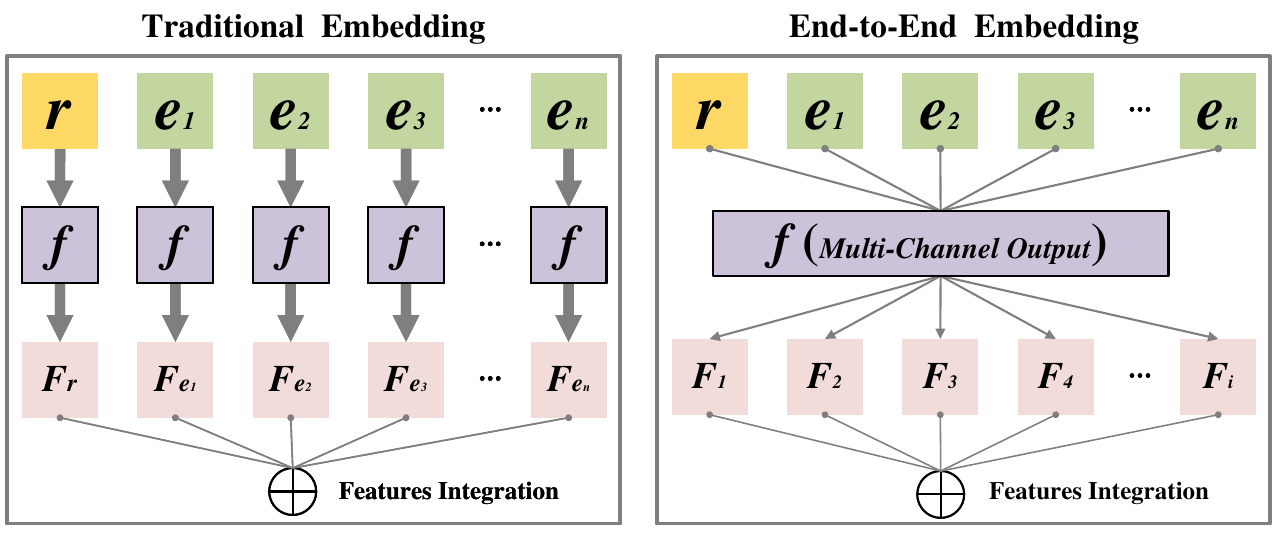}
  \caption{A comparison diagram of traditional embedding and end-to-end embedding of knowledge hypergraphs.}
  \label{fig:EndtoEnd}
\end{figure}

As described earlier, HyCubE always pursues a better trade-off between model effectiveness and efficiency, and the key is to build a new end-to-end knowledge hypergraph embedding architecture. As shown in Fig.~\ref{fig:EndtoEnd}, compared with the repetitive and redundant feature mapping process of traditional knowledge hypergraph embedding architectures, the end-to-end architecture can complete the efficient embedding of $n$-ary knowledge tuples with a single mapping. Furthermore, the end-to-end architecture can realize the embedding of the relation and all entities together, which further enhances the degree of feature interaction and helps to improve the performance of knowledge hypergraph embedding. However, the arity of each $n$-ary knowledge tuple in a knowledge hypergraph is different (i.e., $n$ is a variable), which makes it extremely difficult to construct an end-to-end efficient embedding architecture for knowledge hypergraphs.

Our proposed performance-enhanced \textit{3D circular convolutional neural network} has more flexible feature embedding dimensions, and it provides an excellent technical foundation to design the end-to-end embedding architecture for knowledge hypergraphs. Concretely, we optimize the construction of 3D circular padding feature tensor cubes for each $n$-ary knowledge tuple by constraining the circular padding operations of relations and entities in the concatenation dimension (depth dimension). This optimization operation can lay the foundation for HyCubE to achieve end-to-end efficient knowledge hypergraph embedding and also further reduce the number of redundant model parameters. First, the 3D feature tensor cube $\mathscr{C}_{n}$ of each $n$-ary knowledge tuple is enhanced by the \textit{alternate mask stack} strategy so that the degree of feature interaction between relations and entities is sufficient. Then, the end-to-end knowledge hypergraph embedding architecture enables the $n$-ary relation and all entities to totally interact and embed together without requiring feature interaction enhancement of circular padding operations. Finally, this makes it easier to implement adaptive adjustment of structural parameters in the HyCubE end-to-end model architecture. The 3D circular padding feature tensor cube $\boldsymbol{\mathcal{C}_n}$ is simplified as
\begin{equation}
\label{circular_sample}
     \boldsymbol{\mathcal{C}_n} = \overset{\left \lfloor k/2 \right \rfloor}{\underset{x=-\left \lfloor k/2 \right \rfloor}{\boldsymbol{\Lambda}}} \overset{\left \lfloor k/2 \right \rfloor}{\underset{y=-\left \lfloor k/2 \right \rfloor}{\boldsymbol{\Lambda}}} \mathscr{C}_{n}\left [ :||x, :||y, : \right ]
\end{equation}
where the size of the 3D circular convolution kernel is set to $k$, and the size of the 3D circular padding feature tensor cube is $\boldsymbol{\mathcal{C}_n} \in \mathbb{R}^{(d_1+2\left \lfloor k/2 \right \rfloor) \times (d_2+2\left \lfloor k/2 \right \rfloor) \times 2(n-1)}$.

Furthermore, to achieve the end-to-end efficient embedding architecture for HyCubE, the depth dimension of the \textit{3D circular convolutional neural network} is adaptively matched to the depth of $\boldsymbol{\mathcal{C}_n}$. Thus, HyCubE can embed $n$-ary knowledge tuples of different arities in the knowledge hypergraph simultaneously without redundant operations such as $n$-ary knowledge tuples decomposition and summation of other methods. The feature maps obtained after 3D circular convolution are
\begin{equation}
\label{convolution_sample}
      \boldsymbol{\mathcal{F}_{i}} = \boldsymbol{\mathcal{C}_n} \circledast \boldsymbol{w_c}\left ( k_h=k, k_w=k, k_d=\theta_{adp} \right )
\end{equation}
where $\boldsymbol{\mathcal{F}_{i}} \in \mathbb{R}^{d_1 \times d_2}$, $i=1,2, ..., n_1$, and $n_1$ is the number of output channels of the 3D circular convolution. $\theta_{adp}$ is a structural parameter of the \textit{3D circular convolutional neural network} that adjusts adaptively with each $n$-ary knowledge tuple in the knowledge hypergraph, always $\theta_{adp}=2(n-1)$.

To ensure model training efficiency, we use the 3D MaxPool layer to extract salient features, reduce model parameters, and mitigate overfitting.
\begin{equation}
      \boldsymbol{\mathcal{F}_{i}^{MP}} = \mathbf{MaxPool3D}\left ( \boldsymbol{\mathcal{F}_{i}} \right )
\end{equation}
where the $\mathbf{MaxPool3D}$ layer size is set to (4,1,1), $\boldsymbol{\mathcal{F}_{i}^{MP}} \in \mathbb{R}^{d_1 \times d_2}$, $i=1,2, ..., n_2$, and $n_2=n_1/4$.

After the convolution and pooling layers, we concatenate and flatten the feature maps. Finally, a $d$-dimensional vector is output through the fully connected layer as follows
\begin{equation}
      \boldsymbol{v_{out}} = \mathbf{FC}\left ( \mathbf{Flatten} \left ( \boldsymbol{\mathcal{F}_{1}^{MP}}||\boldsymbol{\mathcal{F}_{2}^{MP}}||\cdots||\boldsymbol{\mathcal{F}_{n_2}^{MP}} \right ) \right )
\end{equation}
where $\boldsymbol{v_{out}} \in \mathbb{R}^{d}$ is the output feature vector, $\mathbf{Flatten (\cdot)}$ denotes the flatten operation, $||$ is the concatenation operation.

\subsection{Model Training}
The 1-N scoring strategy has been proven to speed up model training in binary relational knowledge graph embedding~\cite{ConvE}. Inspired by this, we extend it to a 1-N multilinear scoring way for knowledge hypergraph embedding, which is used to further improve the model training efficiency of HyCubE. Specifically, we treat each entity in the knowledge hypergraph as a candidate prediction entity set, and use the output feature vector $\boldsymbol{v_{out}}$ with each entity embedding to compute the knowledge tuple plausibility score $\phi(x)$, as follows
\begin{equation}
\label{score}
    \phi(x) = \mathbf{softmax} \left( \boldsymbol{v_{out}} \cdot \boldsymbol{e_{m}^{\mathrm{T}}} + \boldsymbol{b} \right)
\end{equation}
where $\boldsymbol{b}$ is a bias term. Furthermore, HyCubE uses the necessary dropout~\cite{Dropout} and batch normalization~\cite{BatchNormalization} during the convolution process to prevent model overfitting and complete more stable convergence.

Based on the above scoring function, we develop the model training loss and learning objectives for our proposed model. At each iteration of the learning process, a batch of positive tuples is selected from the knowledge hypergraph datasets. As with other state-of-the-art works, it is necessary to train the proposed model on negative instances. Therefore, we design a negative sampling strategy for knowledge hypergraph embedding. Specifically, for each positive (true) tuple $x \in \mathcal{T}$, we generate a set of negative samples by replacing each entity in the tuple with $N$ random entities:
\begin{equation}  
	\bigcup^{n}_{i=1} \mathcal{N}^{(i)}_{x}
	\equiv
	\bigcup^{n}_{i=1}
	\left \lbrace (
	e_1, ..., 
	\bar{e}_i, ...,
	e_{n} ) \notin \mathcal{T}|\bar{e}_i \in \mathcal{E}, \bar{e}_i \neq e_i
	\right\rbrace
\end{equation}
where $\mathcal{N}^{(i)}_{x}$ represents the set of knowledge tuples after replacing the $i$-th position entity, which is a generalization of the negative sampling strategy of binary relational knowledge graph~\cite{ConvE}. Our proposed model is trained using mini-batch Stochastic Gradient Descent and AdaGrad for tuning the learning rate~\cite{AdaGrad}. The instantaneous multi-class log loss used by HyCubE can be defined as
\begin{equation}
\label{loss_function}
    \mathcal{L} = \sum_{x\in\mathcal{T}} \sum_{i=1}^{n} -\log
	\left[ 
	e^{\phi(x)} / \left( e^{\phi(x)} + 
	\sum\nolimits_{y\in\mathcal{N}^{(i)}_{x}} e^{\phi(y)}
	\right) 
	\right]
\end{equation}
where the instantaneous multi-class log loss is essentially the cross-entropy loss, which is a generalization of the binary cross-entropy loss commonly used in traditional binary relational knowledge graph completion. Algorithm~\ref{alg:training} summarizes the training procedure for HyCubE.

\SetKwInput{KwInit}{Init}
\begin{algorithm}[]
	\SetAlgoNoLine
	\caption{Training procedure for HyCubE}
        \label{alg:training}
	\KwIn{The knowledge hypergraph $\mathcal{H}=(\mathcal{E},\mathcal{R},\mathcal{T})$, the negative sampling rate $N$, the maximum number of iterations $n_{\textnormal{iter}}$ = 500}
	\KwOut{The score of each knowledge tuple}
	\KwInit{{Entity embedding matrix $\mathbf{E}$ for $e_i \in \mathcal{E}$, Relation embedding matrix $\mathbf{R}$ for $r_i \in \mathcal{R}$}}
	\For{$i=1,2, \cdots, n_\textnormal{iter}$}{
		Sample a mini-batch $\mathcal{T}_{\text{batch}}\in \mathcal{T}$; 
		\\
		\For{ \textnormal{each fact} $x \coloneqq \{r, e_1, e_2, ..., e_n\}\in\mathcal{T}_{\textnormal{batch}}$}
		{
		  \raggedright{Construct negative samples for fact $x$;} \\
		  \raggedright{$\mathscr{C}_n \leftarrow$ construct $n$-ary knowledge tuple 3D feature tensor cube using the \textit{alternate mask stack} strategy~\eqref{cube};} \\
		  \raggedright{$\boldsymbol{\mathcal{C}_n} \leftarrow$ 3D circular padding using~\eqref{circular_sample};} \\
            \raggedright{$\boldsymbol{\mathcal{F}_i} \leftarrow$ get the feature maps using~\textit{3D circular convolutional neural network}~\eqref{convolution_sample};} \\
		  \raggedright{$ \phi(x) \leftarrow$ get the final score of each knowledge tuple using 1-N multilinear scoring~\eqref{score};} \\
		}
		\raggedright{Update learnable parameters w.r.t. gradients based on the whole objective~\eqref{loss_function};}	
	}
\end{algorithm}

\section{Model Analysis and Variants}
Our proposed \textit{alternate mask stack} and \textit{3D circular padding} strategies enhance the effectiveness of knowledge hypergraph embedding by increasing the number and receptive area of relations and entities feature interactions. The effectiveness of these strategies is easily verified through some downstream tasks and ablation studies. However, a novel embedding architecture based on the \textit{3D circular convolutional neural network} can improve both effectiveness and efficiency, and its performance enhancement effect on knowledge hypergraph embedding is challenging to verify directly. Hence, we purposely propose a 2D variant of HyCubE as a direct comparison baseline model. Since the superiority of 2D convolutional embeddings over 1D convolutional embeddings has been widely proven, no comparison of the 1D variant is made in this paper. In addition, to prevent the problem of gradient vanishing, we propose a 3D residual-enhanced variant of HyCubE.

\subsection{2D Convolutional Embedding Variant}
The 2D variant of HyCubE is likewise intuitively called HyPlanE, which retains the \textit{alternate mask stack}, \textit{circular padding}, and 1-N multilinear scoring strategies to fairly and better compare model performance. Since 2D convolution cannot directly process the 3D feature tensor cube, separate convolution operations are required for the 2D reshaping embeddings of the relation and entities in an $n$-ary knowledge tuple. Specifically, circular padding the relation 2D reshaping $\overline{\boldsymbol{r}}$ and each entity 2D reshaping $\overline{\boldsymbol{e}}_{i}$ are required, defined as
\begin{equation}
      \mathscr{R} = \overset{\left \lfloor k/2 \right \rfloor}{\underset{x=-\left \lfloor k/2 \right \rfloor}{\boldsymbol{\Lambda}}} \overset{\left \lfloor k/2 \right \rfloor}{\underset{y=-\left \lfloor k/2 \right \rfloor}{\boldsymbol{\Lambda}}} \overline{\boldsymbol{r}}\left [ :||x, :||y \right ]
\end{equation}
\begin{equation}
      \mathscr{E}_i = \overset{\left \lfloor k/2 \right \rfloor}{\underset{x=-\left \lfloor k/2 \right \rfloor}{\boldsymbol{\Lambda}}} \overset{\left \lfloor k/2 \right \rfloor}{\underset{y=-\left \lfloor k/2 \right \rfloor}{\boldsymbol{\Lambda}}} \overline{\boldsymbol{e}}_i \left [ :||x, :||y \right ]
\end{equation}
where $\mathscr{R}, \mathscr{E}_i \in \mathbb{R}^{(d_1+2\left \lfloor k/2 \right \rfloor) \times (d_2+2\left \lfloor k/2 \right \rfloor)}$ are the 2D circular padding of relation and entities, respectively. Then, construct the feature map pairs of $n$-ary relation and each entity in the knowledge tuple as follows
\begin{equation}
     \boldsymbol{\mathcal{F}^{2D}_{i}} = \left ( \mathbf{Conv2D}\left ( \mathscr{R} \right ), \mathbf{Conv2D}\left ( \mathscr{E}_i \right ) \right )
\end{equation}
where $\mathbf{Conv2D(\cdot)}$ represents the traditional 2D convolution operation. Next, the feature map pairs are concatenated using the \textit{alternate mask stack} strategy as follows
\begin{equation}
     \boldsymbol{v_{conv}^{2D}} = \mathbf{vec} \left ( \boldsymbol{\mathcal{F}^{2D}_{1}} ||\! \cdots \!|| \boldsymbol{\mathcal{F}^{2D}_{m-1}} || \boldsymbol{\mathcal{F}^{2D}_{m+1}} ||\! \cdots \!|| \boldsymbol{\mathcal{F}^{2D}_{n}} \right )
\end{equation}
where $\mathbf{vec(\cdot)}$ represents vectorization operation. The output vector $\boldsymbol{v_{out}}$ of HyPlanE in the knowledge tuple plausibility score $\phi(x)$ (Equation~\ref{score}) is defined as
\begin{equation}
      \boldsymbol{v_{out}}=\boldsymbol{v_{out}^{2D}} = \mathbf{FC} \left ( \mathbf{Flatten} \left ( \boldsymbol{v_{conv}^{2D}} \right ) \right )
\end{equation}
where $\boldsymbol{v_{out}^{2D}} \in \mathbb{R}^{d}$ after vectorization, flatten, and fully connected layer operations.

\subsection{3D Residual-Enhanced Convolutional Embedding Variant}
When the knowledge hypergraph dataset has less relation-specific and the number of relation-arity, it has less inherent $n$-ary implicit semantic information. This will lead to a model training process prone to the gradient vanishing problem, affecting knowledge hypergraph embedding performance. To enhance the embedding performance of knowledge hypergraphs with less $n$-ary semantic information, we propose a 3D residual-enhanced variant named HyCubE+. Specifically, the feature maps of the 3D circular convolution output of HyCubE are flattened, as follows
\begin{equation}
      \boldsymbol{v_{conv}} = \mathbf{Flatten} \left ( \boldsymbol{\mathcal{F}_{1}^{MP}}||\boldsymbol{\mathcal{F}_{2}^{MP}}||\cdots||\boldsymbol{\mathcal{F}_{n_2}^{MP}} \right )
\end{equation}

The residual module is designed for the 3D feature tensor cube $\mathscr{C}_n$ constructed by the \textit{alternate mask stack} strategy, and the 3D residual output vector of HyCubE+ is defined as
\begin{equation}
      \boldsymbol{v_{res}} = \mathbf{Flatten} \left ( \mathbf{vec}(\mathscr{C}_n) \right )
\end{equation}
where $\mathbf{vec(\cdot)}$ represents vectorization operation. Then, the output vector $\boldsymbol{v_{out}}$ of HyCubE+ in the knowledge tuple plausibility score $\phi(x)$ (Equation~\ref{score}) is defined as
\begin{equation}
      \boldsymbol{v_{out}}=\boldsymbol{v_{out}^{res}} = \mathbf{FC}\left ( \boldsymbol{v_{conv}}+\boldsymbol{v_{res}} \right )
\end{equation}
where $\boldsymbol{v_{out}^{res}} \in \mathbb{R}^{d}$ after vectorization, flatten, and fully connected layer operations.

\begin{table*}[ht]
  \begin{center}
    \caption{Dataset Statistics}
    \label{dataset}
    \begin{tabular}{c|c|cccccc|cccc}
        \toprule
        \multicolumn{2}{c|}{\textbf{Dataset}} & $\boldsymbol{|\mathcal{E}|}$ & $\boldsymbol{|\mathcal{R}|}$ & \textbf{Arity} & \textbf{\#Train} & \textbf{\#Valid} & \textbf{\#Test} & \textbf{\#Arity=2} & \textbf{\#Arity=3} & \textbf{\#Arity=4} & \textbf{\#Arity$\boldsymbol{\geq}$ 5} \\
        \hline
        \hline
        ~\multirow{3}{*}{\textbf{\makecell{Mixed \\ Arity}}}
        & \textbf{JF17K} & 28,645 & 322 & 2-6 & 61,104 & 15,275 & 24,568 & 54,627 & 34,544 & 9,509 & 2,267 \\
        & \textbf{WikiPeople} & 47,765 & 707 & 2-9 & 305,725 & 38,223 & 38,281 & 337,914 & 25,820 & 15,188 & 3,307 \\
        & \textbf{FB-AUTO} & 3,388 & 8 & 2, 4, 5 & 6,778 & 2,255 & 2,180 & 3,786 & - & 215 & 7,212 \\
        \hline
        ~\multirow{4}{*}{\textbf{\makecell{Fixed \\ Arity}}}
        & \textbf{JF17K-3} & 11,541 & 104 & 3 & 18,910 & 4,904 & 10,730 & - & 34,544 & - & - \\
        & \textbf{JF17K-4} & 6,536 & 23 & 4 & 5,641 & 1,296 & 2,572 & - & - & 9,509 & - \\
        & \textbf{WikiPeople-3} & 12,270 & 205 & 3 & 20,509 & 2,669 & 2,642 & - & 25,820 & - & - \\
        & \textbf{WikiPeople-4} & 9,528 & 177 & 4 & 12,319 & 1,422 & 1,447 & - & - & 15,188 & - \\
        \bottomrule
    \end{tabular}
  \end{center}
% \vspace{-3px}
\thanks{The size of the \#Train, \#Valid, and \#Test columns represent the number of tuples, respectively. The "Arity" denotes the involved arities of relations.}
\end{table*}

\section{Experiments}
\subsection{Experimental Setup}
\subsubsection{Datasets}
We extensively investigate standard datasets from relevant references and select a representative set of widely used benchmarks. The experiments of \textit{mixed arity} knowledge hypergraph link prediction were conducted on three common benchmarks, i.e.,  JF17K~\cite{RAM}, WikiPeople~\cite{NaLP}, and FB-AUTO~\cite{HypE-HSimplE}. The experiments of \textit{fixed arity} knowledge hypergraph link prediction were conducted on four subsets extracted from JF17K and WikiPeople with 3/4-arity, specifically JF17K-3, JF17K-4, WikiPeople-3, and WikiPeople-4. It should be noted that previous classical works supporting \textit{fixed arity} knowledge hypergraph link prediction have blindly referenced the benchmarks in GETD~\cite{GETD}. However, our investigation and analysis revealed that the \textit{fixed arity} benchmarks provided by GETD are not directly extracted from the original data. Precisely, the number of relations, the number of entities, and the ratio of train/valid/test set divisions do not match the original dataset. The benchmark datasets specially constructed in GETD detract from the fairness and convincingness, so this paper directly extracts the original data to perform \textit{fixed arity} knowledge hypergraph link prediction. A detailed summary of the knowledge hypergraph datasets is provided in TABLE~\ref{dataset}.

\subsubsection{Baselines}
To verify the effectiveness of HyCubE, this paper conducts the knowledge hypergraph link prediction task. We select \textbf{12} classical and state-of-the-art baseline methods for comparison, including translation-based models (RAE and NaLP), semantic matching models (GETD, HypE, HSimplE, RAM, PosKHG, and ReAlE), and neural network models (HyperMLN, tNaLP+, RD-MPNN, and HyConvE).

\subsubsection{Evaluation Metrics}
In line with prior works~\cite{HyConvE}~\cite{ReAlE}~\cite{RAM}, we employ two standard evaluation metrics, namely mean reciprocal rank (MRR) and Hits@$k$, where $k$ is set to $1$, $3$, and $10$.
Each entity in each position of the tuple is first replaced by all entities in the entity set to form a collection of candidate facts, among which those facts that exist in the train/valid/test set are filtered. Specificially, let $\tau_{test}$ be the set of the test set, for each tuple $r(e_1, e_2, ..., e_{n})$ in $\tau_{test}$ and each position within the tuple, $\left |\mathcal{E}\right|-1$ corrupted tuples are generated by replacing $e_i$ with each of the entities in $\mathcal{E} \backslash \left\{e_i \right\}$. Then, We score each given tuple together with its corresponding candidate set and sort them in descending order. MRR is the mean of the inverse of rankings over all testing facts, while Hits@$k$ measures the proportion of top $k$ rankings.

% Let $rank_i(r(e_i, ..., e_{k}))$ be the ranking of $r(e_i, ..., e_{k})$, denote MRR as $\frac{1}{K}\sum_{r(e_i, ..., e_{k}) \in \tau_{test}}\sum_{i=1}^k\frac{1}{rank_i(r(e_i, ..., e_{k}))}$, where $K=\sum_{r(e_i, ..., e_{k}) \in \tau_{test}}$, We count the number of tuples in the test set that score within the top k and calculate the value of the Hit@k metric, which is a ratio determined by the top $k$ counts and the number of test set tuples.

\subsubsection{Hyperparameters}
In our experiments, the embedding dimension $d$ is set to $400$ and the batch size is taken from $\{64, 128, 256, 384, 512\}$. The maximum number of iterations is set to $500$ epochs. The training iteration is terminated during the model training process if the MRR metric does not improve for $50$ epochs continuously. The number of 3D circular convolution output channels is set to $n_1=8$, and the 3D MaxPool layer size is set to $(4, 1, 1)$. The 3D circular padding size is taken from $\{1, 2, 3, 4, 5\}$. To control the tensor size of the convolution process, the 3D circular convolution kernel size $k$ is set to $2padding+1$. Furthermore, the learning rate is selected from $0.00001$ to $0.00100$, the decay rate is chosen from $0.900$ to $0.999$, and the dropout rate is selected from $0.0$ to $0.9$.

All local experiments are obtained on 3 NVIDIA GeForce RTX 3090 GPUs and PyTorch 1.12.0 to ensure a consistent hardware and software environment. Furthermore, all baseline model hyperparameters for the local experiments are taken from the original paper or open-source code (reporting optimal experimental results).
% The detailed implementation information and hyperparameters for the HyCubE (HyCubE+) model are available at this GitHub link~\footnote{\url{https://github.com/LZ-LAB/HyCubE}}.

\begin{table*}[ht]
  \begin{center}
    \caption{Results of Link Prediction on Mixed Arity Knowledge Hypergraph Datasets}
    \vspace{-10px}
    \label{result_mixed}
\resizebox{\linewidth}{!}{ 
    \begin{tabular}{c|cccc|cccc|cccc}
        \toprule
        ~\multirow{2}{*}{\textbf{Model}} & \multicolumn{4}{c|}{\textbf{JF17K}} & \multicolumn{4}{c|}{\textbf{WikiPeople}} & \multicolumn{4}{c}{\textbf{FB-AUTO}}\\
        \cline{2-13}
        \rule{0pt}{10pt} & \textbf{MRR} & \textbf{Hits@1} & \textbf{Hits@3} & \textbf{Hits@10} & \textbf{MRR} & \textbf{Hits@1} & \textbf{Hits@3} & \textbf{Hits@10} & \textbf{MRR} & \textbf{Hits@1} & \textbf{Hits@3} & \textbf{Hits@10} \\
        \hline
        \hline
        \textbf{RAE}~\cite{RAE} & {0.392} & {0.312} & {0.433} & {0.561} & {0.253} & {0.118} & {0.343} & {0.463} & {0.703} & {0.614} & {0.764} & {0.854} \\
        \textbf{NaLP}~\cite{NaLP} & {0.310} & {0.239} & {0.334} & {0.450} & {0.338} & {0.272} & {0.362} & {0.466} & {0.672} & {0.611} & {0.712} & {0.774} \\
        \textbf{HypE}~\cite{HypE-HSimplE} & {0.494} & {0.399} & {0.532} & {0.650} & {0.263} & {0.127} & {0.355} & {0.486} & {0.804} & {0.774} & {0.824} & {0.856} \\
        \textbf{RAM}~\cite{RAM} & {0.539} & {0.463} & {0.573} & {0.690} & ({0.363}) & {0.271} & {0.405} & {0.500} & {0.830} & {0.803} & {0.851} & {0.876} \\

        \textbf{HyperMLN}~\cite{HyperMLN} & {0.556} & ({0.482}) & {0.597} & {0.717} & ${0.351}^{\dag}$ & ${0.270}^{\dag}$ & ${0.394}^{\dag}$ & ${0.497}^{\dag}$ & {0.831} & {0.803} & {0.851} & {0.877} \\
        \textbf{tNaLP+}~\cite{tNaLP} & {0.449} & {0.370} & {0.484} & {0.598} & {0.339} & {0.269} & {0.369} & {0.473} & {0.729} & {0.645} & {0.748} & {0.826} \\
        \textbf{PosKHG}~\cite{PosKHG} & {0.545} & {0.469} & {0.582} & {0.706} & ${0.315}^{\dag}$ & ${0.214}^{\dag}$ & ${0.377}^{\dag}$ & ${0.475}^{\dag}$ & {0.856} & {0.821} & {0.876} & {0.895} \\
        \textbf{ReAlE}~\cite{ReAlE} & {0.530} & {0.454} & {0.563} & {0.677} & ${0.332}^{\dag}$ & ${0.207}^{\dag}$ & (${0.417}^{\dag}$) & (${0.514}^{\dag}$) & ({0.861}) & ({0.836}) & {0.877} & ({0.908}) \\   
        \textbf{RD-MPNN}~\cite{RD-MPNN} & {0.512} & {0.445} & {0.573} & {0.685} & {-} & {-} & {-} & {-} & {0.810} & {0.714} & ({0.880}) & {0.888} \\        
        \textbf{HyConvE}~\cite{HyConvE} & ({0.580}) & {0.478} & ({0.610}) & (\underline{0.729}) & {0.362} & ({0.275}) & {0.388} & ({0.501}) & {0.847} & {0.820} & {0.872} & {0.901} \\

        \hline
        \textbf{HyCubE} (Ours) & \textbf{0.584} & \underline{0.508} & \textbf{0.616} & \textbf{0.730} & \textbf{0.448} & \textbf{0.368} & \textbf{0.490} & \textbf{0.592} & \underline{0.881} & \underline{0.860} & \underline{0.894} & \underline{0.918} \\
        \textbf{HyCubE+} (Ours) & \underline{0.582} & \textbf{0.511} & \underline{0.611} & {0.720} & \underline{0.433} & \underline{0.347} & \underline{0.478} & \underline{0.591} & \textbf{0.891} & \textbf{0.872} & \textbf{0.901} & \textbf{0.923} \\
        \textbf{HyPlanE} (Ours) & {0.569} & {0.496} & {0.600} & {0.708} & {0.402} & {0.323} & {0.443} & {0.549} & {0.866} & {0.843} & {0.880} & {0.909} \\
        \bottomrule
    \end{tabular}
    }
  \end{center}
  \thanks{The best results are in boldface, the second-best results are underlined, and the optimal baseline results are labeled with (). Experimental results for HyperMLN, PosKHG, ReAlE, RD-MPNN, and HyConvE are from the original paper. Experimental results for RAE, NaLP, HypE, RAM, and tNaLP+ are from~\cite{HyConvE}. The experimental results not presented in the original paper and obtained locally are marked with "$\dag$".}
\end{table*}

\begin{table*}[ht]
  \begin{center}
    \caption{Results of Link Prediction on Fixed Arity Knowledge Hypergraph Datasets}
    \vspace{-10px}
    \label{result_fixed}
    \resizebox{\linewidth}{!}{
    \begin{tabular}{c|ccc|ccc|ccc|ccc}
        \toprule
        ~ \multirow{2}{*}{\textbf{Model}} & \multicolumn{3}{c|}{\textbf{JF17K-3}} & \multicolumn{3}{c|}{\textbf{JF17K-4}} & \multicolumn{3}{c|}{\textbf{WikiPeople-3}} & \multicolumn{3}{c}{\textbf{WikiPeople-4}} \\
        \cline{2-13}
        \rule{0pt}{10pt} & \textbf{MRR} & \textbf{Hits@1} & \textbf{Hits@10} & \textbf{MRR} & \textbf{Hits@1} & \textbf{Hits@10} & \textbf{MRR} & \textbf{Hits@1} & \textbf{Hits@10} & \textbf{MRR} & \textbf{Hits@1} & \textbf{Hits@10} \\
        \hline
        \hline
        \textbf{GETD}~\cite{GETD} & (\underline{0.602}) & {0.525} & {0.726} & {0.750} & ({0.700}) & {0.842} & {0.303} & ({0.226}) & ({0.463}) & {0.340} & ({0.233}) & {0.531} \\
        \textbf{HypE}~\cite{HypE-HSimplE} & {0.364} & {0.255} & {0.573} & {0.408} & {0.300} & {0.627} & {0.266} & {0.183} & {0.443} & {0.304} & {0.191} & {0.527} \\
        \textbf{HSimplE}~\cite{HypE-HSimplE} & {0.429} & {0.326} & {0.612} & {0.575} & {0.508} & {0.703} & {0.233} & {0.181} & {0.342} & {0.177} & {0.151} & {0.218} \\
        \textbf{RAM}~\cite{RAM} & {0.591} & {0.516} & {0.725} & {0.717} & {0.661} & {0.813} & {0.270} & {0.205} & {0.401} & {0.223} & {0.150} & {0.378} \\

        \textbf{HyperMLN}~\cite{HyperMLN} & {0.574} & {0.501} & {0.711} & {0.734} & {0.687} & {0.831} & {0.252} & {0.193} & {0.385} & {0.224} & {0.167} & {0.370} \\
        \textbf{tNaLP+}~\cite{tNaLP} & {0.411} & {0.325} & {0.617} & {0.630} & {0.531} & {0.722} & {0.270} & {0.185} & {0.444} & ({0.344}) & {0.223} & (\underline{0.578}) \\
        \textbf{PosKHG}~\cite{PosKHG} & {0.597} & ({0.532}) & (\underline{0.727}) & {0.749} & {0.692} & ({0.855}) & {0.283} & {0.207} & {0.435} & {0.284} & {0.192} & {0.468} \\

        \textbf{ReAlE}~\cite{ReAlE} & {0.587} & {0.511} & {0.724} & {0.702} & {0.642} & {0.819} & {0.304} & {0.218} & {0.461} & {0.303} & {0.190} & {0.540} \\
        
        \textbf{RD-MPNN}~\cite{RD-MPNN} & {0.581} & {0.497} & {0.716} & {0.727} & {0.661} & {0.822} & {0.247} & {0.195} & {0.367} & {0.234} & {0.187} & {0.401} \\
        \textbf{HyConvE}~\cite{HyConvE} & {0.573} & {0.490} & {0.709} & ({0.751}) & {0.670} & {0.831} & ({0.309}) & {0.217} & {0.457} & {0.336} & {0.227} & {0.507} \\
        \hline        
        \textbf{HyCubE} (Ours) & {0.599} & \underline{0.534} & {0.723} & \underline{0.793} & \underline{0.742} & \textbf{0.887} & \underline{0.336} & \underline{0.256} & \underline{0.499} & \underline{0.367} & \underline{0.258} & \underline{0.578} \\
        \textbf{HyCubE+} (Ours) & \textbf{0.603} & \textbf{0.537} & \textbf{0.728} & \textbf{0.794} & \textbf{0.743} & \underline{0.886} & \textbf{0.345} & \textbf{0.260} & \textbf{0.515} & \textbf{0.383}& \textbf{0.269} & \textbf{0.602} \\
        \textbf{HyPlanE} (Ours) & {0.574} & {0.510} & {0.700} & {0.757} & {0.699} & {0.862} & {0.318} & {0.237} & {0.479} & {0.324}& {0.210} & {0.559} \\
        \bottomrule
    \end{tabular}
    }
  \end{center}
  \thanks{The best results are in boldface, the second-best results are underlined, and the optimal baseline results are labeled with (). All experimental results for the baseline models are obtained locally.}
\end{table*}

\begin{table*}[h]
  \begin{center}
    \caption{Results of Model Efficiency Comparison on Mixed Arity Knowledge Hypergraph Datasets}
    \label{result_mixed_eff}
    \begin{tabular}{c|ccc|ccc|ccc}
        \toprule
        ~ \multirow{2}{*}{\textbf{Model}} & \multicolumn{3}{c|}{\textbf{Parameters (Millions)}} & \multicolumn{3}{c|}{\textbf{GPU Memory Usage (MB)}} & \multicolumn{3}{c}{\textbf{Time Usage (s)}} \\
        \cline{2-10}
        \rule{0pt}{10pt} & \textbf{JF17K} & \textbf{WikiPeople} & \textbf{FB-AUTO} & \textbf{JF17K} & \textbf{WikiPeople} & \textbf{FB-AUTO} & \textbf{JF17K} & \textbf{WikiPeople} & \textbf{FB-AUTO} \\
        \hline
        \hline
        \textbf{RAM}~\cite{RAM} & {$\approx$ 14.24} & {$\approx$ 27.34} & {$\approx$ 1.63} & {15,174} & {20,742} & {2,860} & {74.2} & {215.8} & {4.0} \\
        \textbf{PosKHG}~\cite{PosKHG} & {$\approx$ 14.34} & {$\approx$ 27.53} & {$\approx$ 1.65} & {15,388} & {21,147} & {2,874} & {87.4} & {229.7} & {4.3} \\
        \textbf{HyConvE}~\cite{HyConvE} & {$\approx$ 12.80} & {$\approx$ 21.44} & {$\approx$ 4.80} & {7,718} & {15,430} & {3,032} & {98.7} & {247.3} & {4.9} \\
        \textbf{ReAlE}~\cite{ReAlE} & {$\approx$ 14.88} & {$\approx$ 29.61} & {$\approx$ 1.64} & {16,316} & {17,970} & {14,488} & {333.7} & {507.9} & {13.9} \\
        
        \hline
        \textbf{HyCubE} (Ours) & \textbf{$\approx$ 1.28} & \textbf{$\approx$ 2.24} & \textbf{$\approx$ 0.96} & \underline{2,536} & \textbf{3,108} & \textbf{1,704} & \textbf{9.7} & \underline{40.1} & \underline{2.9} \\
        \textbf{HyCubE+} (Ours) & \underline{$\approx$ 5.77} & \underline{$\approx$ 13.46} & \underline{$\approx$ 1.28} & \textbf{2,290} & \underline{3,424} & \underline{1,914} & \underline{14.9} & \textbf{33.7} & \textbf{1.9} \\
        \textbf{HyPlanE} (Ours) & {$\approx$ 11.52} & {$\approx$ 19.20} & {$\approx$ 3.84} & {2,628} & {3,482} & {1,972} & {17.4} & {42.1} & {4.1} \\
        \bottomrule
    \end{tabular}
  \end{center}
  \thanks{The best results are in boldface and the second-best results are underlined.}
\end{table*}

\begin{table*}[h]
  \begin{center}
    \caption{Results of Model Efficiency Comparison on Fixed Arity Knowledge Hypergraph Datasets}
    \label{result_fixed_eff} 
    \begin{tabular}{c|cccc|cccc|cccc}
        \toprule
         ~ \multirow{2}{*}{\textbf{Model}} & \multicolumn{4}{c|}{\textbf{Parameters (Millions)}} & \multicolumn{4}{c|}{\textbf{GPU Memory Usage (MB)}} & \multicolumn{4}{c}{\textbf{Time Usage (s)}} \\
        \cline{2-13}
        \rule{0pt}{10pt} & \textbf{JF-3} & \textbf{JF-4} & \textbf{WP-3} & \textbf{WP-4} & \textbf{JF-3} & \textbf{JF-4} & \textbf{WP-3} & \textbf{WP-4} & \textbf{JF-3} & \textbf{JF-4} & \textbf{WP-3} & \textbf{WP-4} \\
        \hline
        \hline
        \textbf{RAM}~\cite{RAM} & {$\approx$ 4.62} & {$\approx$ 2.61} & {$\approx$ 4.91} & {$\approx$ 3.81} & {3,114} & {2,370} & {1,982} & {1,962} & {14.9} & {5.4} & {11.9} & {10.9} \\
        \textbf{PosKHG}~\cite{PosKHG} & {$\approx$ 4.67} & {$\approx$ 2.64} & {$\approx$ 4.95} & {$\approx$ 3.85} & {3,214} & {2,375} & {2,180} & {2,007} & {14.8} & {5.3} & {11.7} & {10.5} \\
        \textbf{HyConvE}~\cite{HyConvE} & {$\approx$ 6.72} & {$\approx$ 9.28} & {$\approx$ 6.72} & {$\approx$ 9.28} & {2,774} & {3,354} & {2,580} & {2,248} & {22.4} & {9.5} & {19.1} & {17.7} \\
        \textbf{ReAlE}~\cite{ReAlE} & {$\approx$ 4.66} & {$\approx$ 2.62} & {$\approx$ 4.99} & {$\approx$ 3.88} & {4,346} & {5,562} & {4,348} & {5,564} & {46.7} & {14.6} & {41.5} & {30.6} \\

        \hline        
        \textbf{HyCubE} (Ours) & \textbf{$\approx$ 0.32} & \textbf{$\approx$ 0.32} & \textbf{$\approx$ 0.32} & \textbf{$\approx$ 0.32} & \underline{1,972} & \textbf{1,760} & \textbf{1,906} & \textbf{1,790} & \textbf{4.9} & \textbf{1.9} & \textbf{5.4} & \underline{8.9} \\
        \textbf{HyCubE+} (Ours) & \underline{$\approx$ 1.60} & \underline{$\approx$ 1.93} & \underline{$\approx$ 1.60} & \underline{$\approx$ 1.93} & \textbf{1,928} & \underline{1,890} & \underline{1,940} & \underline{1,906} & \underline{7.2} & \underline{2.2} & \underline{6.4} & \textbf{7.4} \\
        \textbf{HyPlanE} (Ours) & {$\approx$ 6.40} & {$\approx$ 8.96} & {$\approx$ 6.40} & {$\approx$ 8.96} & {2,052} & {1,914} & {1,968} & {1,948} & {9.4} & {4.4} & {9.7} & {10.2} \\
        \bottomrule
    \end{tabular}
  \end{center}
  \thanks{The best results are in boldface and the second-best results are underlined. JF and WP are short for JF7K and WikiPeople, respectively.}
\end{table*}

\subsection{Mixed Arity Knowledge Hypergraph Results}
The experimental results of the \textit{mixed arity} knowledge hypergraph link prediction are shown in TABLE~\ref{result_mixed}, which demonstrate the effectiveness of our proposed model. We can clearly see that the optimal baseline model is different across datasets and metrics. Because state-of-the-art baselines have different strengths, it is challenging for knowledge hypergraph embedding models to gain the best results for most metrics. It is encouraging that our proposed model achieves consistently optimal results in terms of effectiveness, further demonstrating the superiority of the 3D circular convolutional embedding architecture of HyCubE (HyCubE+). Our proposed model consistently outperforms all the baselines on this core task of \textit{mixed arity} knowledge hypergraph embedding, with an average improvement of 9.13\% and a maximum improvement of 33.82\% over the best baseline on all metrics.

In particular, HyCubE (HyCubE+) shows an average improvement of 5.46\% on all metrics compared with the 2D variant model HyPlanE. The \textit{mixed arity} knowledge hypergraph link prediction experiments have demonstrated that 3D convolutional embedding outperforms 2D convolutional embedding in feature interaction and extraction. HyCubE (HyCubE+) relies on the powerful nonlinear modeling capability of the performance-enhanced \textit{3D circular convolutional neural network} to capture knowledge hypergraph latent and deep complex semantic information. Moreover, the \textit{alternate mask stack} strategy of HyCubE (HyCubE+) enables $n$-ary relations to fully interact with each entity, which improves the expressiveness and robustness of the model. It is worth stating that the performance of HyPlanE is also good enough to compete with state-of-the-art baselines since HyPlanE uses the \textit{circular padding} and \textit{alternate mask stack} strategy of the 2D convolutional version. The performance of HyPlanE further demonstrates the superiority of our proposed the \textit{3D circular convolutional neural network} and \textit{alternate mask stack} strategy.

From the details of the \textit{mixed arity} knowledge hypergraph datasets in TABLE~\ref{dataset}, it can be seen that JF17K (322) and WikiPeople (707) have high relation-specific, and FB-AUTO (8) has low relation-specific. Besides, JF17K (2-6) and WikiPeople (2-9) have a higher number of relation-arity, while FB-AUTO (2, 4, 5) has a lower number of relation-arity. Therefore, it can be assumed that JF17K and WikiPeople have more $n$-ary implicit semantic information, while FB-AUTO has less $n$-ary implicit information. HyCubE generally outperforms HyCubE+ on datasets (JF17K and WikiPeople) with more $n$-ary implicit semantic information, and HyCubE+ outperforms HyCubE on datasets (FB-AUTO) with less $n$-ary implicit semantic information. This experimental result demonstrates that the 3D residual module of HyCubE+ can alleviate the problem of model training on datasets with less $n$-ary implicit semantic information, which is prone to gradient vanishing.

\subsection{Fixed Arity Knowledge Hypergraph Results}
As the problem described in the dataset subsection, we must implement all baseline models locally. The results of the \textit{fixed arity} knowledge hypergraph link prediction experiments are shown in TABLE~\ref{result_fixed}, which demonstrate the effectiveness of our proposed model. Our proposed model consistently outperforms all the baselines on \textit{fixed arity} knowledge hypergraphs, with an average improvement of 7.30\% and a maximum improvement of 15.45\% over the best baseline on all metrics. It should be noted that the average effectiveness of the \textit{fixed arity} knowledge hypergraph link prediction experiment results contains the Hits@3 metric (not shown due to space constraints). Notably, GETD gains the optimal baseline results at most for the \textit{fixed arity} knowledge hypergraph experiments. GETD is explicitly designed to model \textit{fixed arity} knowledge hypergraph data, and it deserves such results. However, a good knowledge hypergraph embedding model should be able to handle both \textit{mixed arity} and \textit{fixed arity} knowledge hypergraph datasets.

In particular, HyCubE (HyCubE+) shows an average improvement of 9.09\% on all metrics compared with the 2D variant HyPlanE. The \textit{fixed arity} knowledge hypergraph link prediction experiments have also demonstrated that 3D convolutional embedding outperforms 2D convolutional embedding in feature interaction and extraction. The effectiveness of HyPlanE on \textit{fixed arity} knowledge hypergraph datasets is still comparable to state-of-the-art baseline methods. Furthermore, \textit{fixed arity} knowledge hypergraph datasets have significantly less $n$-ary implicit semantic information than \textit{mixed arity} knowledge hypergraph datasets, so 3D residual-enhanced HyCubE+ can achieve better effectiveness in \textit{fixed arity} knowledge hypergraphs. HyCubE (HyCubE+) continues to show excellent model stability on \textit{fixed arity} knowledge hypergraphs.

\begin{figure*}[h]
	\centering
	\subfigure[JF17K]{
		\includegraphics[width=0.199\textwidth, height=0.159\textwidth]{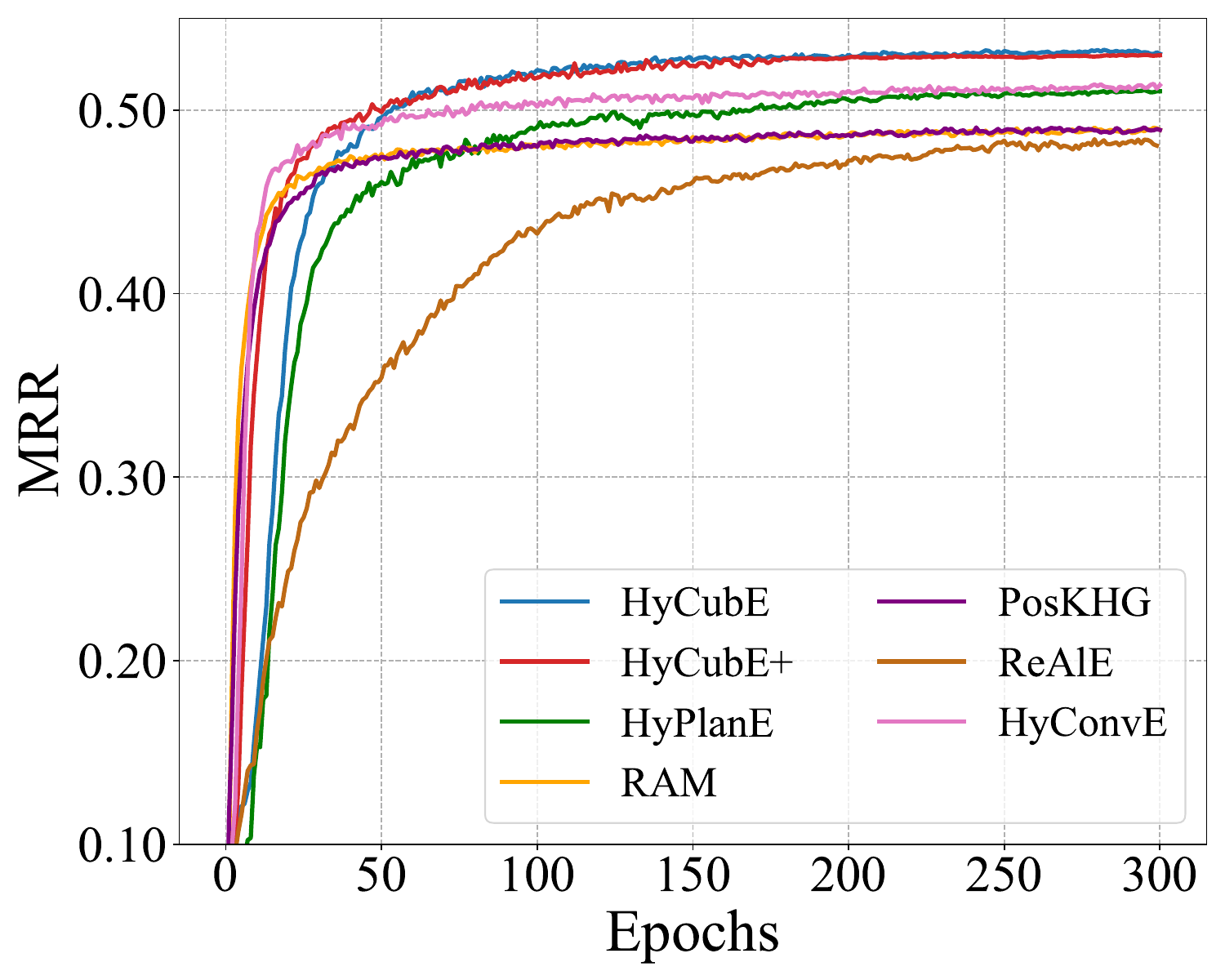}}
	\hspace{-10px}
	\subfigure[WikiPeople]{
		\includegraphics[width=0.199\textwidth, height=0.159\textwidth]{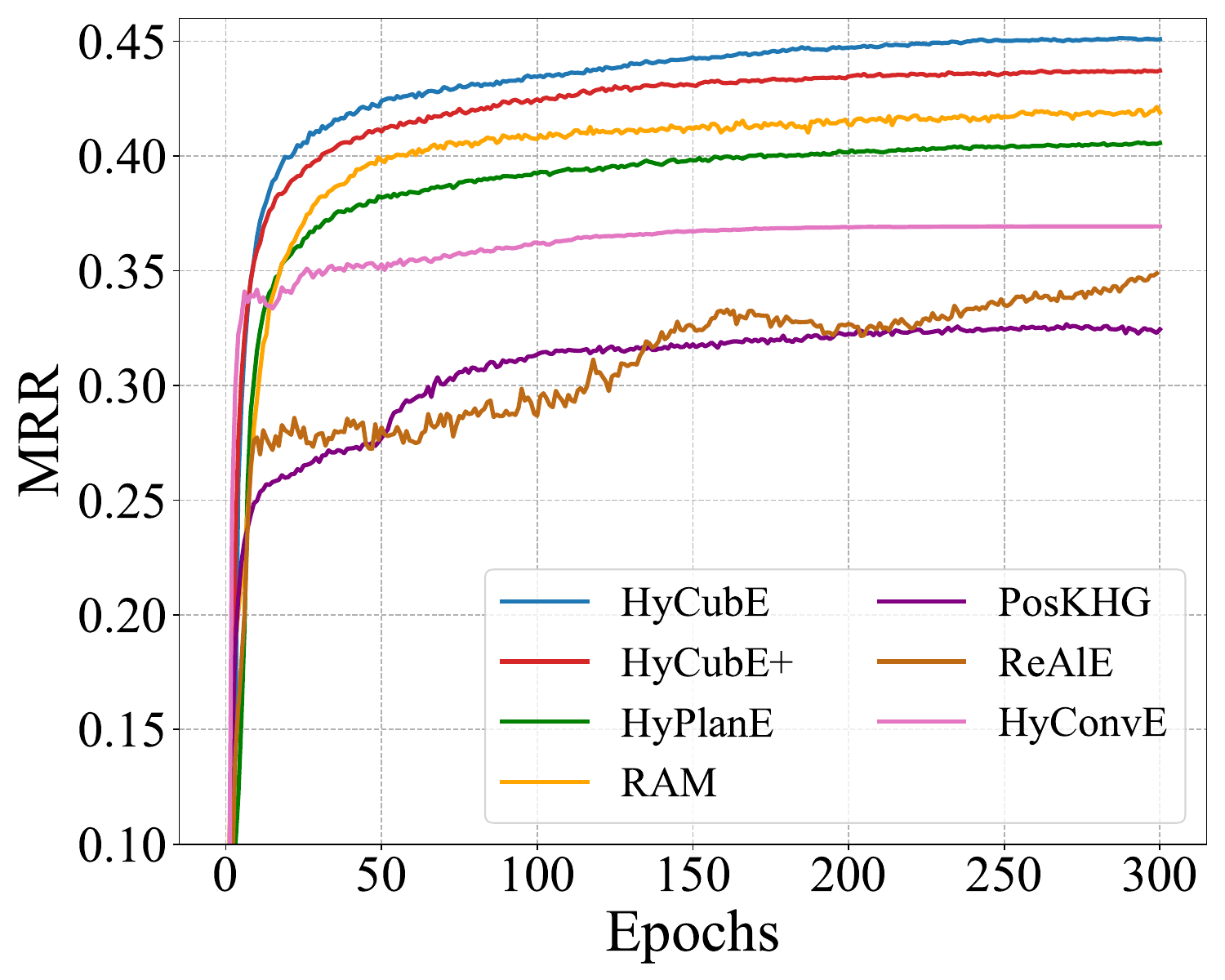}}
	\hspace{-10px}
	\subfigure[FB-AUTO]{
		\includegraphics[width=0.199\textwidth, height=0.159\textwidth]{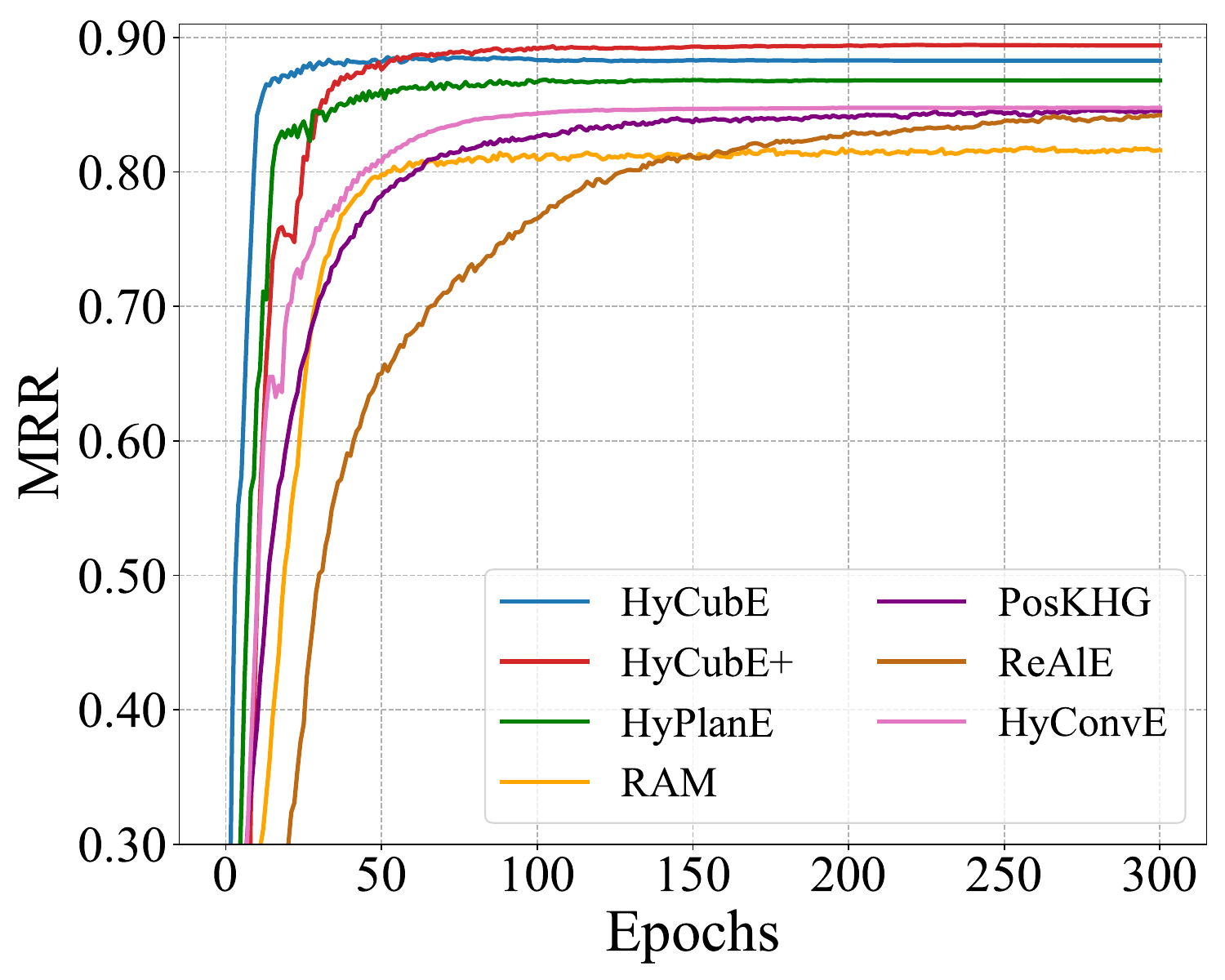}}
	\hspace{-10px}
	\subfigure[WikiPeople-3]{
		\includegraphics[width=0.199\textwidth, height=0.159\textwidth]{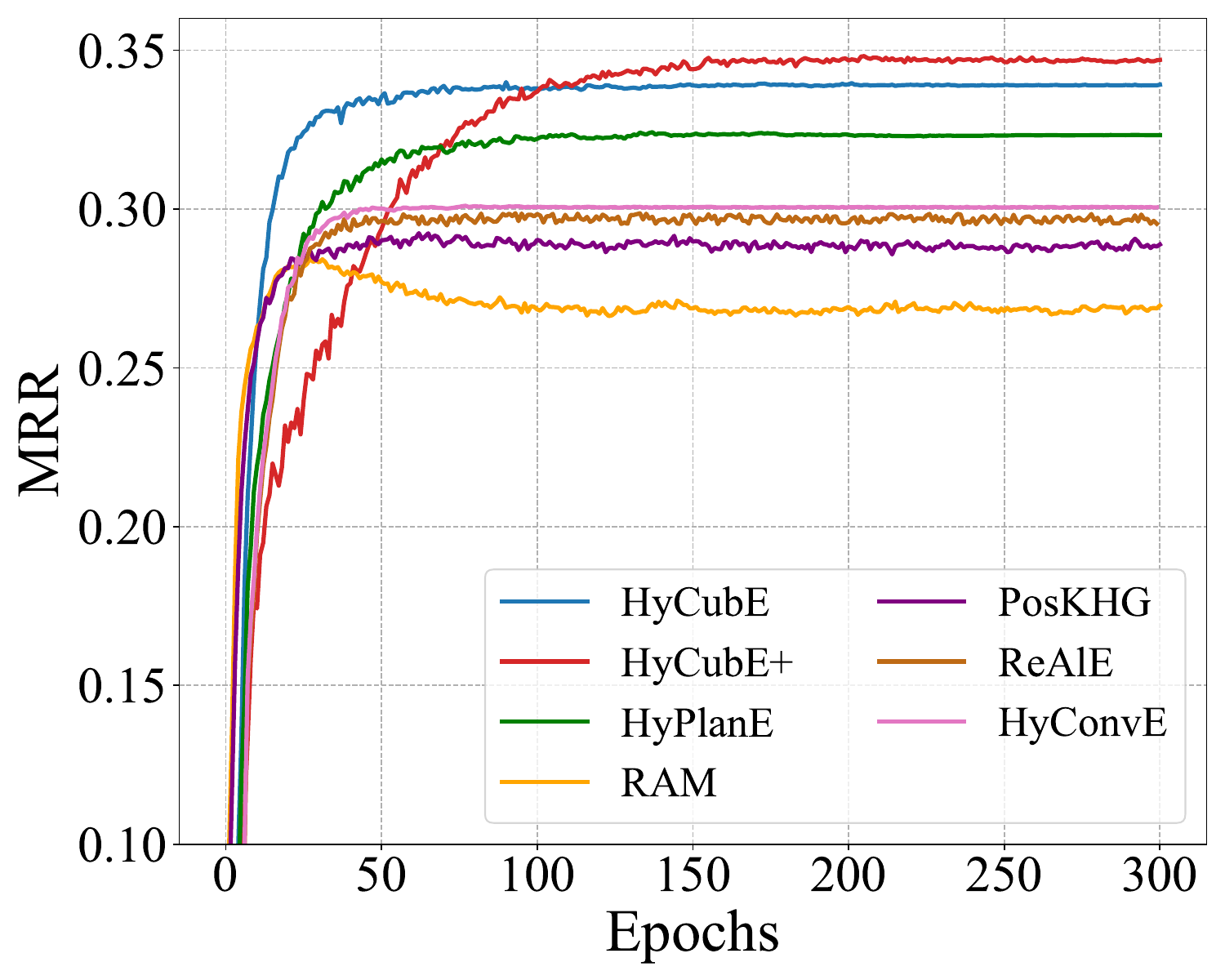}}
	\hspace{-10px}
	\subfigure[JF17K-4]{
		\includegraphics[width=0.199\textwidth, height=0.159\textwidth]{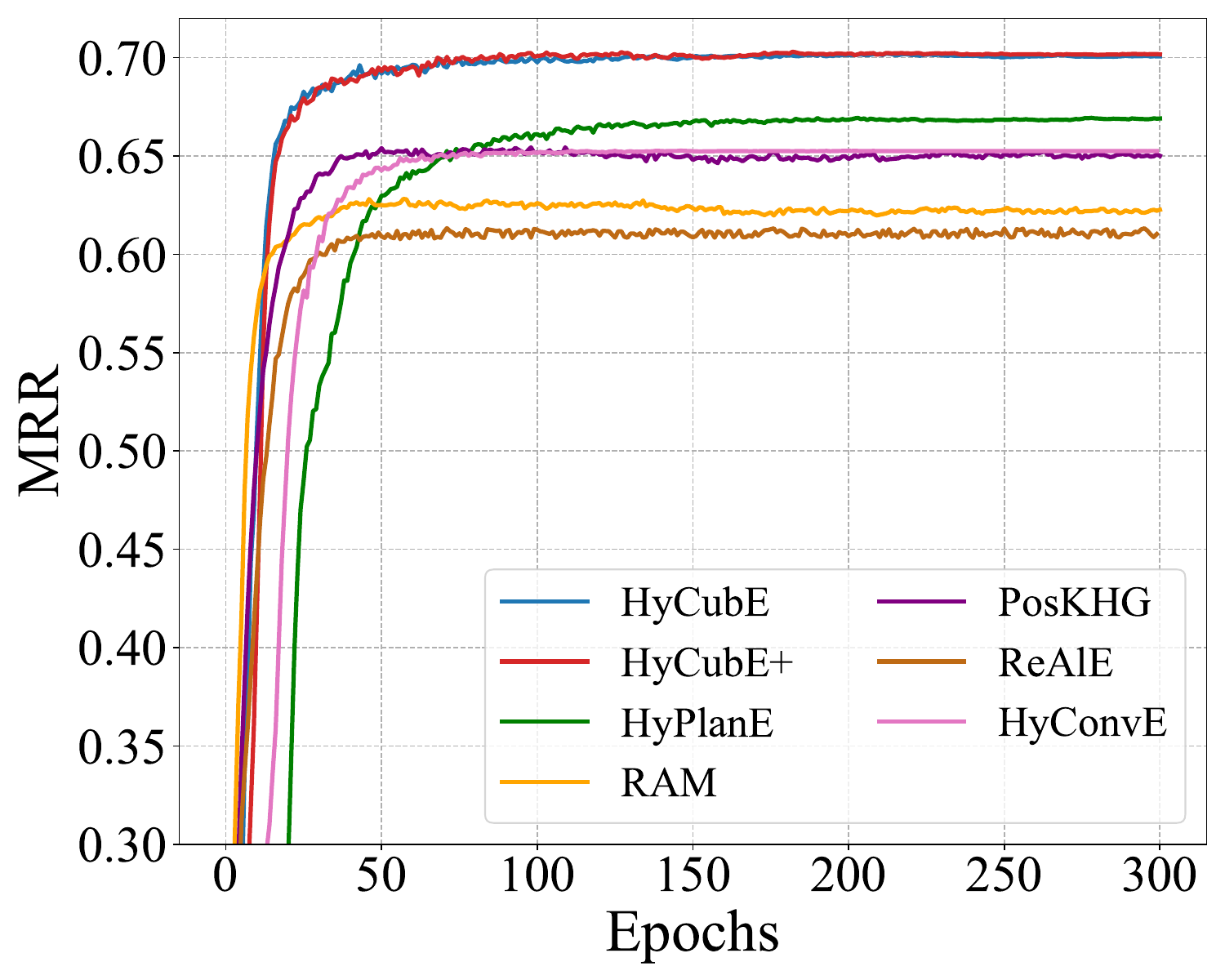}}
	\caption{The model training convergence curves (MRR metric) on knowledge hypergraph datasets.}
 \label{curve:ModelTraining_MRR}
\end{figure*}

\begin{figure*}[h]
	\centering
	\subfigure[JF17K]{
		\includegraphics[width=0.199\textwidth, height=0.159\textwidth]{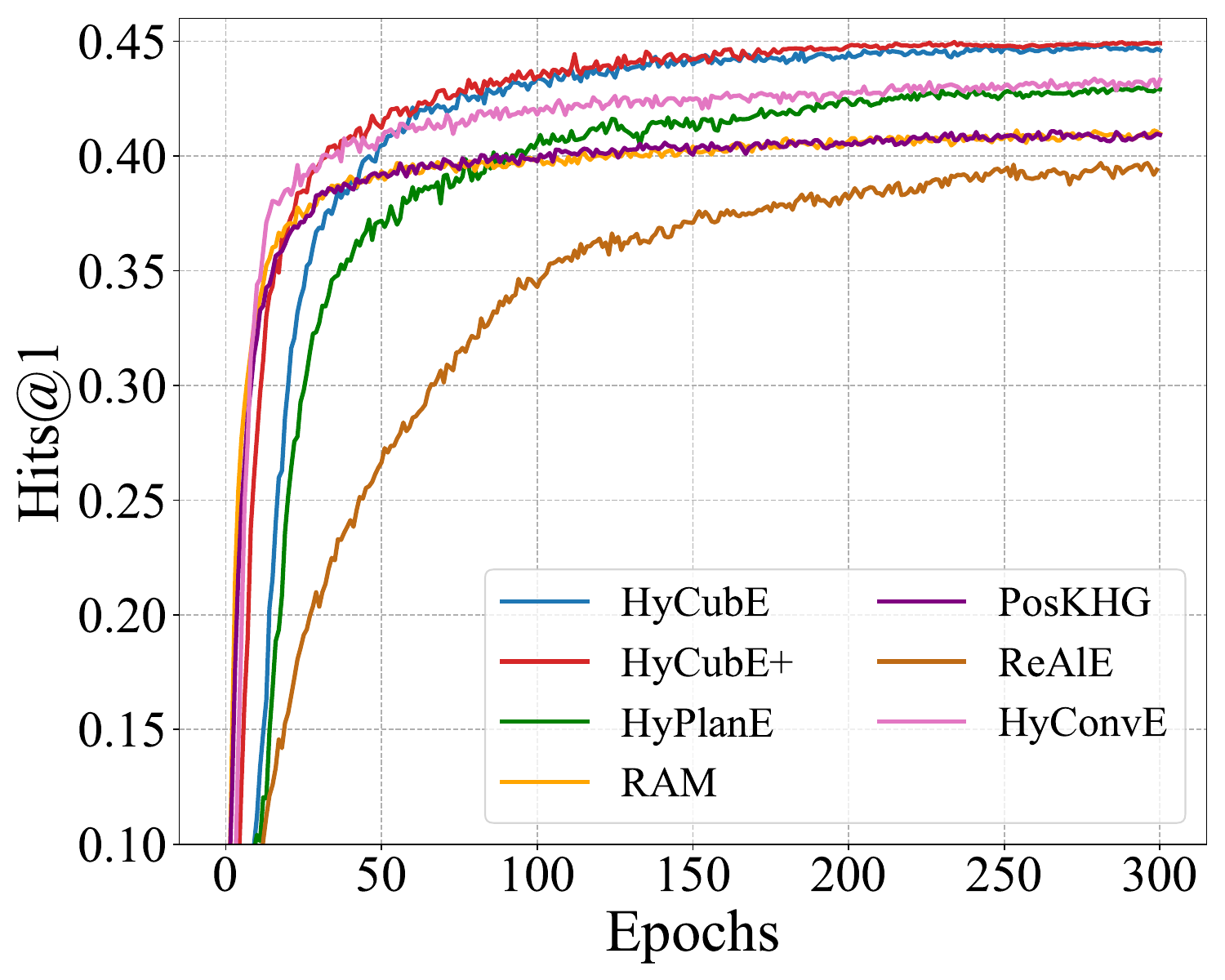}}
	\hspace{-10px}
	\subfigure[WikiPeople]{
		\includegraphics[width=0.199\textwidth, height=0.159\textwidth]{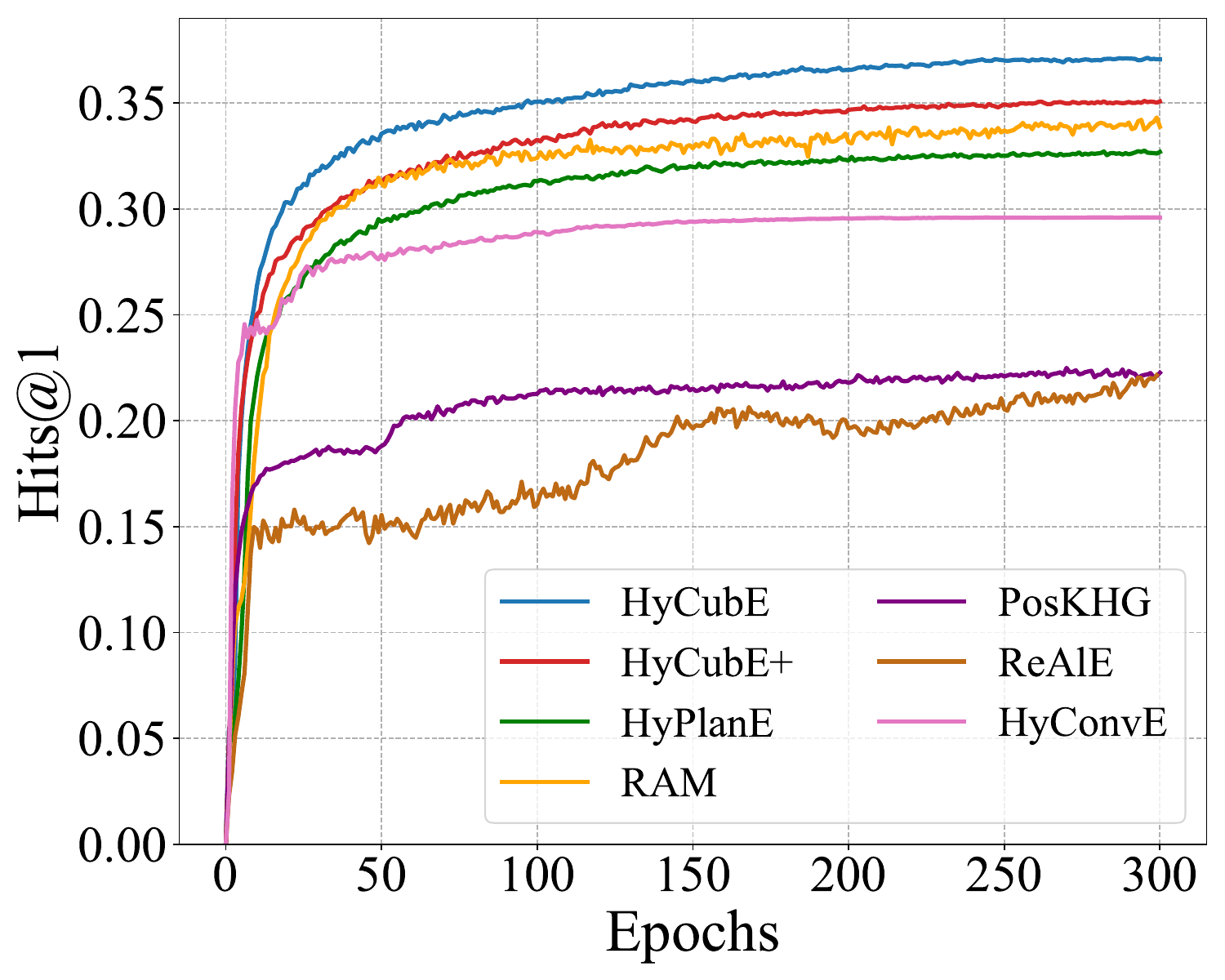}}
	\hspace{-10px}
	\subfigure[FB-AUTO]{
		\includegraphics[width=0.199\textwidth, height=0.159\textwidth]{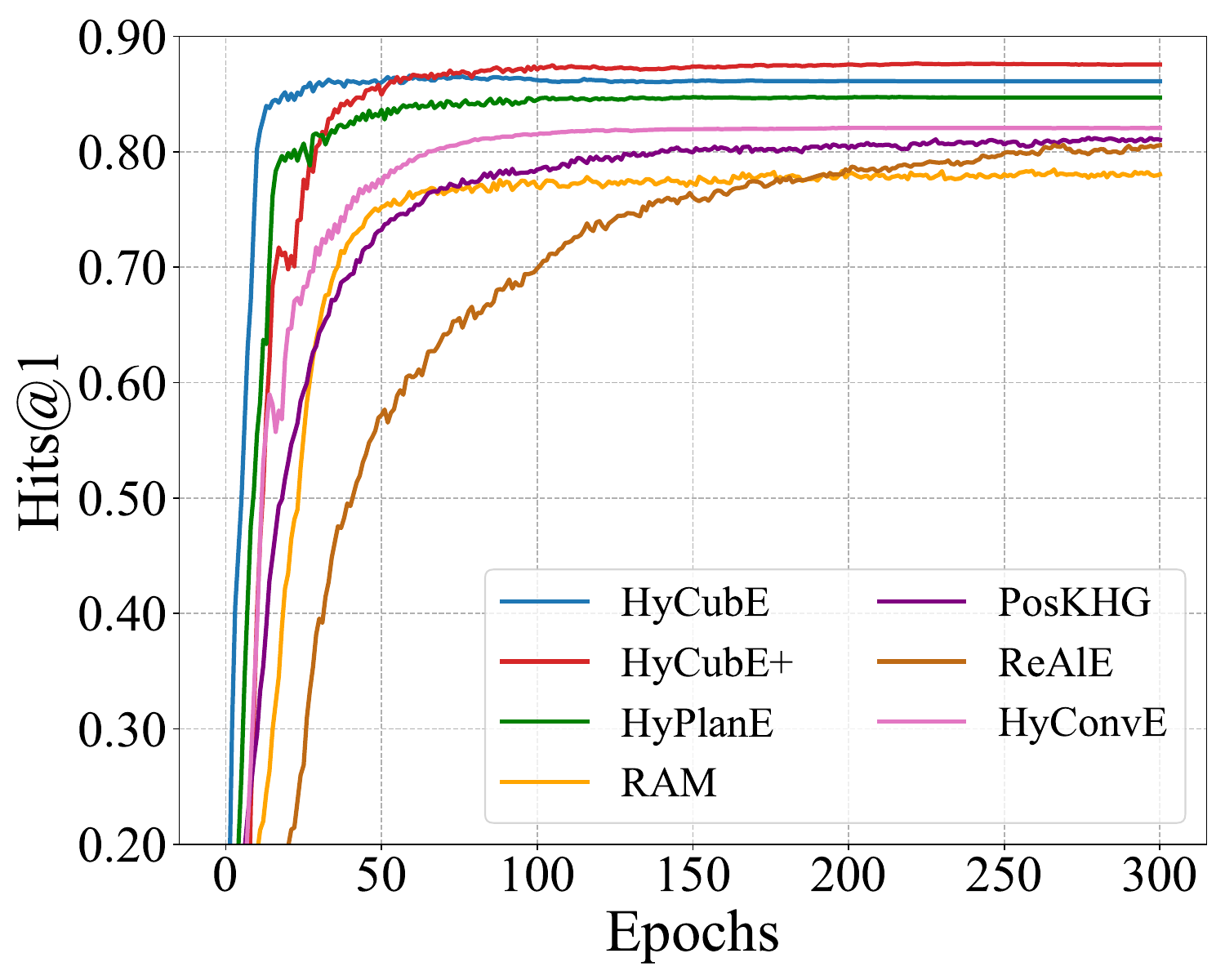}}
	\hspace{-10px}
	\subfigure[WikiPeople-3]{
		\includegraphics[width=0.199\textwidth, height=0.159\textwidth]{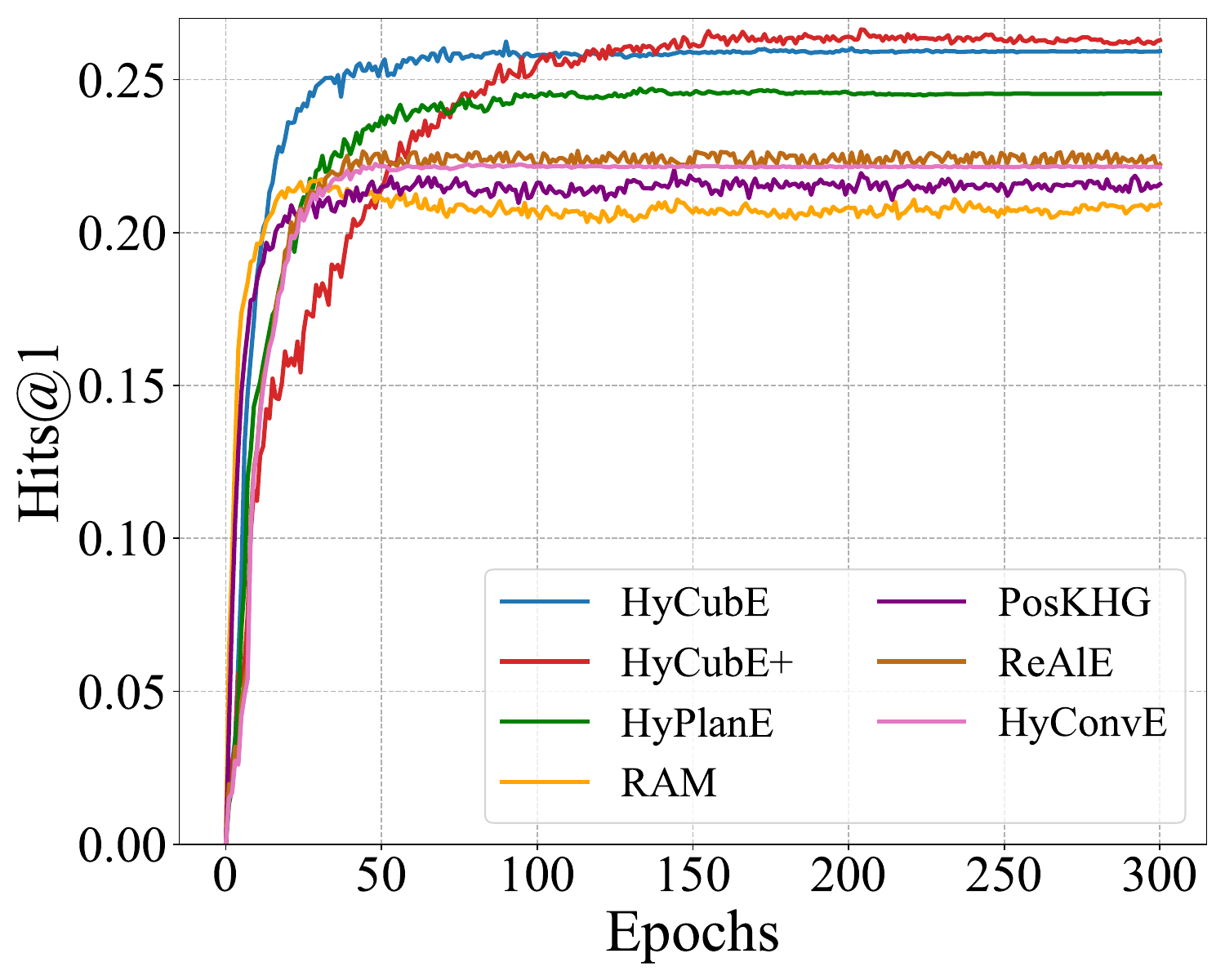}}
	\hspace{-10px}
	\subfigure[JF17K-4]{
		\includegraphics[width=0.199\textwidth, height=0.159\textwidth]{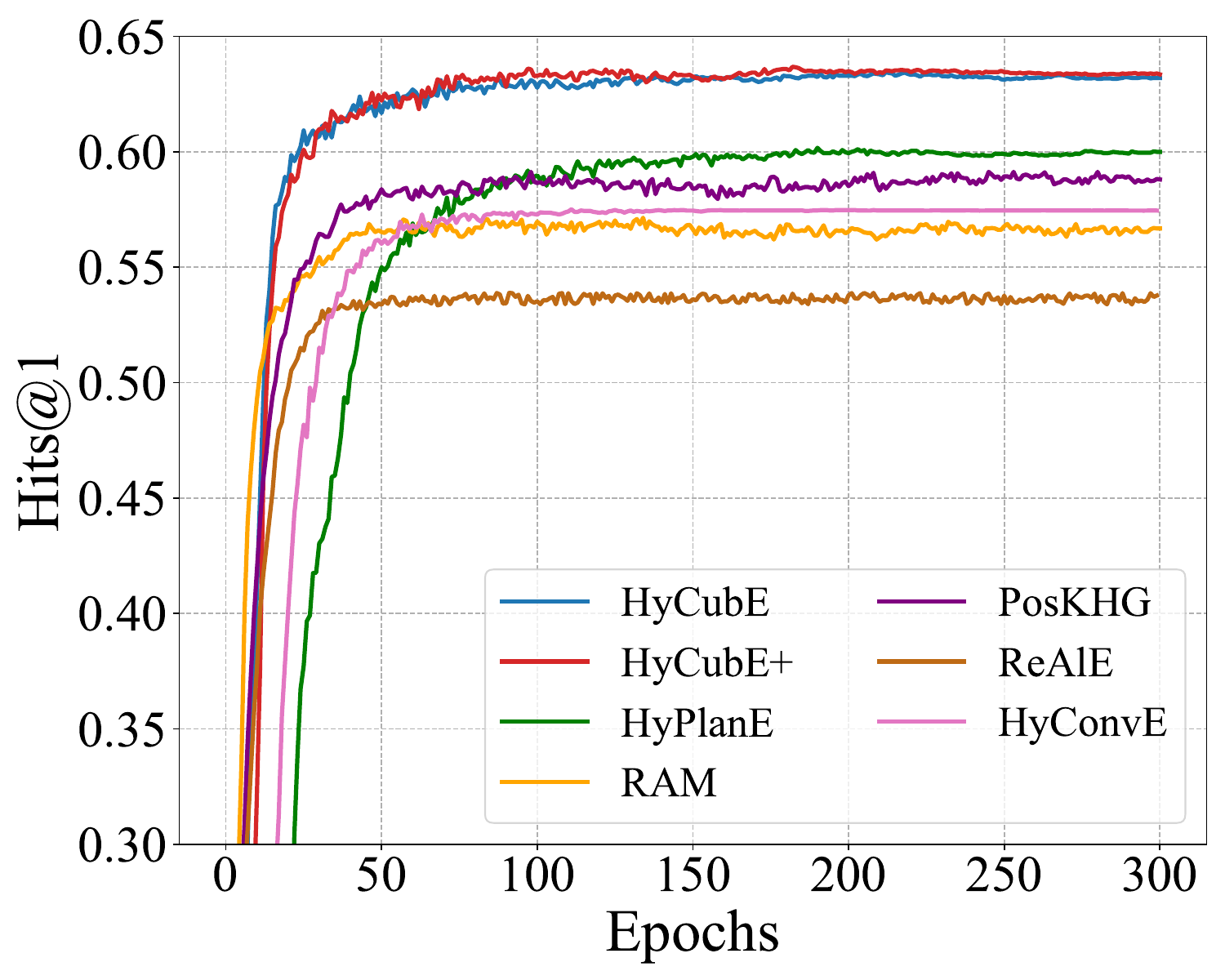}}
	\caption{The model training convergence curves (Hits@1 metric) on knowledge hypergraph datasets.}
 \label{curve:ModelTraining_H1}
\end{figure*}

\begin{figure*}[h]
	\centering
	\subfigure[JF17K]{
		\includegraphics[width=0.199\textwidth, height=0.159\textwidth]{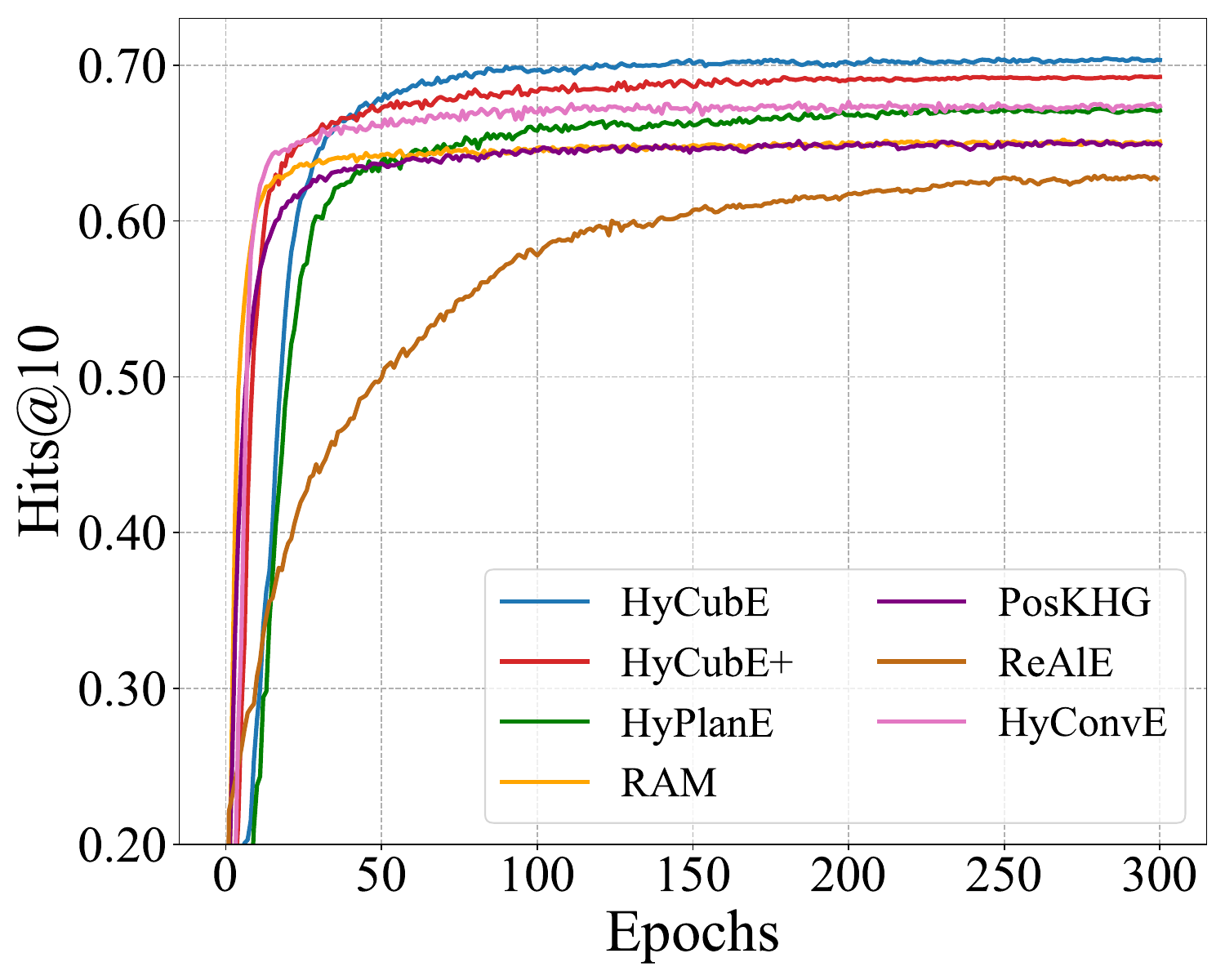}}
	\hspace{-10px}
	\subfigure[WikiPeople]{
		\includegraphics[width=0.199\textwidth, height=0.159\textwidth]{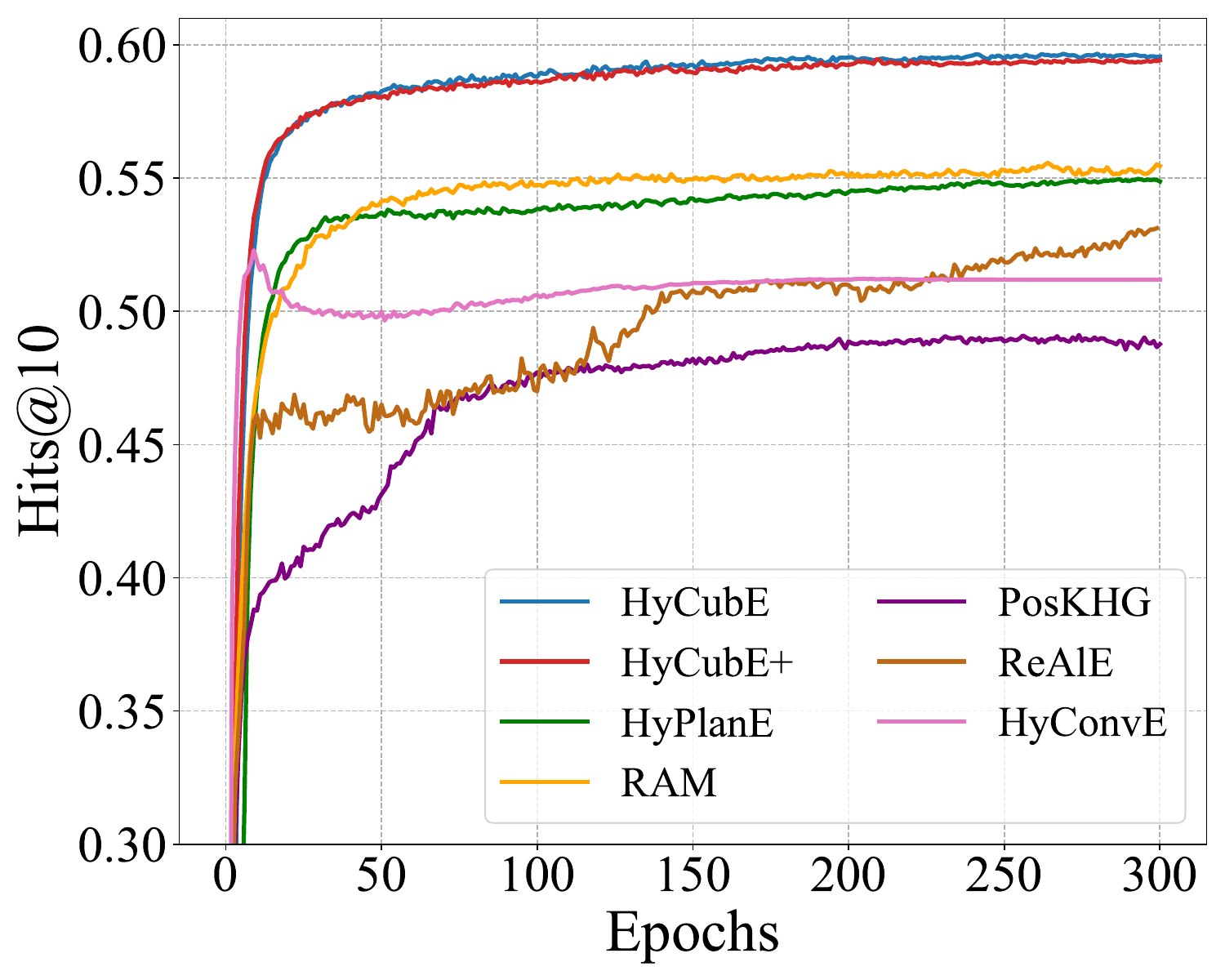}}
	\hspace{-10px}
	\subfigure[FB-AUTO]{
		\includegraphics[width=0.199\textwidth, height=0.159\textwidth]{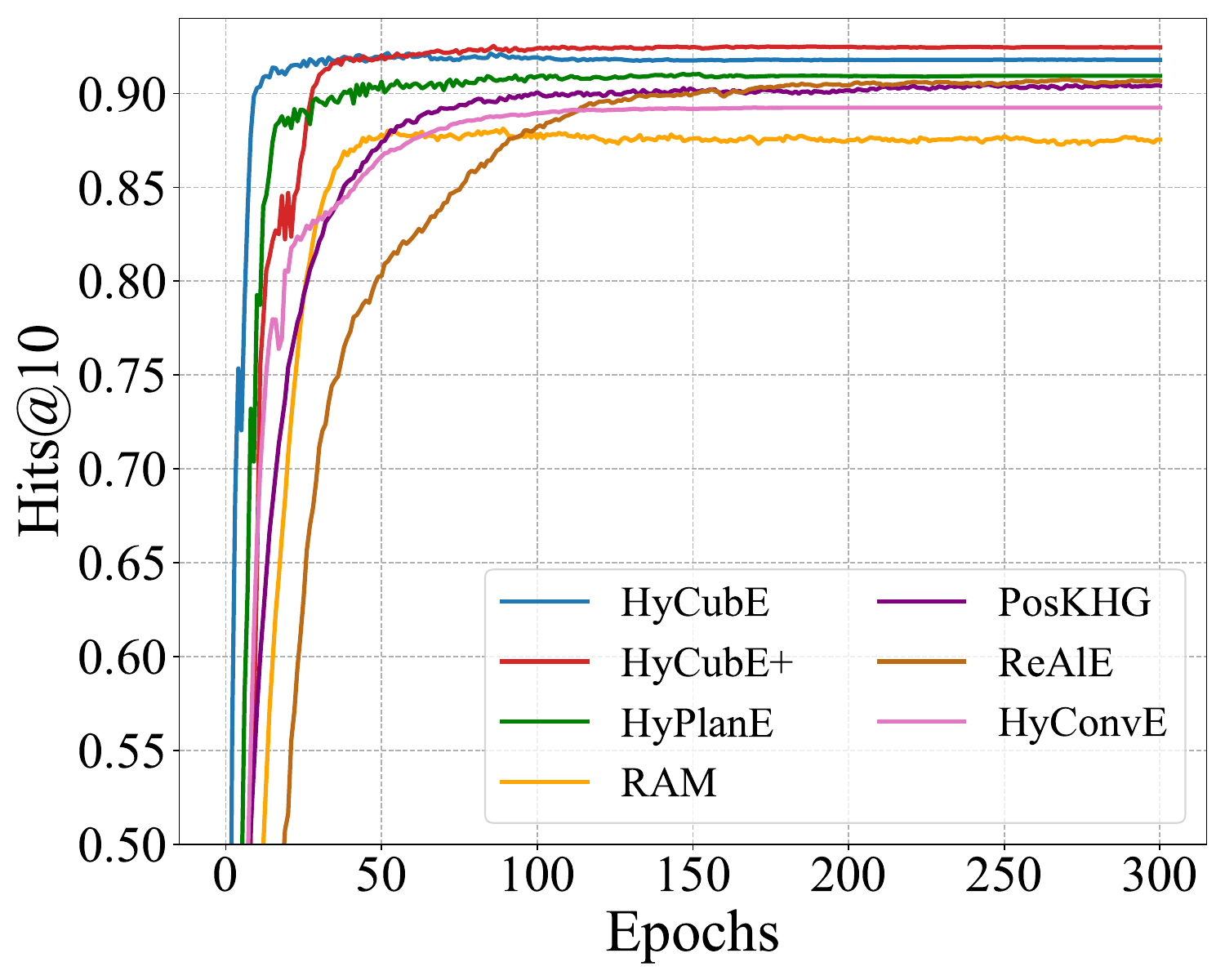}}
	\hspace{-10px}
	\subfigure[WikiPeople-3]{
		\includegraphics[width=0.199\textwidth, height=0.159\textwidth]{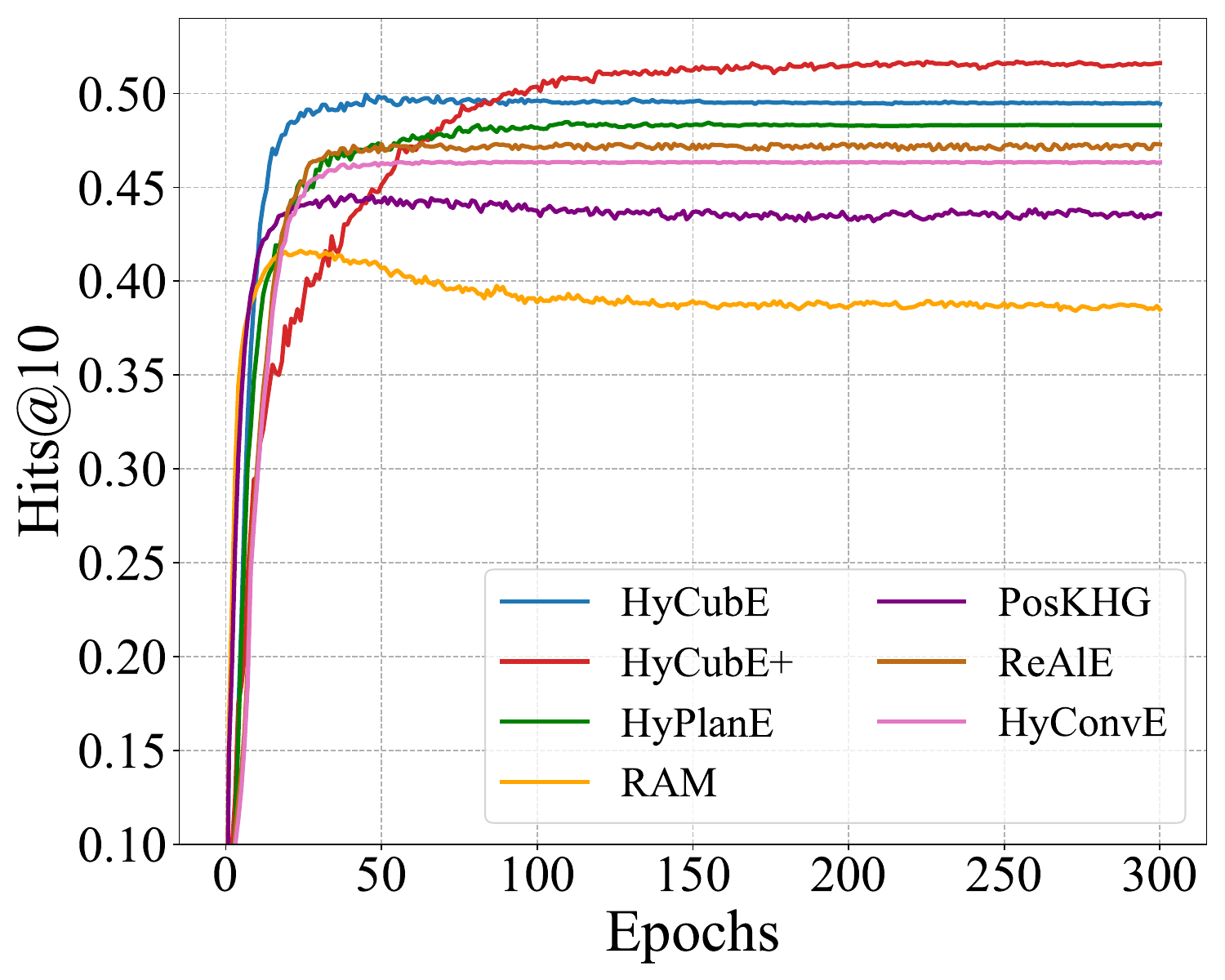}}
	\hspace{-10px}
	\subfigure[JF17K-4]{
		\includegraphics[width=0.199\textwidth, height=0.159\textwidth]{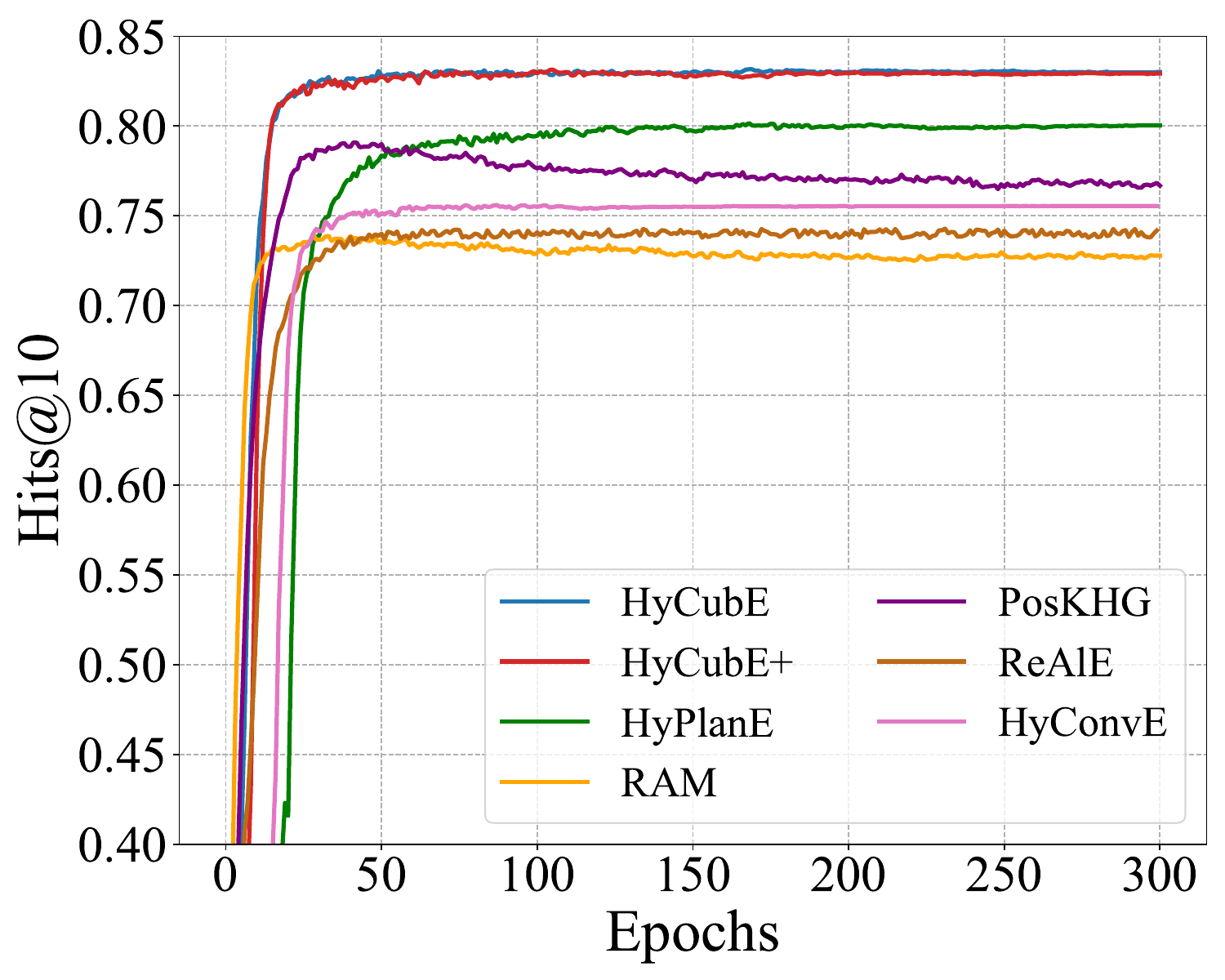}}
	\caption{The model training convergence curves (Hits@10 metric) on knowledge hypergraph datasets.}
 \label{curve:ModelTraining_H10}
\end{figure*}

\subsection{Model Efficiency Comparison}
An outstanding knowledge hypergraph embedding model must achieve a trade-off between effectiveness and efficiency, which is crucial for practical and large-scale dataset applications and generalizations. The experiments in this subsection include model efficiency metrics evaluation and model training convergence curves. All experimental results are obtained locally.

\subsubsection{Model Efficiency Results}
The efficiency evaluation metric of previous binary relational knowledge graph embedding methods is usually the number of parameters~\cite{HittER}~\cite{SAttLE}~\cite{JointE}, which is an indirect metric. However, some works have shown that indirect metrics cannot accurately evaluate the efficiency of a model~\cite{ShuffleNetV2}~\cite{RepVGG}~\cite{CSPNet}. Additionally, knowledge hypergraph embedding models have different types of methods (translation-based, semantic matching, and neural network methods), and accurately calculating the indirect metrics for each type of method is highly challenging. Therefore, it is more reasonable to use direct metrics (i.e., memory and time usage) to evaluate model efficiency. It should be noted that although the direct metric will be closer to the real-world application scenario, the direct metric is also slightly affected by some of the hyperparameters of the model (e.g., batch size). To ensure that the efficiency evaluation experiments are convincing and fair, we use both indirect metrics (the number of parameters) and direct metrics (memory and time usage) to evaluate model efficiency. Besides, the time usage contains the training, validation, and testing time of the knowledge hypergraph embedding models. The time usage refers to the time required for each epoch iteration of the knowledge hypergraph embedding models, calculated from the average of 10 epochs iteration time.

The experimental results of the \textit{mixed arity} and \textit{fixed arity} model efficiency comparison are shown in TABLE~\ref{result_mixed_eff} and TABLE~\ref{result_fixed_eff}. As mentioned earlier, HyCubE focuses on improving model efficiency while pursuing effectiveness. Compared with state-of-the-art baseline methods on all datasets, the HyCubE model speed is 6.12x faster on average, GPU memory is 52.67\% less on average, and the number of parameters is 85.21\% fewer on average; the HyCubE+ model speed is 5.32x faster on average, GPU memory is 50.63\% less on average, and the number of parameters is 53.03\% fewer on average. Commendably, HyCubE (HyCubE+) consistently reaches optimal effectiveness while ensuring model training efficiency, proving that HyCubE (HyCubE+) has an excellent knowledge hypergraph embedding model architecture.

\subsubsection{Model Training Convergence Curves}
The training efficiency curves are plotted from 300 epochs iteration of knowledge hypergraph models, and the results are shown in Fig.~\ref{curve:ModelTraining_MRR}, Fig.~\ref{curve:ModelTraining_H1}, and Fig.~\ref{curve:ModelTraining_H10}. From the experimental results, it can be seen that HyCubE (HyCubE+) has the fastest training convergence while achieving the best performance. Our proposed method is based on a novel 3D circular convolutional embedding architecture, which can efficiently process $n$-ary knowledge tuples of different arities with fewer parameters by adaptively adjusting the model structure parameters to accomplish end-to-end knowledge hypergraph embedding. This experimental result demonstrates that our proposed novel 3D circular convolutional embedding architecture attains an excellent trade-off between effectiveness and efficiency.

\subsubsection{Efficiency Comparison of 2D Variant}
To further validate the efficiency of our proposed model, we purposely perform a direct comparison with the 2D variant model HyPlanE. Compared with HyPlanE on average, HyCubE and HyCubE+ are 1.61x and 1.54x faster, 7.54\% and 4.27\% less GPU memory, and the number of model parameters is reduced by 89.90\% and 62.78\%, respectively. In addition, both HyCubE and HyCubE+ can converge faster and better than the model training curve of HyPlanE. Since we still retain the knowledge hypergraph 1-N multilinear scoring way for HyPlanE, it also has some efficiency advantages over state-of-the-art baselines. This experimental result can further demonstrate the superiority of our proposed new 3D circular convolutional embedding architecture, which can efficiently accomplish knowledge hypergraph embedding with fewer parameters, achieving a satisfactory trade-off between model effectiveness and efficiency.

\begin{table*}[ht]
  \begin{center}
    \caption{Results of Ablation Study on Mixed Arity Knowledge Hypergraph Datasets}
    \vspace{-10px}
    \label{mixed_ablation_study}
    \centering
    \resizebox{\linewidth}{!}{
    \begin{tabular}{c|cccc|cccc|cccc}
        \toprule
        ~ \multirow{2}{*}{\textbf{Model}} & \multicolumn{4}{c|}{\textbf{JF17K}} & \multicolumn{4}{c|}{\textbf{WikiPeople}} & \multicolumn{4}{c}{\textbf{FB-\footnotesize AUTO}}\\
        \cline{2-13}
        \rule{0pt}{10pt} & \textbf{MRR} & \textbf{Hits@1} & \textbf{Hits@3} & \textbf{Hits@10} & \textbf{MRR}   & \textbf{Hits@1} & \textbf{Hits@3} & \textbf{Hits@10} & \textbf{MRR}   & \textbf{Hits@1} & \textbf{Hits@3} & \textbf{Hits@10} \\
        \hline
        \hline
        \textbf{HyCubE} & \textbf{0.584} & \textbf{0.508} & \textbf{0.616} & \textbf{0.730} & \textbf{0.448} & \textbf{0.368} & \textbf{0.490} & \textbf{0.592} & \textbf{0.881} & \textbf{0.860} & \textbf{0.894} & \textbf{0.918} \\
        \textbf{w/o Alternate} & {0.579} & {0.506} & {0.612} & {0.723} & {0.447} & {0.367} & {0.487} & {0.591} & {0.875} & {0.853} & {0.888} & {0.911} \\
        \textbf{w/o Circular} & {0.583} & {0.507} & {0.614} & {0.727} & {0.437} & {0.356} & {0.479} & {0.584} & {0.877} & {0.855} & {0.890} & {0.911} \\
        \hline
        \textbf{HyCubE+} & \textbf{0.582} & \textbf{0.511} & \textbf{0.611} & \textbf{0.720} & \textbf{0.433} & \textbf{0.347} & \textbf{0.478} & \textbf{0.591} & \textbf{0.891} & \textbf{0.872} & \textbf{0.901} & \textbf{0.923} \\
        \textbf{w/o Alternate} & {0.577} & {0.505} & {0.607} & {0.719} & {0.432} & {0.345} & {0.477} & {0.590} & {0.887} & {0.869} & {0.897} & {0.919} \\
        \textbf{w/o Circular} & {0.532} & {0.451} & {0.568} & {0.690} & {0.432} & {0.346} & {0.477} & {0.589} & {0.888} & {0.871} & {0.898} & {0.919} \\
        \hline
        \textbf{HyPlanE} & {0.569} & {0.496} & {0.600} & {0.708} & {0.402} & {0.323} & {0.443} & {0.549} & {0.866} & {0.843} & {0.880} & {0.909} \\
        \bottomrule
    \end{tabular}
    }
  \end{center}
  \thanks{The best results are in boldface.}
\end{table*} 

\begin{table*}[ht]
  \begin{center}
    \caption{Results of Ablation Study on Fixed Arity Knowledge Hypergraph Datasets}
    \vspace{-10px}    
    \label{fixed_ablation_study}
    \centering
    \resizebox{\linewidth}{!}{
    \begin{tabular}{c|ccc|ccc|ccc|ccc}
        \toprule
        ~ \multirow{2}{*}{\textbf{Model}} & \multicolumn{3}{c|}{\textbf{JF17K-3}} & \multicolumn{3}{c|}{\textbf{JF17K-4}} & \multicolumn{3}{c|}{\textbf{WikiPeople-3}} & \multicolumn{3}{c}{\textbf{WikiPeople-4}} \\
        \cline{2-13}
        \rule{0pt}{10pt} & \textbf{MRR} & \textbf{Hits@1} & \textbf{Hits@10} & \textbf{MRR} & \textbf{Hits@1} & \textbf{Hits@10} & \textbf{MRR} & \textbf{Hits@1} & \textbf{Hits@10} & \textbf{MRR} & \textbf{Hits@1} & \textbf{Hits@10} \\
        \hline
        \hline
        \textbf{HyCubE} & \textbf{0.599} & \textbf{0.534} & \textbf{0.723} & \textbf{0.793} & \textbf{0.742} & \textbf{0.887} & \textbf{0.336} & \textbf{0.256} & \textbf{0.499} & \textbf{0.367} & \textbf{0.258} & \textbf{0.578} \\
        \textbf{w/o Alternate} & {0.596} & {0.532} & {0.718} & {0.789} & {0.738} & {0.884} & {0.332} & {0.252} & {0.492} & {0.363} & {0.248} & {0.577} \\
        \textbf{w/o Circular} & {0.590} & {0.525} & {0.714} & {0.790} & {0.738} & {0.886} & {0.330} & {0.252} & {0.490} & {0.361} & {0.249} & {0.576} \\
        \hline
        \textbf{HyCubE+} & \textbf{0.603} & \textbf{0.537} & \textbf{0.728} & \textbf{0.794} & \textbf{0.743} & \textbf{0.886} & \textbf{0.345} & \textbf{0.260} & \textbf{0.515} & \textbf{0.383}& \textbf{0.269} & \textbf{0.602} \\
        \textbf{w/o Alternate} & {0.601} & {0.534} & {0.727} & {0.791} & {0.740} & {0.882} & {0.343} & {0.255} & {0.514} & {0.377} & {0.260} & {0.597} \\
        \textbf{w/o Circular} & {0.601} & {0.535} & {0.726} & {0.792} & {0.741} & {0.885} & {0.340} & {0.254} & {0.512} & {0.373} & {0.253} & {0.597} \\
        \hline
        \textbf{HyPlanE} & {0.574} & {0.510} & {0.700} & {0.757} & {0.699} & {0.862} & {0.318} & {0.237} & {0.479} & {0.324}& {0.210} & {0.559} \\
        \bottomrule
    \end{tabular}
    }
  \end{center}
  \thanks{The best results are in boldface.}
\end{table*}

\begin{figure*}[h]
	\centering
	\subfigure[JF17K]{
		\includegraphics[width=0.199\textwidth, height=0.149\textwidth]{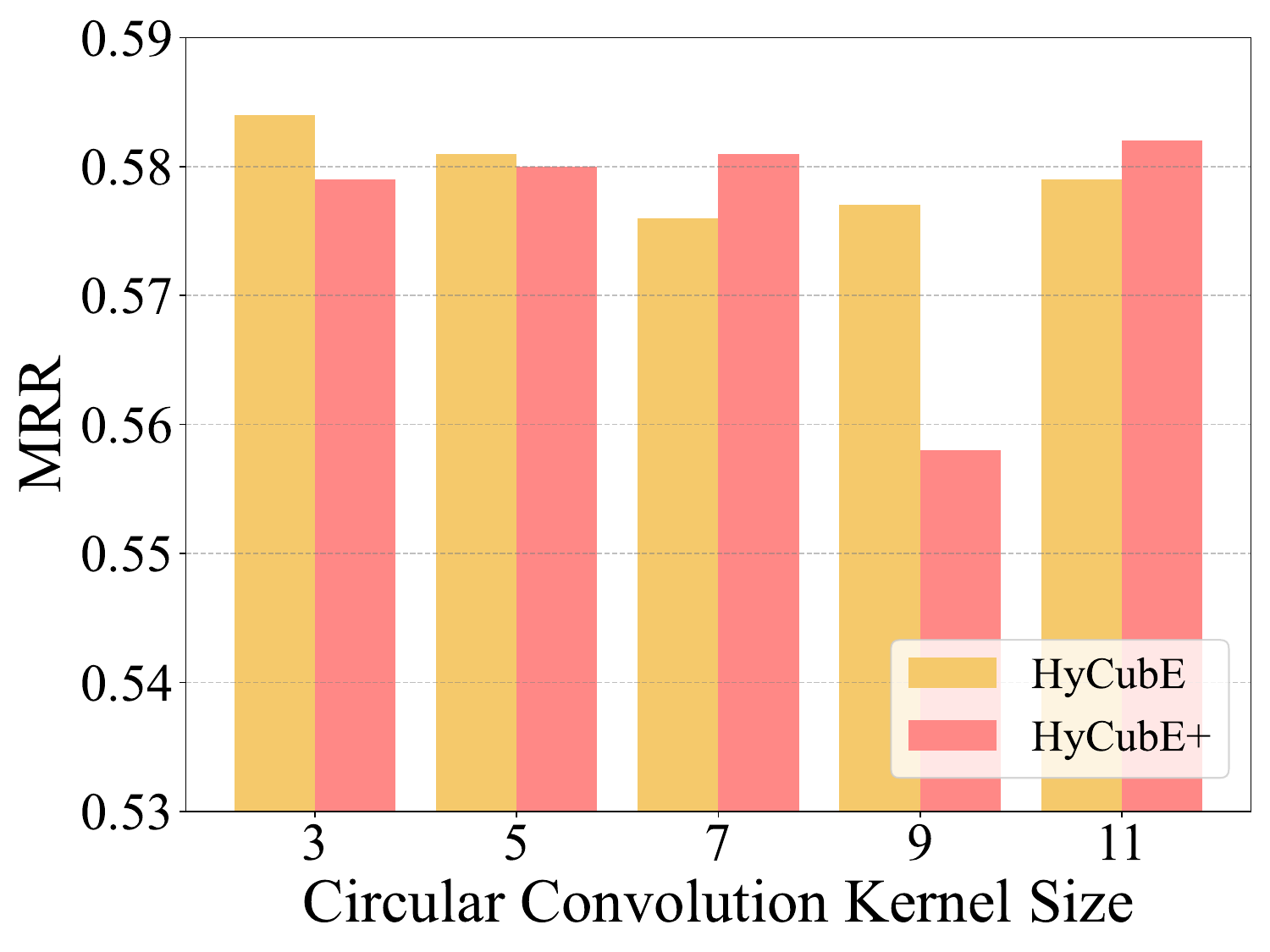}}
 	\hspace{-10px}
	\subfigure[WikiPeople]{
		\includegraphics[width=0.199\textwidth, height=0.149\textwidth]{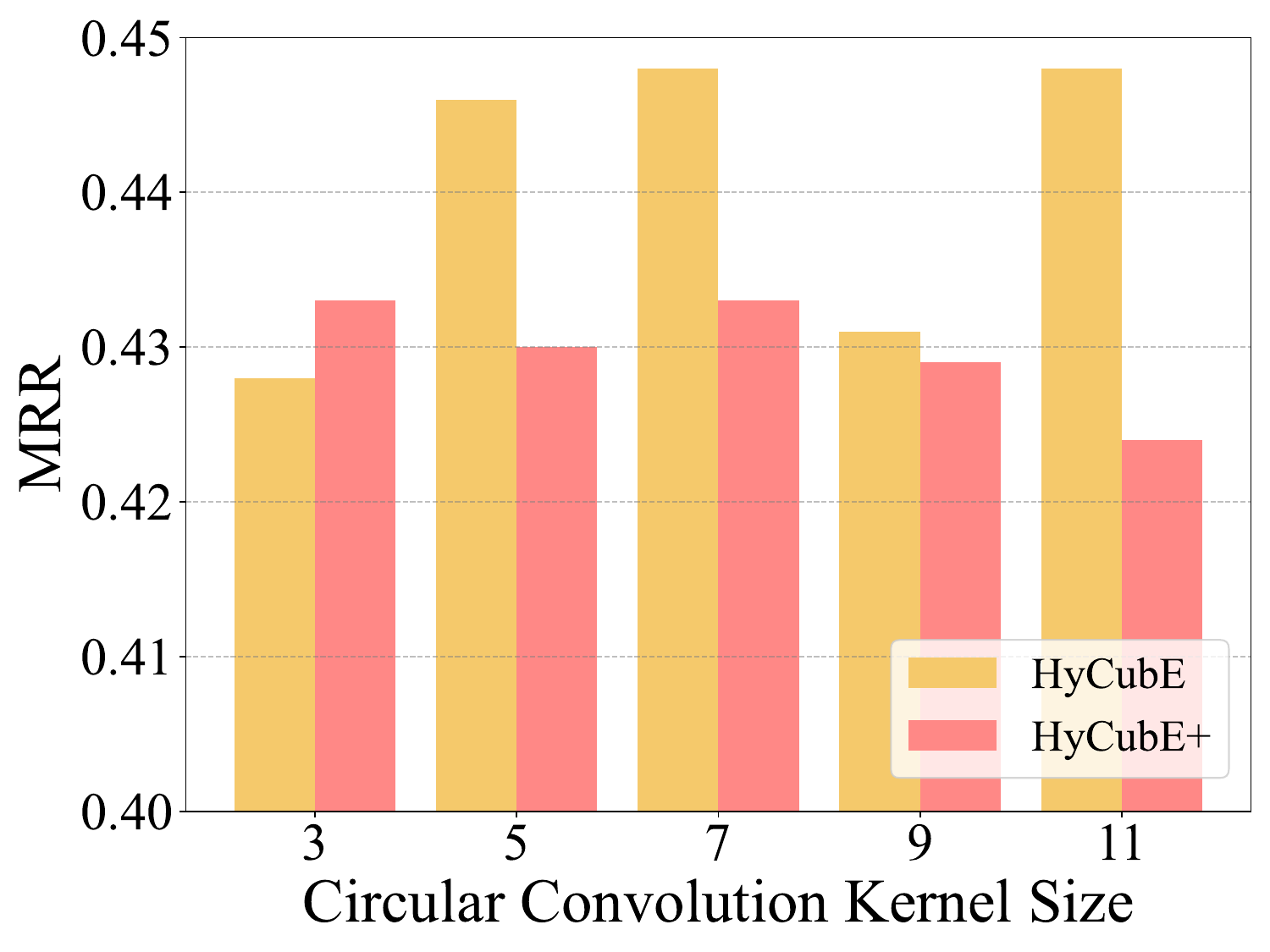}}
 	\hspace{-10px}
	\subfigure[FB-AUTO]{
		\includegraphics[width=0.199\textwidth, height=0.149\textwidth]{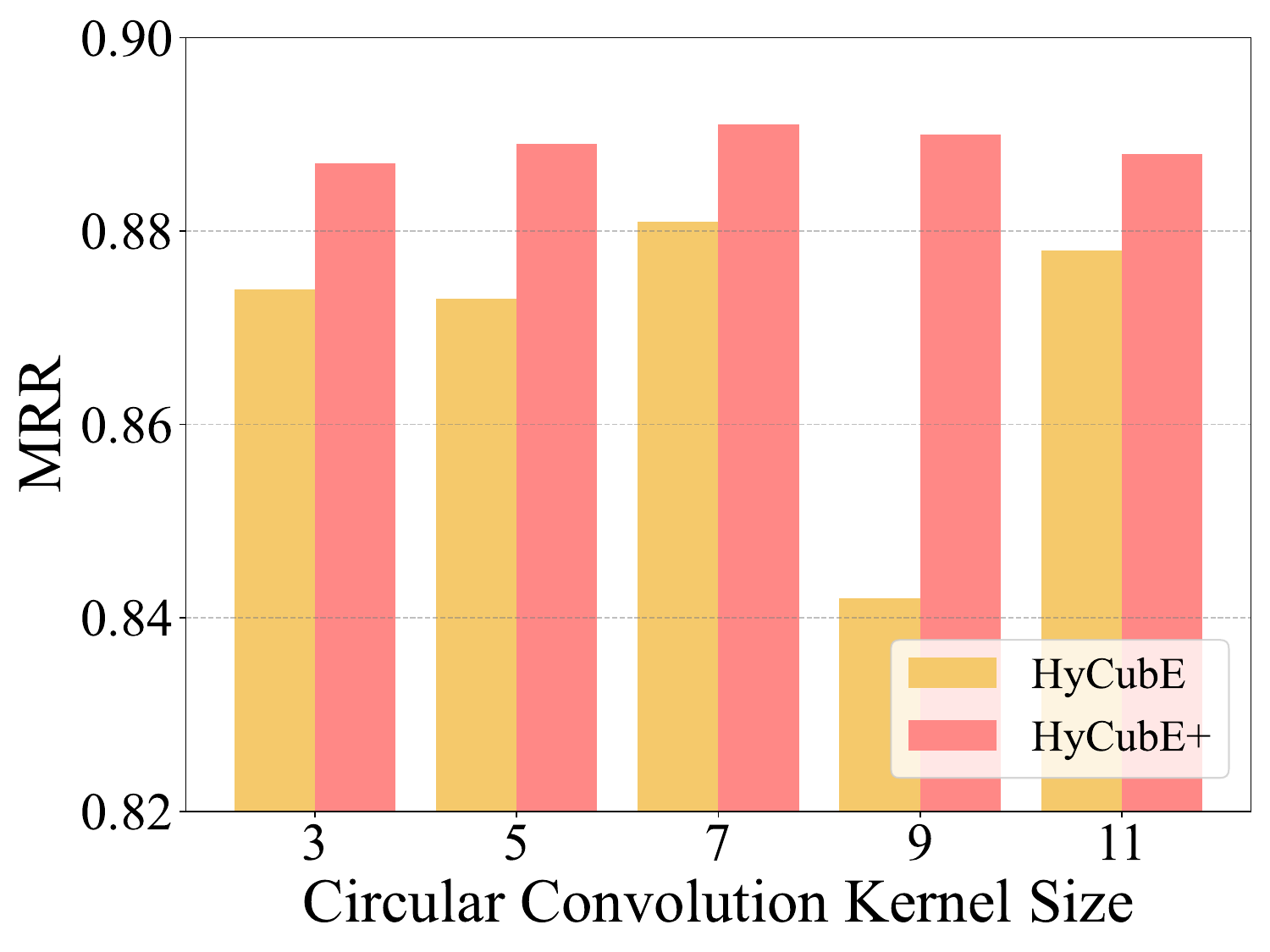}}
 	\hspace{-10px}
	\subfigure[WikiPeople-3]{
		\includegraphics[width=0.199\textwidth, height=0.149\textwidth]{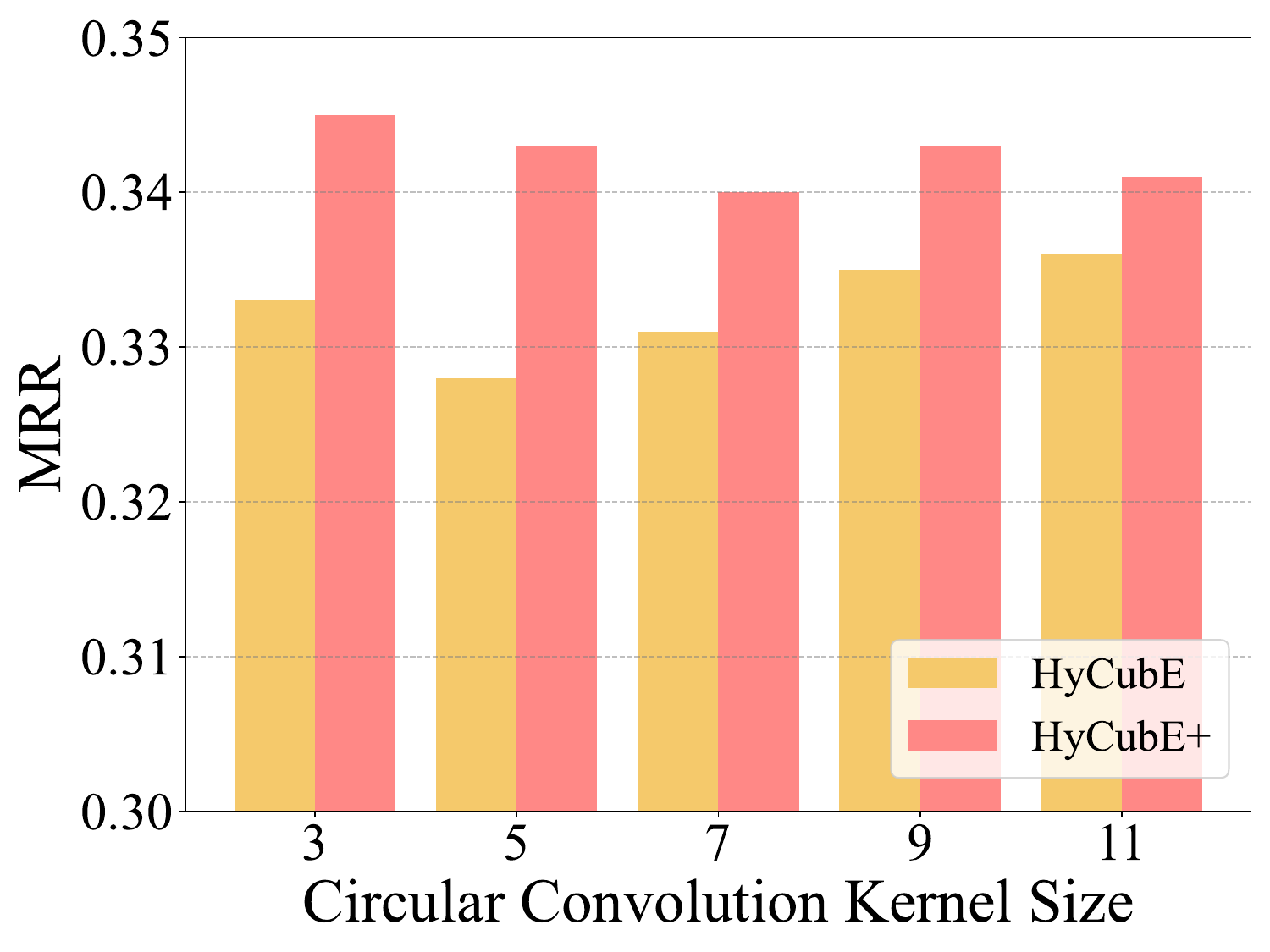}}
 	\hspace{-10px}
	\subfigure[JF17K-4]{
		\includegraphics[width=0.199\textwidth, height=0.149\textwidth]{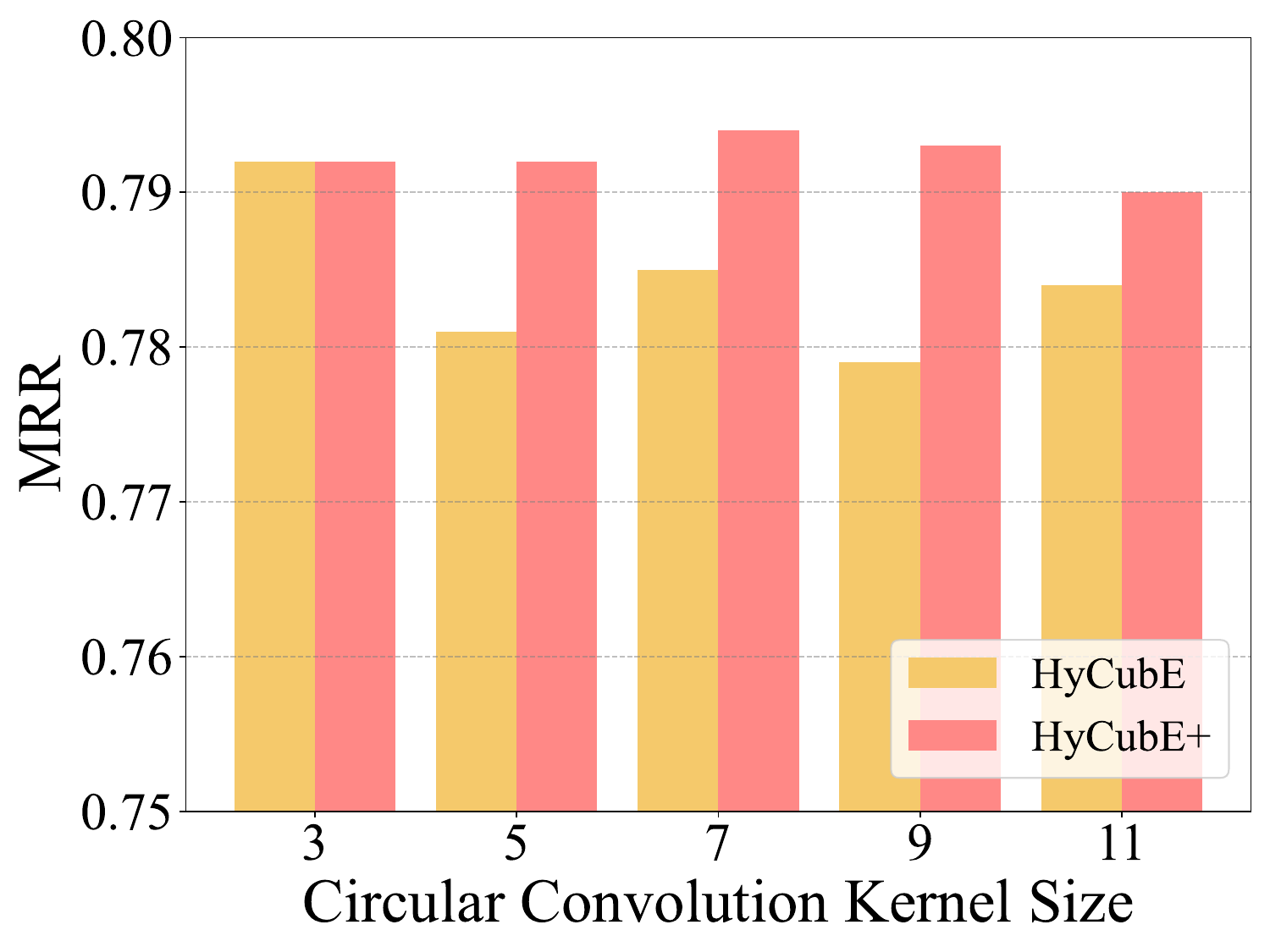}}
	\caption{The 3D circular convolution kernel size parameter analysis on knowledge hypergraph datasets.}
 \label{curve:ConvolutionSize}
\end{figure*}

\begin{figure*}[h]
	\centering
	\subfigure[JF17K]{
		\includegraphics[width=0.199\textwidth, height=0.149\textwidth]{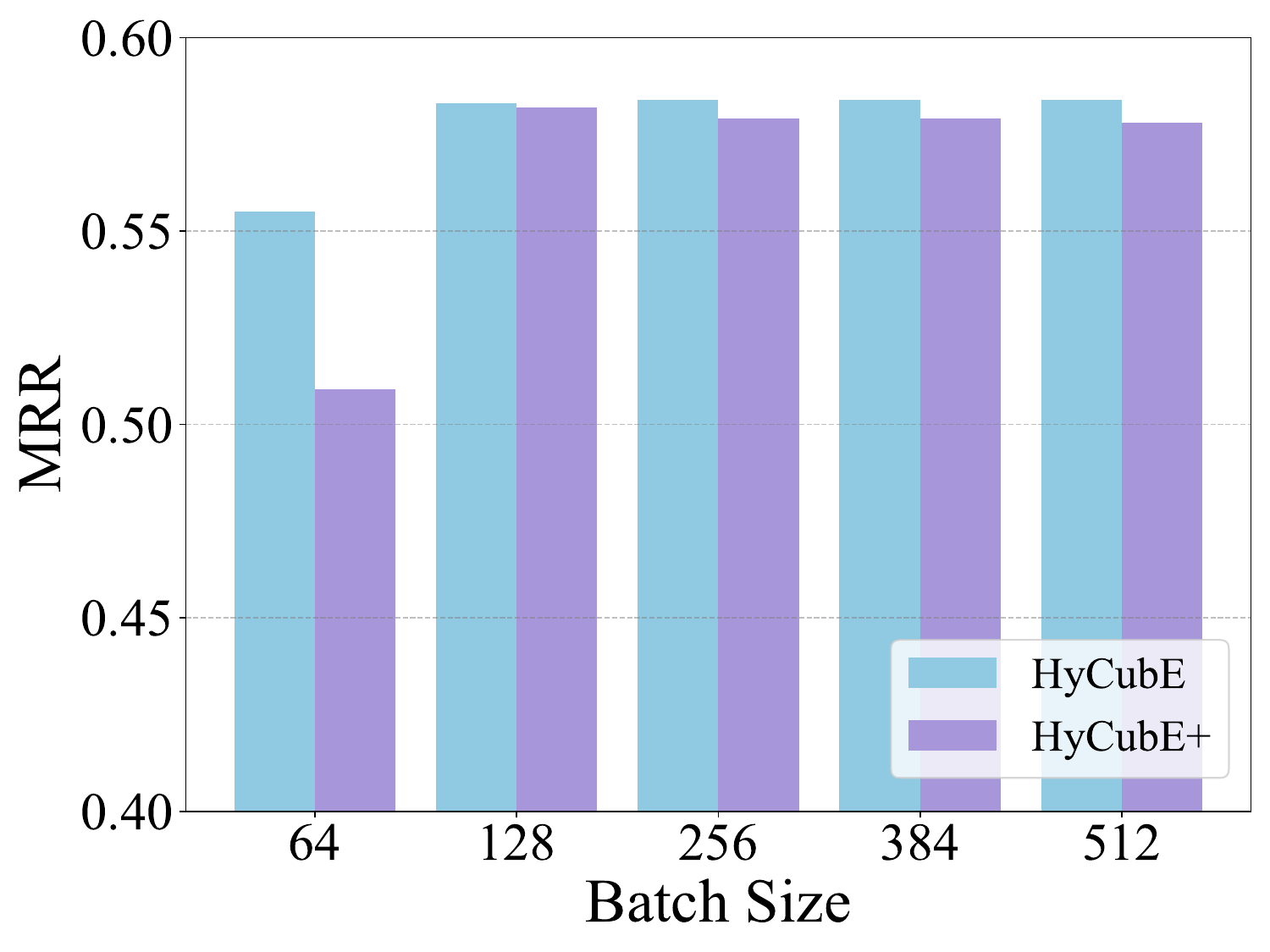}}
 	\hspace{-10px}
	\subfigure[WikiPeople]{
		\includegraphics[width=0.199\textwidth, height=0.149\textwidth]{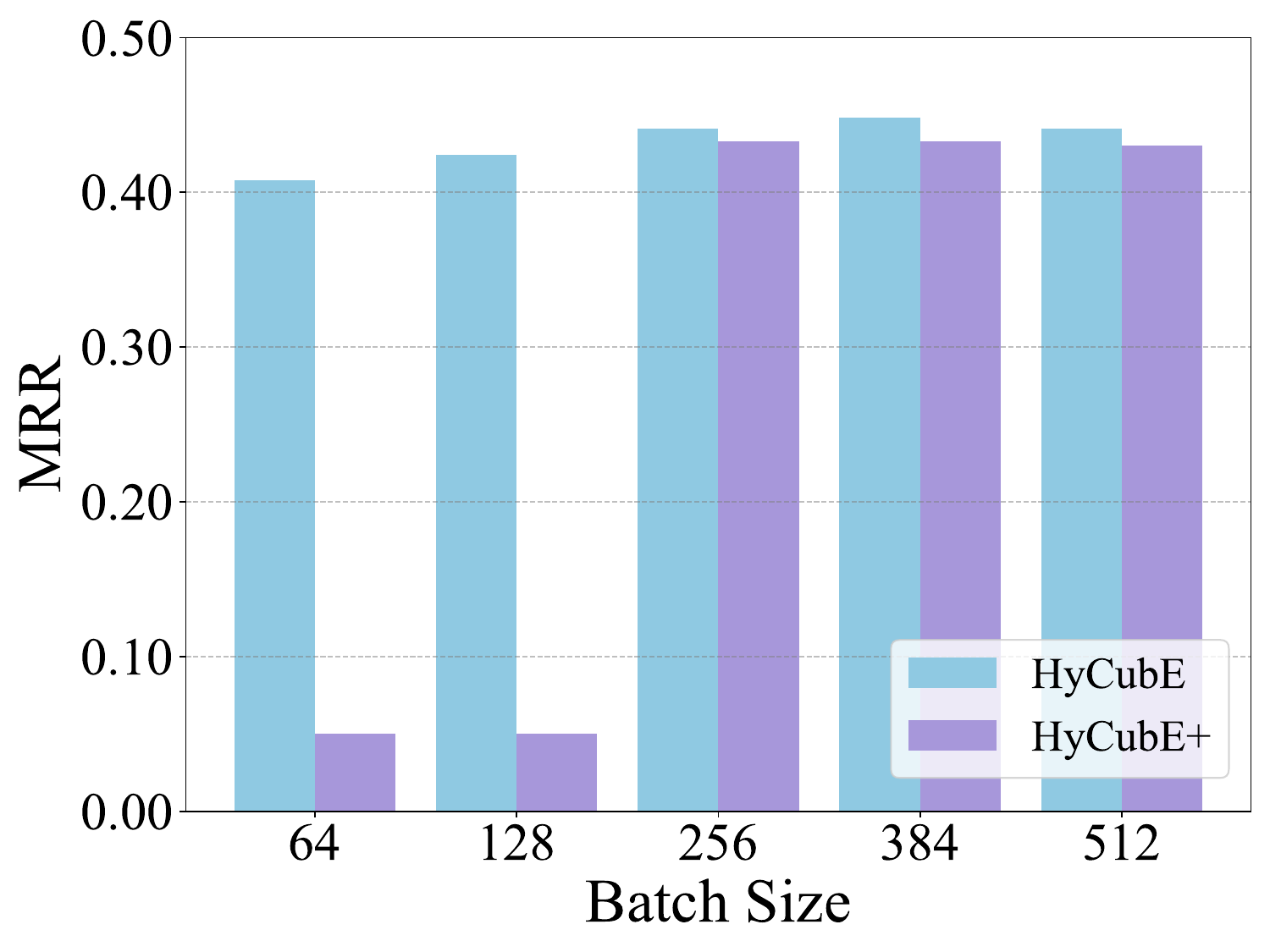}}
 	\hspace{-10px}
	\subfigure[FB-AUTO]{
		\includegraphics[width=0.199\textwidth, height=0.149\textwidth]{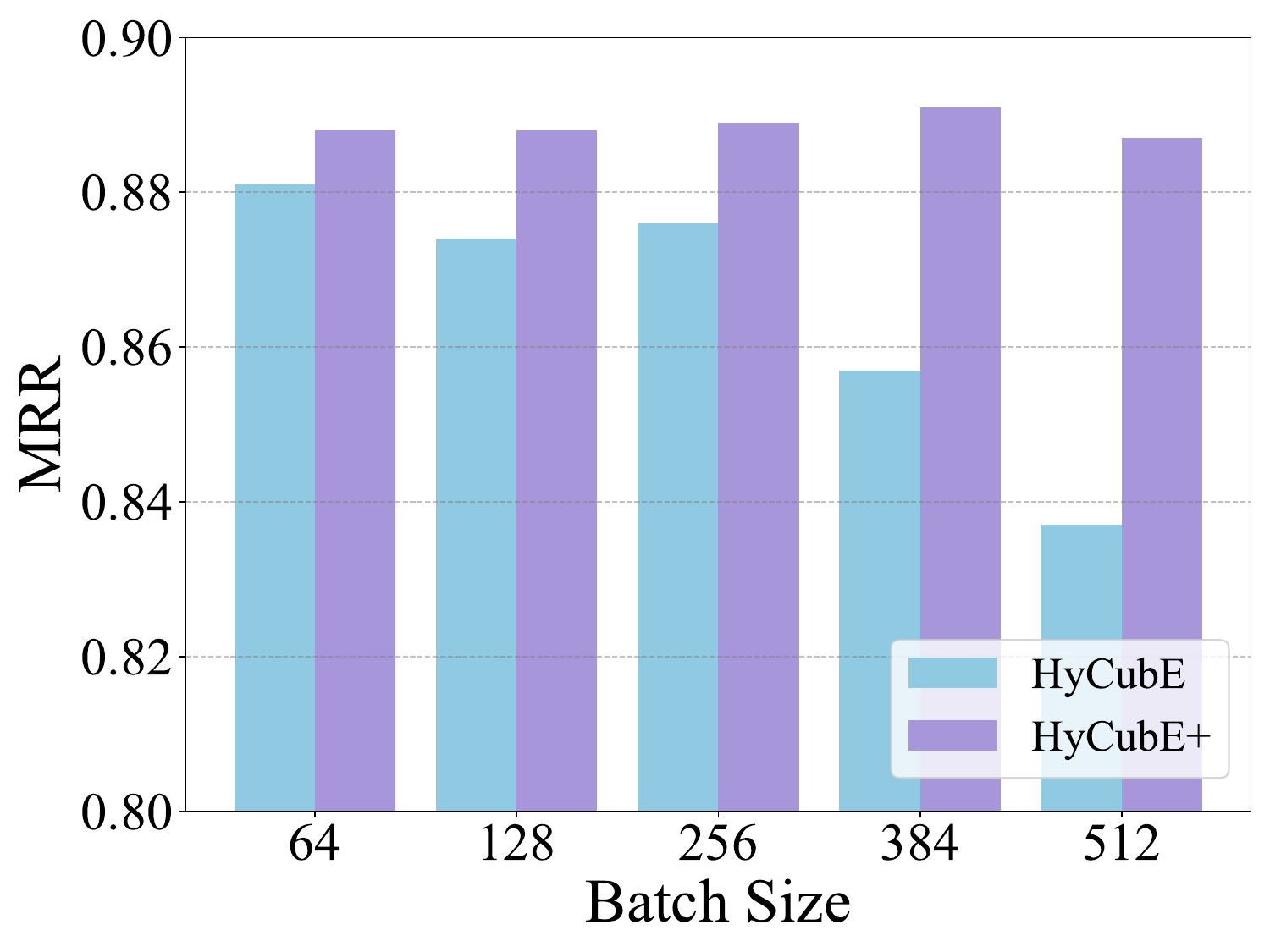}}
 	\hspace{-10px}
	\subfigure[WikiPeople-3]{
		\includegraphics[width=0.199\textwidth, height=0.149\textwidth]{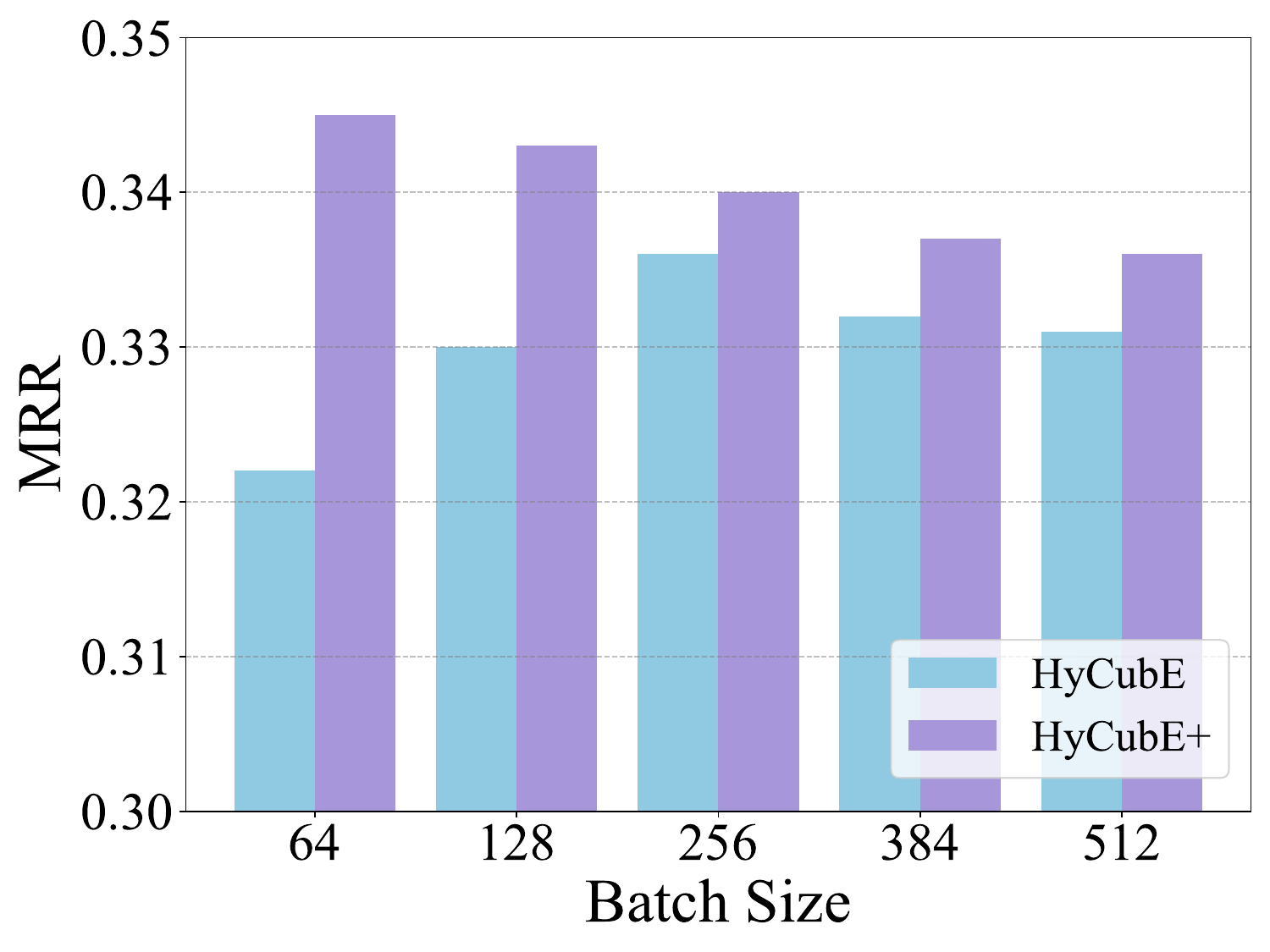}}
 	\hspace{-10px}
	\subfigure[JF17K-4]{
		\includegraphics[width=0.199\textwidth, height=0.149\textwidth]{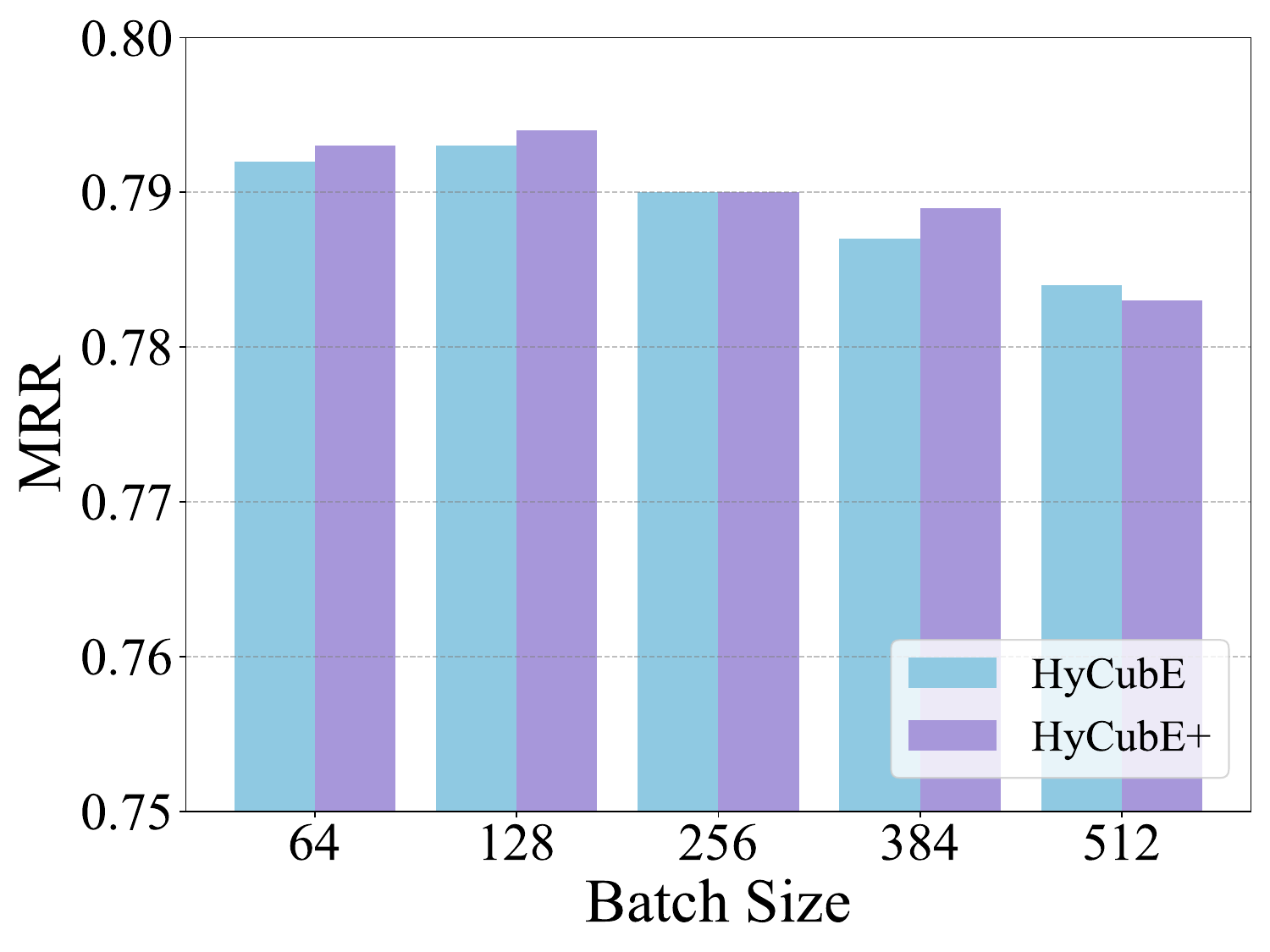}}
	\caption{The batch size parameter analysis on knowledge hypergraph datasets.}
 \label{curve:BatchSize}
\end{figure*}

\subsection{Ablation Study}
The ablation studies were conducted on \textit{mixed arity} and \textit{fixed arity} knowledge hypergraph datasets with hyperparameters consistent with the knowledge hypergraph link prediction task. For the core part of the HyCubE and HyCubE+, we conduct two ablation study experiments: removing the alternate stack strategy (w/o Alternate) and removing the circular padding of the 3D circular convolution (w/o Circular). Because the mask mechanism of the \textit{alternate mask stack} strategy couples the 1-N multilinear scoring function of the model, the mask mechanism is maintained in the ablation study. Furthermore, the 2D variant of HyCubE, HyPlanE, is used as a comparative baseline for ablation study experiments to evaluate the superiority of 3D convolutional embedding further. To control the convolutional feature size and ensure fairness, we set HyCubE (w/o Circular) and HyCubE+ (w/o Circular) to use zero-value padding, and HyCubE (w/o Alternate) and HyCubE+ (w/o Alternate) to use the traditional standard stack.

The experimental results of the \textit{mixed arity} and \textit{fixed arity} knowledge hypergraph ablation studies are shown in TABLE~\ref{mixed_ablation_study} and TABLE~\ref{fixed_ablation_study}, respectively.
HyCubE (HyCubE+) decreases performance on all evaluation metrics when removing any core components for \textit{mixed arity} and \textit{fixed arity} knowledge hypergraph link prediction task. Specifically, on the \textit{mixed arity} knowledge hypergraph datasets, the performance of HyCubE (w/o Alternate) and HyCubE (w/o Circular) decreases by an average of 0.59\% and 1.08\%, respectively; the performance of HyCubE+ (w/o Alternate) and HyCubE+ (w/o Circular) reduces by an average of 0.48\% and 3.08\%, respectively. On the \textit{fixed arity} knowledge hypergraph datasets, the performance of HyCubE (w/o Alternate) and HyCubE (w/o Circular) decreases by an average of 0.96\% and 1.35\%, respectively; the performance of HyCubE+ (w/o Alternate) and HyCubE+ (w/o Circular) reduces by an average of 0.81\% and 1.20\%, respectively. Consequently, the circular padding of 3D circular convolution is superior to the alternate stack strategy in enhancing the expressive power of knowledge hypergraph embedding. In addition, the experimental results of the \textit{mixed arity} and \textit{fixed arity} knowledge hypergraph ablation studies show that HyCubE has an average performance improvement of 6.11\% over HyPlanE, and HyCubE+ has an average performance improvement of 6.97\% over HyPlanE. Obviously, the 3D convolutional embedding architecture plays the most significant role in improving the expressiveness of the knowledge hypergraph embedding model.

\subsection{Hyperparameter Analysis}
The experiments of hyperparameter analysis were conducted on \textit{mixed arity} and \textit{fixed arity} knowledge hypergraph datasets with all other hyperparameters consistent with the knowledge hypergraph link prediction task. It is well known that the convolution kernel size is an essential hyperparameter for convolutional neural network models, and the batch size impacts both time and memory usage during model training. Therefore, two core hyperparameters, \textit{Convolution Kernel Size} and \textit{Batch Size}, are used for sensitivity analysis to evaluate the stability of the knowledge hypergraph embedding model. The experimental results of 3D circular convolution kernel size parameter analysis are shown in Fig.~\ref{curve:ConvolutionSize}, and the experimental results of batch size parameter analysis are shown in Fig.~\ref{curve:BatchSize}.

It can be seen from the experimental results that 3D circular convolution kernel size and batch size have an impact on the performance of knowledge hypergraph embedding. However, the variations of all evaluation metrics are moderate, proving that our proposed model is robust and stable. Moreover, in knowledge hypergraph datasets (FB-AUTO, WikiPeople-3, and JF17K-4) with less $n$-ary implicit semantic information, 3D residual-enhanced HyCubE+ usually performs better and is more stable than HyCubE. Importantly, in the batch size parameter analysis experiments, when the batch size is 64 and 128, the loss function of the HyCubE+ model does not converge during the training process. For the aesthetics of the figures in this paper, we set the HyCubE+ experimental results to 0.05. In particular, we perform batch size parameter analysis experiments on the WikiPeople dataset using some of the state-of-the-art baseline models (RAM, PosKHG, and HyConvE) and find that all of these baselines are prone to loss function non-convergence on the WikiPeople dataset.

As mentioned above, the $n$-ary semantic information inherent in the WikiPeople dataset is the most complex, with the highest number of relation-arity and relation-specific. Hyperparameter analysis experiments show that the inherent $n$-ary semantic information in the benchmark dataset is too complex, which brings many challenges to the hyperparameter optimization of knowledge hypergraph embedding models. Consequently, HyCubE (HyCubE+) has the most significant performance gains on the WikiPeople dataset, which further shows the superiority of our proposed model architecture.

\section{Conclusion}
In this paper, we propose a performance-enhanced 3D circular convolutional embedding model, HyCubE, which designs a novel \textit{3D circular convolutional neural network} and introduces the \textit{alternate mask stack} strategy to achieve effective knowledge hypergraph embedding. Our proposed model enables a better trade-off between effectiveness and efficiency by adaptively adjusting the 3D circular convolutional layer structure to construct end-to-end efficient knowledge hypergraph embedding with fewer parameters. Additionally, HyCubE+ with the 3D residual module is proposed to effectively alleviate the gradient vanishing problem that occurs easily in knowledge hypergraph datasets with less $n$-ary semantic information. Extensive experimental studies have shown that our proposed method consistently outperforms state-of-the-art baseline models in terms of both knowledge hypergraph embedding effectiveness and efficiency.

\bibliographystyle{IEEEtran}
\bibliography{TKDEHyCubE}

% Generated by IEEEtran.bst, version: 1.14 (2015/08/26)
\begin{thebibliography}{10}
\providecommand{\url}[1]{#1}
\csname url@samestyle\endcsname
\providecommand{\newblock}{\relax}
\providecommand{\bibinfo}[2]{#2}
\providecommand{\BIBentrySTDinterwordspacing}{\spaceskip=0pt\relax}
\providecommand{\BIBentryALTinterwordstretchfactor}{4}
\providecommand{\BIBentryALTinterwordspacing}{\spaceskip=\fontdimen2\font plus
\BIBentryALTinterwordstretchfactor\fontdimen3\font minus \fontdimen4\font\relax}
\providecommand{\BIBforeignlanguage}[2]{{%
\expandafter\ifx\csname l@#1\endcsname\relax
\typeout{** WARNING: IEEEtran.bst: No hyphenation pattern has been}%
\typeout{** loaded for the language `#1'. Using the pattern for}%
\typeout{** the default language instead.}%
\else
\language=\csname l@#1\endcsname
\fi
#2}}
\providecommand{\BIBdecl}{\relax}
\BIBdecl

\bibitem{m-TransH}
J.~Wen, J.~Li, Y.~Mao, S.~Chen, and R.~Zhang, ``On the representation and embedding of knowledge bases beyond binary relations,'' in \emph{Proceedings of the Twenty-Fifth International Joint Conference on Artificial Intelligence}, ser. IJCAI'16, 2016, pp. 1300--1307.

\bibitem{HypE-HSimplE}
B.~Fatemi, P.~Taslakian, D.~Vazquez, and D.~Poole, ``Knowledge hypergraphs: prediction beyond binary relations,'' in \emph{Proceedings of the Twenty-Ninth International Conference on International Joint Conferences on Artificial Intelligence}, ser. IJCAI'21, 2021, pp. 2191--2197.

\bibitem{ReAlE}
------, ``Knowledge hypergraph embedding meets relational algebra,'' \emph{Journal of Machine Learning Research}, vol.~24, no. 105, pp. 1--34, 2023.

\bibitem{DASFAA}
Z.~Li, ``Knowledge hypergraph reasoning based on representation learning,'' in \emph{International Conference on Database Systems for Advanced Applications}, ser. DASFAA'23, 2023, pp. 743--747.

\bibitem{RAM}
Y.~Liu, Q.~Yao, and Y.~Li, ``Role-aware modeling for n-ary relational knowledge bases,'' in \emph{Proceedings of The Web Conference 2021}, ser. WWW'21, 2021, pp. 2660--2671.

\bibitem{PosKHG}
Z.~Chen, X.~Wang, C.~Wang, and Z.~Li, ``Poskhg: A position-aware knowledge hypergraph model for link prediction,'' \emph{Data Science and Engineering}, vol.~8, no.~2, pp. 135--145, 2023.

\bibitem{HyConvE}
C.~Wang, X.~Wang, Z.~Li, Z.~Chen, and J.~Li, ``Hyconve: A novel embedding model for knowledge hypergraph link prediction with convolutional neural networks,'' in \emph{Proceedings of the ACM Web Conference 2023}, ser. WWW'23, 2023, pp. 188--198.

\bibitem{HJE}
Z.~Li, C.~Wang, X.~Wang, Z.~Chen, and J.~Li, ``Hje: Joint convolutional representation learning for knowledge hypergraph completion,'' \emph{IEEE Transactions on Knowledge and Data Engineering}, vol.~36, no.~8, pp. 3879--3892, 2024.

\bibitem{KGTransformer}
W.~Zhang, Y.~Zhu, M.~Chen, Y.~Geng, Y.~Huang, Y.~Xu, W.~Song, and H.~Chen, ``Structure pretraining and prompt tuning for knowledge graph transfer,'' in \emph{Proceedings of the ACM Web Conference 2023}, ser. WWW'23, 2023, pp. 2581--2590.

\bibitem{kgformer}
X.~Liu, S.~Zhao, K.~Su, Y.~Cen, J.~Qiu, M.~Zhang, W.~Wu, Y.~Dong, and J.~Tang, ``Mask and reason: Pre-training knowledge graph transformers for complex logical queries,'' in \emph{Proceedings of the 28th ACM SIGKDD Conference on Knowledge Discovery and Data Mining}, ser. KDD'22, 2022, pp. 1120--1130.

\bibitem{RD-MPNN}
X.~Zhou, B.~Hui, I.~Zeira, H.~Wu, and L.~Tian, ``Dynamic relation learning for link prediction in knowledge hypergraphs,'' \emph{Applied Intelligence}, vol.~53, no.~22, pp. 26\,580--26\,591, 2023.

\bibitem{KGT5}
A.~Saxena, A.~Kochsiek, and R.~Gemulla, ``Sequence-to-sequence knowledge graph completion and question answering,'' in \emph{Proceedings of the 60th Annual Meeting of the Association for Computational Linguistics (Volume 1: Long Papers)}, ser. ACL'22, 2022, pp. 2814--2828.

\bibitem{HittER}
S.~Chen, X.~Liu, J.~Gao, J.~Jiao, R.~Zhang, and Y.~Ji, ``Hitter: Hierarchical transformers for knowledge graph embeddings,'' in \emph{Proceedings of the 2021 Conference on Empirical Methods in Natural Language Processing}, ser. EMNLP'21, 2021, pp. 10\,395--10\,407.

\bibitem{SAttLE}
P.~Baghershahi, R.~Hosseini, and H.~Moradi, ``Self-attention presents low-dimensional knowledge graph embeddings for link prediction,'' \emph{Knowledge-Based Systems}, vol. 260, pp. 110--124, 2023.

\bibitem{ConvE}
T.~Dettmers, P.~Minervini, P.~Stenetorp, and S.~Riedel, ``Convolutional 2d knowledge graph embeddings,'' in \emph{Proceedings of the Thirty-Second AAAI Conference on Artificial Intelligence}, ser. AAAI'18, 2018, pp. 1811--1818.

\bibitem{InteractE}
S.~Vashishth, S.~Sanyal, V.~Nitin, N.~Agrawal, and P.~Talukdar, ``Interacte: Improving convolution-based knowledge graph embeddings by increasing feature interactions,'' in \emph{Proceedings of the 34th AAAI Conference on Artificial Intelligence}, ser. AAAI'20, 2020, pp. 3009--3016.

\bibitem{Nature}
B.~M. Lake and M.~Baroni, ``Human-like systematic generalization through a meta-learning neural network,'' \emph{Nature}, vol. 623, no. 7985, pp. 115--121, 2023.

\bibitem{JointE}
Z.~Zhou, C.~Wang, Y.~Feng, and D.~Chen, ``Jointe: Jointly utilizing 1d and 2d convolution for knowledge graph embedding,'' \emph{Knowledge-Based Systems}, vol. 240, p. 108100, 2022.

\bibitem{RAE}
R.~Zhang, J.~Li, J.~Mei, and Y.~Mao, ``Scalable instance reconstruction in knowledge bases via relatedness affiliated embedding,'' in \emph{Proceedings of the 2018 World Wide Web Conference}, ser. WWW'18, 2018, pp. 1185--1194.

\bibitem{NaLP}
S.~Guan, X.~Jin, Y.~Wang, and X.~Cheng, ``Link prediction on n-ary relational data,'' in \emph{Proceedings of the 2019 World Wide Web Conference}, ser. WWW'19, 2019, pp. 583--593.

\bibitem{GETD}
Y.~Liu, Q.~Yao, and Y.~Li, ``Generalizing tensor decomposition for n-ary relational knowledge bases,'' in \emph{Proceedings of The Web Conference 2020}, ser. WWW'20, 2020, pp. 1104--1114.

\bibitem{HyperMLN}
Z.~Chen, X.~Wang, C.~Wang, and J.~Li, ``Explainable link prediction in knowledge hypergraphs,'' in \emph{Proceedings of the 31st ACM International Conference on Information \& Knowledge Management}, ser. CIKM'22, 2022, pp. 262--271.

\bibitem{tNaLP}
S.~Guan, X.~Jin, J.~Guo, Y.~Wang, and X.~Cheng, ``Link prediction on n-ary relational data based on relatedness evaluation,'' \emph{IEEE Transactions on Knowledge and Data Engineering}, vol.~35, no.~1, pp. 672--685, 2023.

\bibitem{AcrE}
F.~Ren, J.~Li, H.~Zhang, S.~Liu, B.~Li, R.~Ming, and Y.~Bai, ``Knowledge graph embedding with atrous convolution and residual learning,'' in \emph{Proceedings of the 28th International Conference on Computational Linguistics}, ser. COLING'20, 2020, pp. 1532--1543.

\bibitem{CircularConvolution}
T.-H. Wang, H.-J. Huang, J.-T. Lin, C.-W. Hu, K.-H. Zeng, and M.~Sun, ``Omnidirectional cnn for visual place recognition and navigation,'' in \emph{IEEE International Conference on Robotics and Automation}, ser. ICRA'18, 2018, pp. 2341--2348.

\bibitem{Dropout}
N.~Srivastava, G.~Hinton, A.~Krizhevsky, I.~Sutskever, and R.~Salakhutdinov, ``Dropout: a simple way to prevent neural networks from overfitting,'' \emph{The journal of machine learning research}, vol.~15, no.~1, pp. 1929--1958, 2014.

\bibitem{BatchNormalization}
S.~Ioffe and C.~Szegedy, ``Batch normalization: Accelerating deep network training by reducing internal covariate shift,'' in \emph{International conference on machine learning}, ser. ICML'15, 2015, pp. 448--456.

\bibitem{AdaGrad}
J.~Duchi, E.~Hazan, and Y.~Singer, ``Adaptive subgradient methods for online learning and stochastic optimization.'' \emph{Journal of machine learning research}, vol.~12, no.~7, 2011.

\bibitem{ShuffleNetV2}
N.~Ma, X.~Zhang, H.-T. Zheng, and J.~Sun, ``Shufflenet v2: Practical guidelines for efficient cnn architecture design,'' in \emph{European Conference on Computer Vision}, ser. ECCV'18, 2018, pp. 122--138.

\bibitem{RepVGG}
X.~Ding, X.~Zhang, N.~Ma, J.~Han, G.~Ding, and J.~Sun, ``Repvgg: Making vgg-style convnets great again,'' in \emph{Proceedings of the IEEE/CVF conference on computer vision and pattern recognition}, ser. CVPR'21, 2021, pp. 13\,733--13\,742.

\bibitem{CSPNet}
C.-Y. Wang, H.-Y.~M. Liao, Y.-H. Wu, P.-Y. Chen, J.-W. Hsieh, and I.-H. Yeh, ``Cspnet: A new backbone that can enhance learning capability of cnn,'' in \emph{Proceedings of the IEEE/CVF conference on computer vision and pattern recognition workshops}, 2020, pp. 390--391.

\end{thebibliography}

% \newpage

\begin{IEEEbiography}[{\includegraphics[width=1in,height=1.25in,keepaspectratio]{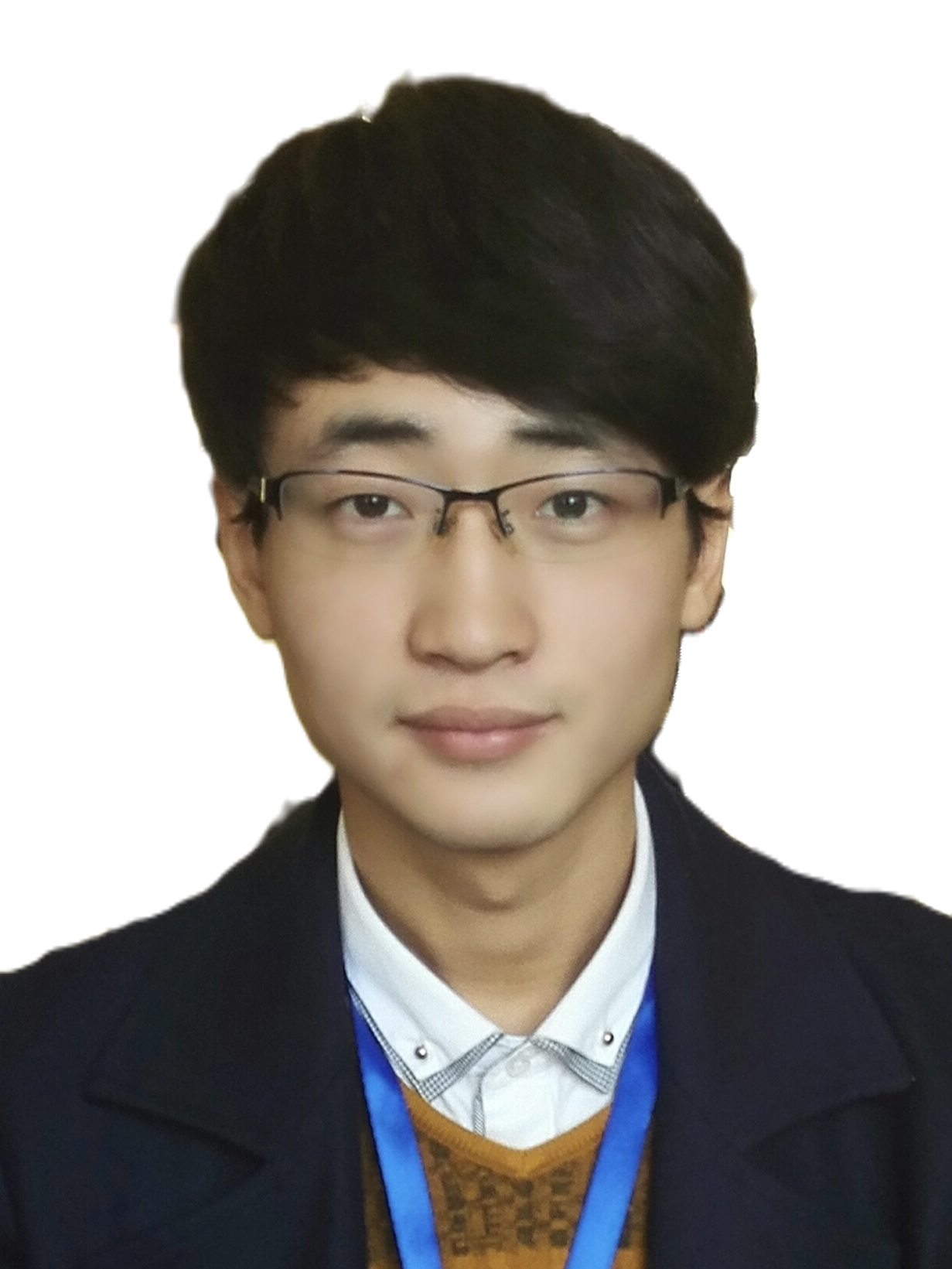}}]{Zhao Li}
is currently pursuing a PhD degree in Computer Science and Technology, College of Intelligence and Computing, Tianjin University, Tianjin, China. His current research interests include knowledge graph, representation learning and reasoning, and Responsible AI. He is the president of the Tianjin University Student Chapter of the China Computer Federation.
\end{IEEEbiography}

\begin{IEEEbiography}[{\includegraphics[width=1in,height=1.25in,keepaspectratio]{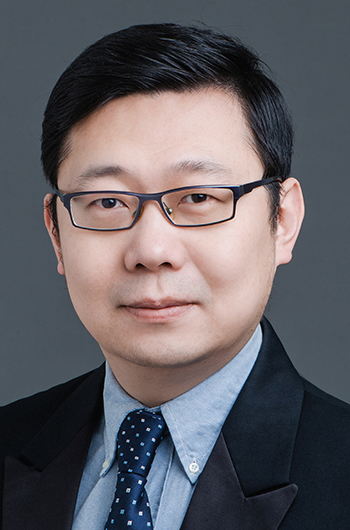}}]{Xin Wang}
received the BE and PhD degrees in computer science and technology from Nankai University in 2004 and 2009, respectively. He is currently a professor in the College of Intelligence and Computing, Tianjin University. His research interests include knowledge graph data management, graph databases, and big data distributed processing. He is a member of the IEEE and the IEEE Computer Society.
\end{IEEEbiography}

\begin{IEEEbiography}[{\includegraphics[width=1in,height=1.25in,keepaspectratio]{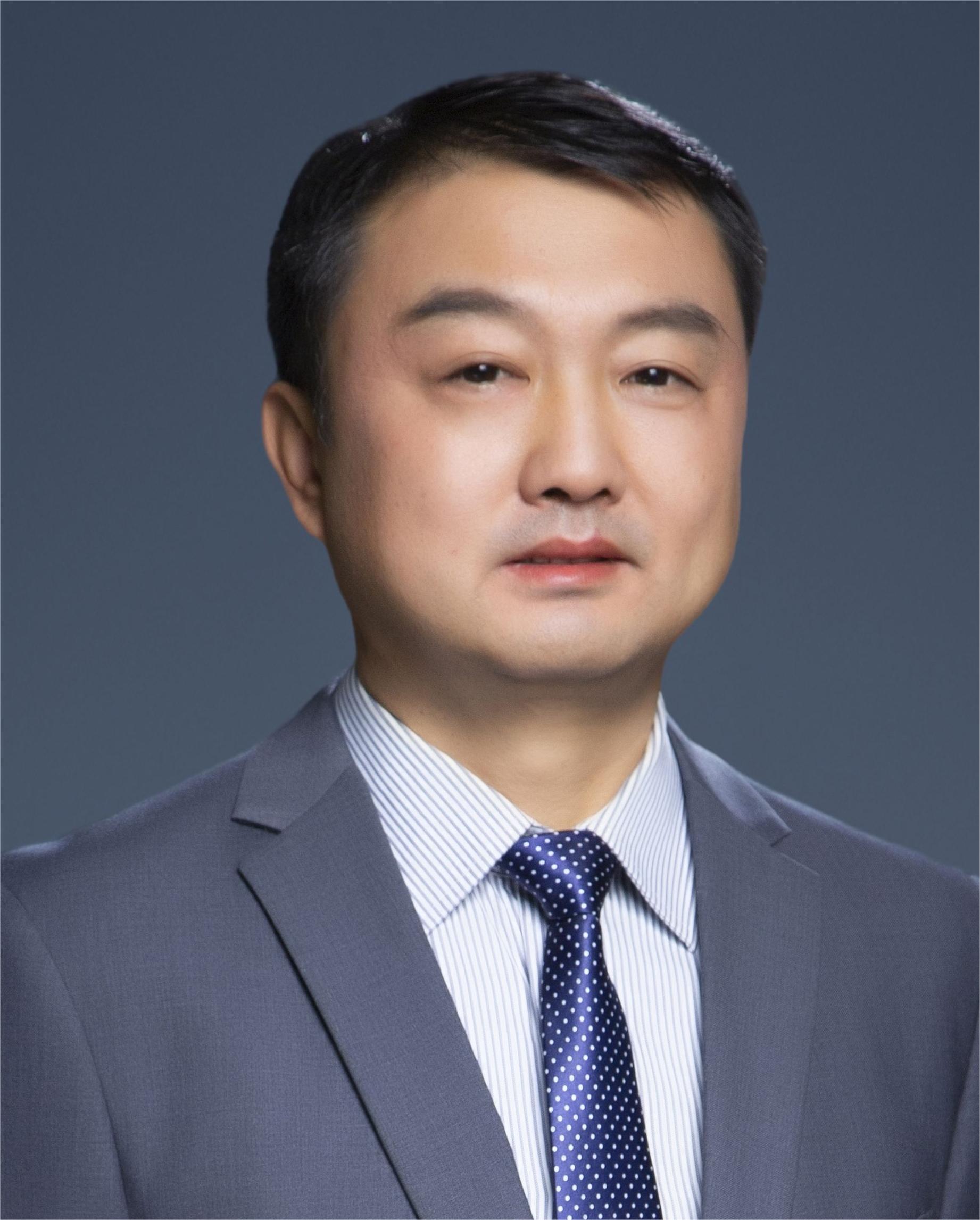}}]{Jun Zhao}
received Ph.D. degree in management science and engineering from the Beijing Institute of Technology, China, in 2006. He is the former dean of the business school and currently is a professor in the economic and management of school, Ningxia University. His research interests include information system, computational experiments, and big data intelligent applications.
\end{IEEEbiography}

% \begin{IEEEbiography}[{\includegraphics[width=1in,height=1.25in,keepaspectratio]{figure/ff.jpg}}]{Feng Feng} is currently a full professor (PhD) and Dean of the School of Information Engineering, Ningxia University. His research topics are about Information systems integration and application, Internet of Things, and so on. He also obtained many awards, including the "Ningxia Science and Technology Progress Award", the "Ningxia University Teaching Achievement Award" (three times). He is a distinguished expert of "Yinchuan Innovation and Development Think Tank".
% \end{IEEEbiography}

\begin{IEEEbiography}[{\includegraphics[width=1in,height=1.25in,keepaspectratio]{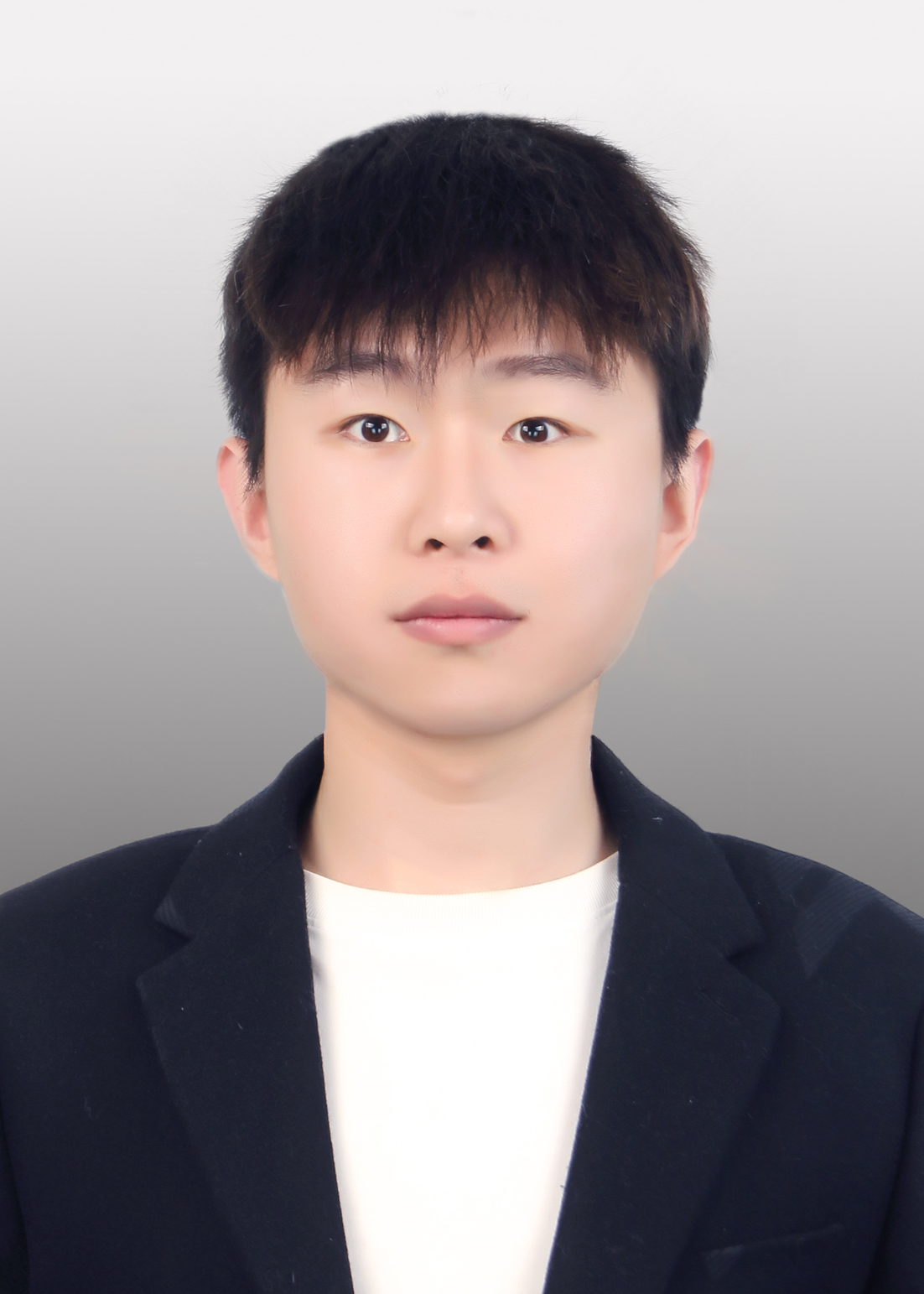}}]{Wenbin Guo}
is currently pursuing a PhD degree in Computer Science and Technology, College of Intelligence and Computing, Tianjin University, Tianjin, China. His current research interests include knowledge representation learning.
\end{IEEEbiography}

\begin{IEEEbiography}[{\includegraphics[width=1in,height=1.25in,keepaspectratio]{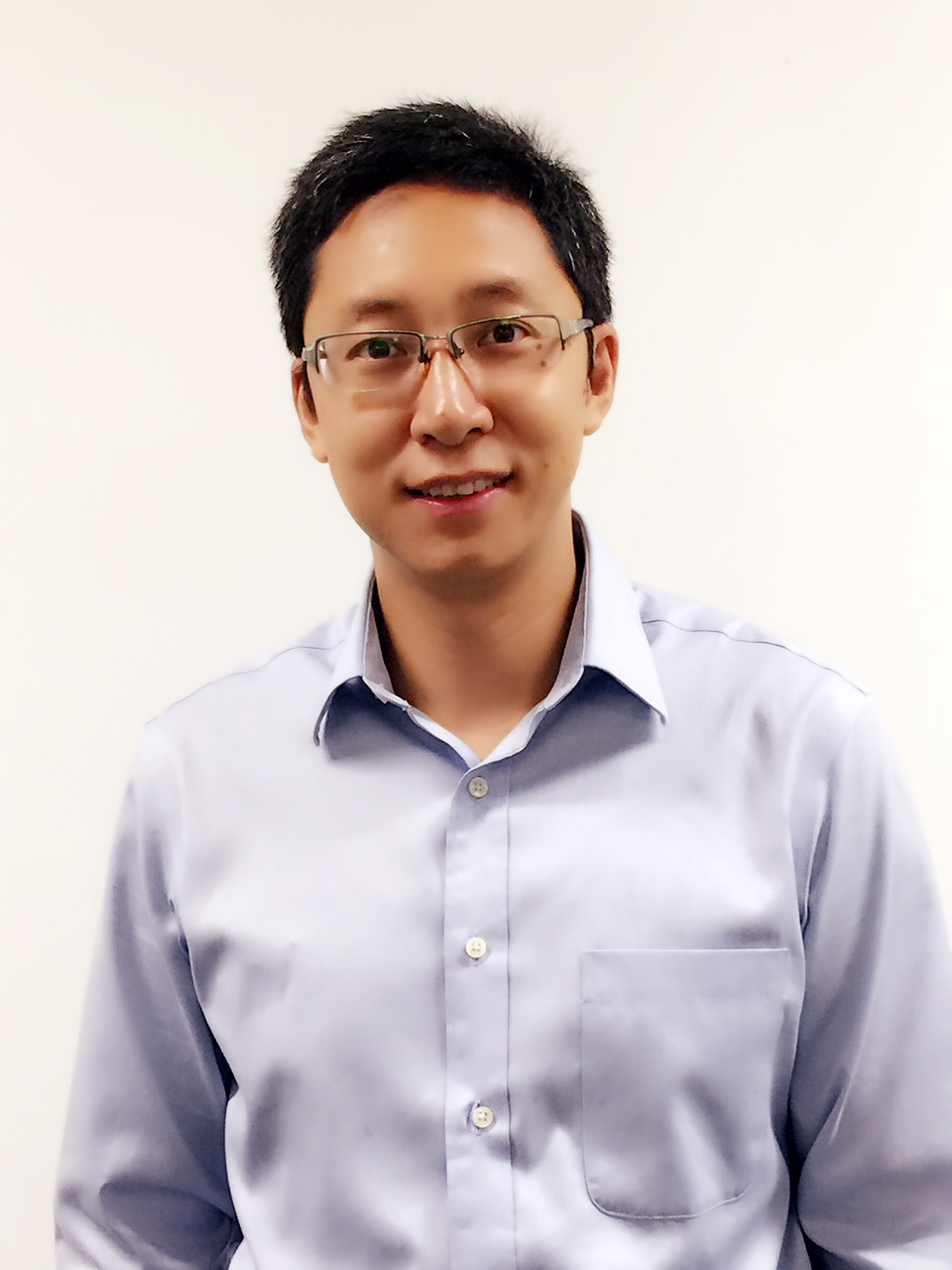}}]{Jianxin Li}
received a Ph.D. degree in computer science from the Swinburne University of Technology, Melbourne, VIC, Australia, in 2009. He is a professor of Information Systems at the School of Business and Law, Edith Cowan University, Joondalup, Australia. He was awarded as World Top 2\% scientists by 2023 Stanford due to his research impact and high citation. His current research interests include database query processing and optimization, social network analytics, and traffic network data processing. He is a senior member of the IEEE Computer Society.
\end{IEEEbiography}

% \vfill

\end{document}